# RareLens: Towards End-to-End Rare Disease Care via Aligning Divergent Large Language Model Reasoning

Xi Chen[1,†], Hongru Zhou[2,†], Shiyu Feng[3,4,†], Hanyu Zhou[5], Huahui Yi[4], Rongsheng Wang[6], Tiancheng He[7], Kun Wang[8], Pingping Liu[9], Qiankun Li[8], Sicheng Lin[10], Huiying Ou[11], Xiaohong Zheng[12], Tianying Zang[13], Zhuohang Wu[10], Leheng Jiang[2], Kexin Cao[10], Wenhan Zhang[10], ChengYi Li[10], Zhiyang Wang[10], Songlin Li[14], Benyou Wang[3], Ningbei Yin[2], Shaoting Zhang[15], Weili Fu[1,*], Jian Li[1,*], Kang Li[4,16,*]

*† These authors contributed equally to this work.*

**Affiliations**

[1] Sports Medicine Center, Department of Orthopedics and Orthopedic Research Institute, West China Hospital, Sichuan University, Chengdu, China

[2] Center for Cleft Lip and Palate Treatment, Plastic Surgery Hospital, Chinese Academy of Medical Sciences and Peking Union Medical College, Beijing, China

[3] School of Data Science, The Chinese University of Hong Kong, Shenzhen (CUHKSZ), Shenzhen, China

[4] West China Biomedical Big Data Center, West China Hospital, Sichuan University, Chengdu, China

[5] School of Computer Science, Carnegie Mellon University, Pittsburgh, Pennsylvania, USA

[6] School of Medicine, The Chinese University of Hong Kong, Shenzhen (CUHKSZ), Shenzhen, China

[7] School of Artificial Intelligence and Automation, Huazhong University of Science and Technology, Wuhan 430074, China

[8] College of Computing and Data Science, Nanyang Technological University (NTU), Singapore 639798, Singapore

[9] Plastic Surgery Hospital, Chinese Academy of Medical Sciences and Peking Union Medical College, Beijing, China

[10] Pittsburgh Institute, Sichuan University, Chengdu, China

[11] West China School of Medicine, Sichuan University, Chengdu, Sichuan 610041, P. R. China

[12] The First Affiliated Hospital of Sun Yat-sen University, Guangzhou, Guangdong, China

[13] Department of Aesthetic Plastic Surgery and Laser Medicine, Beijing Anzhen Hospital Affiliated to Capital Medical University, Beijing, China

[14] Department of Orthopedics, Qilu Hospital of Shandong University, Jinan, Shandong, China

[15] Shanghai Artificial Intelligence Laboratory, OpenMedLab, Shanghai, China

[16] Med-X Center for Informatics, Sichuan University, Chengdu, Sichuan 610041, P. R. China

[*] **Correspondence:** Weili Fu (foxwin2008@163.com); Jian Li (lijian_sportsmed@163.com); Kang Li (likang@wchscu.cn).

**RareLens: Towards End-to-End Rare Disease Care via Aligning Divergent Large Language Model Reasoning**

**Abstract**

Rare diseases represent one of the most challenging settings for clinical decision-making, where heterogeneous presentations, sparse evidence and limited expertise create persistent uncertainty throughout the care pathway. Although artificial intelligence could help, existing systems largely address isolated tasks, particularly diagnosis, and usually rely on downstream investigations rather than information available at initial presentation. Here we show that clinical AI performance under uncertainty can be improved not by scaling a single model, but by exploiting the diversity of multiple imperfect reasoning systems. Across heterogeneous large language models, we identify divergent reasoning trajectories with complementary error patterns and develop RareLens, which learns to reconcile these perspectives into actionable decisions across four stages of rare disease care: risk screening, diagnosis, treatment planning and prognosis prediction. Built on RarelensBench, a real-world dataset of 157,525 cases spanning all 33 Orphanet categories and more than 7,000 conditions, RareLens outperformed every frontier model tested, including GPT-5, DeepSeek-R1, Claude-3.7-Sonnet and Gemini-2.5-Pro, across all stages. It achieved an area under the curve of 0.917 for screening and top-1 accuracies of 65.5% and 89.8% for diagnosis and treatment. In an external evaluation involving 1,287 cases and 23 physicians, autonomous RareLens and physicians assisted by RareLens both outperformed unaided physicians, while demonstrating that effective human–AI collaboration requires more than simply providing model outputs. These findings establish divergent model reasoning as an exploitable source of information and suggest a general strategy for building AI systems that operate reliably under high clinical uncertainty.

## Introduction

Rare diseases represent an extreme challenge for clinical decision-making under uncertainty. Although each individual condition is uncommon, collectively rare diseases affect an estimated 3.5% to 5.9% of the population and encompass more than 7,000 distinct disorders across diverse clinical domains (1-3). Despite this substantial burden, patients frequently experience prolonged diagnostic journeys, with more than 70% experiencing at least one incorrect or delayed diagnosis before a definitive diagnosis is established (4, 5). This challenge arises not only from disease rarity, but from a fundamental mismatch between the nature of rare disease presentations and conventional clinical reasoning: early manifestations are heterogeneous, overlap with common disorders, and provide limited evidence on which to base decisions.

A major source of diagnostic delay occurs before diagnosis begins: rare disease is often not considered at the initial clinical encounter, and targeted investigations are therefore never initiated. At first presentation, clinicians make decisions based primarily on demographics, medical history and physical examination findings, while individual exposure to rare diseases remains limited because each condition is encountered infrequently. Consequently, patients with rare diseases are often managed along pathways designed for common conditions, and delay accrues before the correct work-up begins. Identifying these patients at the first encounter is therefore fundamentally a problem of recognition in unselected populations: any solution should operate across all presenting patients, not only those in whom rare disease is already suspected (6-8).

Beyond early recognition, rare disease care is inherently a longitudinal process spanning diagnostic, therapeutic, and prognostic decisions, each stage marked by substantial clinical complexity, uncertainty, and resource demands (4). At each stage, expert clinicians rarely rely on a single line of reasoning; instead, they weigh multiple competing hypotheses and integrate diverse perspectives before reaching a final judgement (9). This form of expertise is scarce: it is concentrated within specialised centres and remains inaccessible to many patients worldwide (10). An approach that can reproduce expert-level reasoning and be deployed where such expertise is absent could therefore begin to narrow this global inequity in rare disease care.

Artificial intelligence has the potential to support clinical decision-making in rare diseases, where knowledge is dispersed and expertise is limited. Large language models (LLMs) have demonstrated the ability to encode medical knowledge and generate clinically plausible reasoning (11, 12), while recent advances in tool-augmented and multi-agent systems have extended their ability to support diagnostic tasks and structured clinical workflows (13-16). However, existing approaches remain constrained in two important ways. First, they predominantly focus on individual stages of care, particularly diagnosis, rather than supporting the full trajectory from initial risk recognition to treatment and prognosis. Second, they generally rely on a single model or seek consensus among models, implicitly treating disagreement as uncertainty or error to be reduced. Yet in complex clinical decision-making, alternative perspectives may contain complementary information. Whether divergent reasoning among AI systems can be systematically exploited to improve decision-making under uncertainty remains unknown.

Here we hypothesised that the divergent reasoning produced by heterogeneous AI systems could serve as an informative source for improving clinical decision-making under uncertainty. We developed RareLens, an artificial intelligence system that leverages this principle across the entire rare disease care trajectory, from primary-visit risk screening to diagnosis, treatment planning and prognosis prediction. Rather than relying on a single model or forcing agreement among models, RareLens generates multiple independent reasoning trajectories and learns to reconcile their complementary signals into a calibrated clinical decision.

To investigate this hypothesis, we established RarelensBench, a real-world clinical benchmark comprising longitudinal rare disease cases spanning four sequential stages of care: primary-visit screening, diagnosis, treatment and prognosis. RarelensBench was designed to evaluate clinical decision-making under progressively increasing information availability while preserving the constraints of real-world workflows. We assessed RareLens through independent evaluations across different stages of care and an external physician study involving 23 clinicians and 1,287 cases. Together, these evaluations demonstrate that aligning divergent reasoning trajectories from heterogeneous AI systems can improve rare disease decision-making beyond individual models and conventional ensemble approaches, suggesting a general framework for developing reliable AI systems in settings characterized by high uncertainty.

## Results

### Overview of RareLens

Rare disease care unfolds as a sequence of decisions made under high but progressively diminishing uncertainty, beginning with the recognition of rare disease risk at the first clinical encounter and extending through diagnosis, treatment and prognosis (Fig. 1a). We developed RareLens, an artificial intelligence system that supports this entire trajectory through a single mechanism: for each patient, a panel of heterogeneous large language models independently generates divergent clinical reasoning, each output comprising a stage-specific prediction, the key clinical factors supporting it, and an explanatory rationale. Rather than selecting one model or averaging their outputs, RareLens treats these parallel reasoning paths as candidate evidence and applies a machine-learning alignment step that calibrates and weights them into a single convergent, actionable decision (Fig. 1b).

RareLens comprises four coordinated modules that mirror successive stages of care, each operating on the information available at its stage: RareAlert performs risk screening at the primary visit, RareDiagnosis generates and confirms diagnostic hypotheses, RareTreatment formulates management strategies, and RarePrognosis predicts outcomes (Fig. 1a). The output of each module provides structured context for the next, so that information propagates along the clinical trajectory as in routine practice. Because screening is intended for universal application at every primary encounter, the aligned reasoning underlying RareAlert is additionally distilled into a compact, locally deployable model (Methods). RareLens was developed and evaluated on RarelensBench, a real-world dataset of 157,525 cases. In the sections that follow, we characterise this dataset, establish that inter-model reasoning divergence constitutes a complementary signal rather than noise, and then demonstrate state-of-the-art performance at each stage of care and in an end-to-end study against physicians.

### RarelensBench

To develop and evaluate RareLens, we curated RarelensBench, a real-world clinical dataset of 157,525 cases, comprising 42,836 rare disease cases and 114,689 non-rare disease cases. The rare disease cohort spans all 33 Orphanet disease categories and covers more than 7,000 distinct conditions, providing broad representation across the rare disease spectrum and across clinical specialties. Fig. 2 provides an overview of the design and construction of RarelensBench.

RarelensBench was designed to meet three requirements that existing rare disease benchmarks do not jointly satisfy. First, it includes a large population of non-rare disease cases alongside rare ones, so that rare disease identification can be evaluated as a screening task in which both false negatives and false positives are clinically consequential, rather than as diagnosis among patients in whom rare disease is already suspected (Fig. 2a). Second, each rare disease case is organised as a longitudinal clinical trajectory of four stages with progressively increasing information content and distinct clinical tasks. The primary visit provides demographics, medical history and physical examination findings for risk assessment and hypothesis generation; the follow-up visit adds laboratory and imaging results for diagnostic confirmation; the treatment visit incorporates the confirmed diagnosis for treatment planning; and the prognosis visit adds treatment and outcome information for prognostic assessment. Each stage exposes only the information available at the corresponding point in real-world care, preventing information leakage across stages (Fig. 2b). Third, to support rare disease screening, the dataset should include non-rare disease cases. Cases were drawn from three sources, namely PubMed-indexed published case reports, MIMIC-IV-Ext-BHC and MIMIC-IV-Ext-CDM, and were processed through a standardised pipeline for quality control, rare disease identification and structured curation (Methods).

**Inter-model Reasoning Divergence**

The alignment mechanism rests on an empirical premise: that different models reason differently about the same case. We observed substantial divergence at every stage of care. In risk screening, two models assigned identical risk scores in fewer than 20% of cases (range 5% to 49%) (Fig. 3f), with mean absolute differences of up to 26 points (Fig. 3g); structured analysis of their rationales identified on average six distinct reasoning chains per case, with any two models co-occurring in the same chain in fewer than 25% of cases (Fig. 3h). In diagnosis, models generated an average of 26.7 unique candidate diagnoses per case at the primary visit (Supplementary Fig. 1a), their top-5 lists overlapping at a mean Jaccard similarity of only 0.18 (Supplementary Fig. 1e), and even when two models proposed the same diagnosis its rank differed by 1.7 positions on average (Supplementary Fig. 2e). In treatment planning, models diverged in both the interventions selected and their prioritisation across pharmacological, rehabilitative, nutritional, preventive and surveillance categories (Supplementary Fig. 3). In prognosis, pairwise agreement ranged from 16.8% to 79.1% for overall outcome (Supplementary Fig. 4b), 53.2% to 78.6% for functional status (Supplementary Fig. 4c) and 75.1% to 89.5% for symptom burden (Supplementary Fig. 4a).

This divergence carried recoverable information. In risk screening, RareAlert correctly reclassified 47% to 74% of the cases misclassified by each individual model, including strong frontier models (Fig. 3k), and its advantage was greatest where reasoning diverged most: among cases failed by at least ten models simultaneously, it still recovered 39% (Fig. 3j). The gain was therefore concentrated on the hardest, most ambiguous cases. Consistent with this, learned alignment outperformed mean, median and mode aggregation of the same model outputs in every task. This informative divergence is the signal RareLens exploits at each stage, as shown in the next section.

**Performance Across the Care Trajectory**

**RareAlert**

Delayed diagnosis frequently originates at the primary visit, where rare disease risk goes unrecognised and targeted investigation is never initiated. RareAlert addresses this step by estimating patient-level rare disease risk from the information available at the first encounter alone, namely demographics, medical history, and physical examination findings. Rather than triaging patients in whom rare disease is already suspected, RareAlert is designed to be applied universally to every patient at presentation, reframing early rare disease detection as a population-scale screening task. The aligned reasoning is distilled into RareAlert, a single locally deployable 32B language model (Methods).

On an independent test set of 31,505 cases, RareAlert achieved an area under the receiver operating characteristic curve (AUC) of 0.917, the highest of all evaluated approaches (Fig. 3b). It exceeded every one of the 16 large language models assessed, including the frontier models (GPT-5, 0.905; o3-mini, 0.897; Claude-3.7-Sonnet, 0.896; DeepSeek-R1, 0.880; Gemini-2.5-Pro, 0.882; Qwen3-235B, 0.859). It also outperformed the machine-learning model that integrated the risk scores of all constituent large language models, the alignment step from which RareAlert is distilled. Further analysis showed AUC performance across clinical specialties (Supplementary Fig. 5).

At the operating threshold defined by the Youden index, RareAlert reached a sensitivity of 0.778 and a specificity of 0.935 (Supplementary Fig. 6f,g), correctly excluding the large majority of non-rare presentations while maintaining a positive predictive value of 0.818 (Fig. 3e). It ranked first among all evaluated models on specificity, positive predictive value, and on composite measures of overall discrimination, with detailed results reported in Supplementary Fig. 6. Further analysis confirmed that RareAlert ranked first in overall accuracy (0.892), balanced accuracy (0.856), F1 score (0.797), F2 score (0.785), and Matthews correlation coefficient (0.668), while ranking second in negative predictive value (0.918) among all models. The complete metric panel is reported in Fig. 3 and Supplementary Fig. 6.

**RareDiagnosis**

RareDiagnosis supports diagnostic reasoning at two points in the workup. At the primary visit it takes demographics and free-text clinical history and returns a ranked list of candidate diseases together with recommended confirmatory investigations; at the follow-up visit it incorporates the resulting laboratory and imaging findings to refine the ranking and converge on the diagnosis.

RareDiagnosis was evaluated against 11 large language models, including GPT-5, o3-mini, DeepSeek-R1, Gemini-2.5-Flash, and Qwen3-235B, on 8,567 cases. It achieved the highest diagnostic accuracy of all evaluated models at both stages (Fig. 4b,c). At the primary visit it reached top-1, top-3, and top-5 accuracies of 65.5%, 75.9%, and 79.6%, exceeding the strongest comparator (GPT-5: 57.1%, 68.6%, and 72.3%) (Fig. 4b). After follow-up information was incorporated, accuracy rose to 88.4%, 94.5%, and 96.1%, again above the best baseline (GPT-5: 81.0%, 90.0%, and 91.9%) (Fig. 4c). RareDiagnosis recovered the correct top-1 diagnosis in 26% to 51% of primary-visit cases misdiagnosed by each individual model, and in 15.3% of cases failed by at least ten models simultaneously (Supplementary Fig. 7c,d).

The ordering of candidates showed the same advantage. RareDiagnosis attained the highest mean reciprocal rank, which captures how near the top of the list the first correct diagnosis appears, reaching 0.710 at the primary visit and 0.915 at follow-up and exceeding the second-best model by 0.080 and 0.060 respectively (Fig. 4f,j). It also ranked first at every cut-off on normalised discounted cumulative gain, which rewards placing the more clinically probable diagnoses higher in the ranking (Fig. 4d,e).

These gains reflect a concentration of the divergent hypothesis space onto the correct diagnosis. From an average of 26.7 distinct candidate diagnoses generated across models per case (Supplementary Fig. 1a,b), alignment produced a single short ranked list that placed the correct diagnosis within the top five in 79.6% of cases at the primary visit and 96.1% at follow-up (Fig. 4b,c).

Stratified by clinical specialty, RareDiagnosis ranked first in the majority of specialties at both stages (top-1 accuracy in 33 of 43 specialties at the primary visit and 30 of 38 at follow-up), with the advantage most consistent in adequately sized specialties (Fig. 4g,h). Full results for the remaining metrics and for the recommended diagnostic investigations are provided in Fig. 4, Supplementary Fig. 7 and Supplementary Fig. 8.

**RareTreatment**

RareTreatment supports management after a diagnosis is established. To ground treatment generation in real-world evidence, we constructed a rare disease treatment knowledge base from the Orphanet database and real-world case records, covering 1,021 diseases, of which 371 are additionally supported by one to two real cases (621 cases in total), and used it for retrieval-augmented generation when the models produced treatment plans. Given patient-specific information, it produces a ranked list of treatment recommendations, each annotated with category, implementation details, expected response, and safety considerations, together with short-, medium-, and long-term goals.

On an independent test set of 452 cases, RareTreatment recommended appropriate treatments more accurately than all 12 large language models evaluated (Fig. 5b). Its top-1 accuracy reached 89.8%, exceeding the strongest comparator (GPT-5, 83.0%) and the remaining models (71.0% to 81.9%), and it attained the highest mean reciprocal rank (0.929 versus 0.779 to 0.846) and the highest normalised discounted cumulative gain at every cut-off, indicating that clinically appropriate interventions were consistently ranked first (Fig. 5c,d).

This ranking advantage reflected a favourable balance between coverage and prioritisation. Although several models proposed comparably broad treatment sets, RareTreatment recommended the largest number of appropriate treatments per case (8.7) while issuing the fewest inappropriate ones among models of similar breadth (1.3), and it produced the lowest proportion of high-risk recommendations of any model (1.8%) (Fig. 5e,f, Supplementary Fig. 9a). Further analysis of completeness, helpfulness, safety, the fraction of recommendations subsequently administered, and per-model rescue is provided in Fig. 5 and Supplementary Fig. 9.

**RarePrognosis**

RarePrognosis operates after treatment planning, predicting a patient's long-term trajectory across three dimensions: overall outcome (from complete recovery to terminal), functional status (none to severe limitation), and symptom burden (none to persistent severe). For each dimension it returns a categorical prediction together with a confidence score and a concise clinical rationale, and it additionally forecasts anticipated clinical events with their expected type and timing.

On an independent test set of 3,734 cases, RarePrognosis predicted all three dimensions more accurately than every one of the 12 large language models evaluated (Fig. 6b). It reached an accuracy of 62.4% for overall outcome (other models, 28.8% to 58.4%), 63.3% for functional status (40.8% to 57.4%), and 67.0% for symptom burden (43.9% to 54.9%), and attained the highest Matthews correlation coefficient on each dimension (Fig. 6c). Under the stricter joint criterion of predicting all three dimensions correctly for the same patient, RarePrognosis succeeded in 33.9% of cases, compared with at most 28.3% for any individual model (Fig. 6b). Further analyses of severe-outcome recall, false-optimism rate, ordinal error, temporal event prediction, and per-model rescue are provided in Fig. 6 and Supplementary Fig. 10.

**Human-AI collaboration evaluation**

To evaluate the clinical value of RareLens in a realistic deployment scenario, we tested the complete cascade on an external cohort independent of all development data, comprising 371 rare disease cases and 916 non-rare cases. We compared three conditions, RareLens alone, physicians unaided, and physicians assisted by RareLens, processing each case sequentially through the full pathway on a common fixed denominator (Fig. 7a). A total of 23 practicing physicians from 10 hospitals, all of whom had completed standardized residency training, participated in the study. Under the full end-to-end criterion, requiring a case to be correct at every stage from alert through prognosis, RareLens succeeded in 21.3% of rare cases, compared with 10.0% for assisted and 0.8% for unaided physicians (Fig. 7b).

This gap emerged progressively along the cascade (Fig. 7c). At the alert gate, overall accuracy was similar across groups (76.4% to 82.9%), but sensitivity, the metric relevant to screening, differed markedly: 0.784 for RareLens versus only 0.369 for unaided physicians, who missed roughly two-thirds of rare cases. RareLens assistance improved physician sensitivity to 0.569 with essentially unchanged specificity. Cumulative accuracy among rare disease cases then fell steeply after primary diagnosis, the principal bottleneck, to 52.3% for RareLens, 32.4% for assisted physicians and 7.0% for unaided physicians, and this ordering persisted through treatment and prognosis. Across every downstream stage, RareLens assistance roughly doubled physician yield while RareLens alone led throughout, indicating that the main barrier to end-to-end care is recognition at the earliest visits.

We computed, on paired cases, the rate at which RareLens assistance rescued physicians' errors

(turning incorrect answers correct) versus regressed their correct answers. Rescue rates consistently exceeded regression rates: 49.8% versus 7.3% at screening (n=1,287), 47.4% versus 4.8% at primary diagnosis (top-5) and 80% versus 0% at treatment (top-5), with a similar pattern at follow-up diagnosis and prognosis, though downstream estimates are limited by small paired samples. Detailed results are provided in Fig 7.

**Efficiency and Cost**

With our current models and hardware, RareLens processed a rare disease case end-to-end in an average of 11 minutes: approximately 1.4 minutes for risk assessment, 2.8 minutes for primary diagnosis, 2.5 minutes for follow-up diagnosis, 2.7 minutes for treatment planning and 1.8 minutes for prognosis. In terms of cost, RareAlert runs locally, whereas the downstream stages call external LLMs, at an average cost of approximately US$0.46 in LLM calls per case.

**Reliability Analysis**

To assess stability under stochastic decoding, we re-ran RareLens three times on a random subset of 500 cases per module. The principal outputs were highly reproducible across runs (screening risk score ICC 0.930; diagnostic hit agreement 0.80 to 0.91; treatment retrieval Gwet's AC1 up to 0.99; prognostic dimensions weighted κ 0.84 to 0.89). Detailed results are available in the Supplementary Information.

**Comparison with other Ensemble Techniques**

We benchmarked each module against a panel of training-free ensemble baselines spanning the score-fusion, rank-aggregation, voting and central-tendency families, matched to each task's output type (Supplementary Table 3). Across all five modules RareLens outperformed every baseline on the primary metric, and no single ensemble rule was competitive across tasks.

## Discussion

We show that divergent reasoning among heterogeneous artificial intelligence systems can be transformed from a source of uncertainty into a source of clinically useful information. By learning to reconcile multiple independent reasoning trajectories, RareLens improved decision-making across the entire rare disease care continuum, from primary-visit risk screening to diagnosis, treatment planning and prognosis prediction. In an external evaluation involving 23 physicians, autonomous RareLens and physicians assisted by RareLens both substantially outperformed unaided physicians in end-to-end rare disease management. These findings suggest that reliable clinical AI may not emerge solely from developing increasingly capable individual models, but from designing systems that can integrate diverse and complementary perspectives to support complex decisions under uncertainty.

**Divergent reasoning**

The central conceptual contribution of this work is to show that divergent AI reasoning can be formalised as a computable and exploitable signal for clinical decision-making. Clinical reasoning under uncertainty is inherently pluralistic. Experienced physicians rarely commit immediately to a single explanation; instead, they maintain several competing hypotheses simultaneously and iteratively update their beliefs as new evidence becomes available (17, 18). We found that heterogeneous LLMs exhibited a comparable pattern of diversity: when presented with the same case, different models generated substantially different risk assessments, diagnostic hypotheses, treatment choices and prognostic predictions across the care trajectory. Such inter-model variability is consistent with previous observations that different LLMs possess complementary capabilities and partially distinct error patterns in clinical tasks, allowing their integration to exceed individual model performance (19, 20). Crucially, our analyses demonstrate that this divergence is informative rather than random noise: across tasks, RareLens recovered a substantial fraction of cases misclassified by individual models, including frontier systems, and its advantage was greatest in cases where model disagreement was highest.

We propose that heterogeneous LLMs can therefore be viewed as representing distinct reasoning priors. These differences may arise from variations in training data distributions, optimisation objectives and architectural inductive biases, causing models to attend preferentially to different aspects of the same clinical presentation. Consequently, no individual model provides a complete representation of the underlying clinical decision landscape. RareLens does not simply aggregate model outputs; instead, it learns when different reasoning priors are informative and calibrates their relative contributions. This may explain why conventional ensemble strategies consistently underperformed supervised alignment across all evaluated tasks.

More broadly, these findings suggest that diversity among foundation models should not necessarily be regarded as a limitation to overcome through model scaling or consensus voting. Instead, diversity itself may represent a computational resource that can be explicitly modeled and exploited (21). Although demonstrated here in rare disease care, this principle may extend to other high-uncertainty decision problems in medicine and beyond, where multiple imperfect systems capture complementary aspects of an underlying problem.

**Rare Disease Screening**

Treating rare disease identification as universal screening at first presentation, rather than as diagnosis applied only to already-suspected patients, represents a fundamental reframing of the clinical problem. Rare diseases collectively affect an estimated 3.5 to 5.9% of the population (3), yet patients endure an average of 4.7 years of diagnostic delay and frequent misdiagnosis before a diagnosis is reached, because early presentations are nonspecific and span every specialty (4). The decisive bottleneck lies at the very first encounter: unless the possibility of rare disease enters clinical consideration, confirmatory investigations such as genetic testing are unlikely to be initiated, and patients may never enter the appropriate diagnostic pathway. Primary-visit recognition is therefore arguably the most consequential step in the entire rare disease care trajectory. Unlike diagnosis, which asks which rare disease a suspected patient has, primary-visit recognition must be applied broadly to determine which patients warrant further evaluation for rare disease.

This detection task is not a simplified form of diagnosis but a distinct inference problem. At the first visit, relevant evidence is present but weak: individually nonspecific signals are distributed across the clinical presentation, and the challenge is to integrate these signals rather than reason more deeply along a single trajectory. Aligning complementary reasoning from heterogeneous models may therefore provide an advantage by assembling dispersed evidence into a calibrated risk estimate. Consistent with this view, RareAlert achieved the highest primary-visit discrimination among all 16 large language models evaluated (AUC 0.917), exceeding frontier reasoning models including GPT-5, o3-mini and Claude-3.7-Sonnet. Its advantage was greatest in cases with maximal model disagreement, recovering 39% of cases missed simultaneously by at least ten models. Because RareAlert produces a calibrated risk estimate rather than a fixed binary decision, thresholds can be adapted according to local resources and the acceptable balance between missed cases and unnecessary referrals. Its compact local deployment design may further extend rare disease risk assessment to settings where specialist expertise is limited.

### End to End Care

Recognising rare disease at the first encounter is only the entry point to a much longer process, and effective support cannot stop there. Rare disease care is inherently longitudinal and dynamic: information accumulates across encounters, uncertainty progressively changes, and decisions at each stage depend on previous assessments while shaping subsequent management. Screening, diagnosis, treatment and prognosis are therefore not independent tasks, but interconnected decisions within a single clinical trajectory. An AI system that addresses only one stage cannot preserve and propagate information across these transitions, leaving critical parts of the care pathway unsupported.

RareLens was designed around this principle. Its four modules function as coordinated stages rather than independent tools, with outputs from each stage providing structured context for the next. This design allows information to propagate along the clinical trajectory, approximating the sequential accumulation of evidence that characterises real-world rare disease care. Unlike systems optimised for isolated diagnostic accuracy, RareLens maintains continuity across recognition, diagnosis, treatment, and prognosis prediction, supporting decision-making across diverse diseases and specialties rather than a limited set of predefined conditions.

### Human-AI collaboration

Our external evaluation highlights an important principle for clinical AI deployment: the benefit of human–AI collaboration is not guaranteed simply by providing clinicians with access to a high-performing model. RareLens alone achieved the highest end-to-end performance, followed by physicians assisted by RareLens and unaided physicians (full end-to-end success of 21.3%, 10.0% and 0.8%, respectively). Although RareLens assistance substantially improved physician performance, the collaborative condition did not reach the performance achieved by the model alone. This pattern is consistent with a meta-analysis of 106 studies showing that human–AI combinations are, on average, less accurate than the better partner and tend to lose performance precisely when the AI alone outperforms the human (22), and with a randomized trial in which LLM access did little to improve physicians' diagnostic reasoning despite the model alone scoring substantially higher (23).

Although we did not directly measure the mechanisms underlying this pattern, several established factors may contribute. Algorithm aversion can reduce the adoption of AI recommendations that conflict with clinicians' own judgement (24), while anchoring on an initial hypothesis may limit the extent to which clinicians revise their assessment after receiving additional information (25). These effects may be particularly relevant in rare disease, where the most valuable AI contributions may involve unusual or counterintuitive presentations that challenge existing clinical expectations.

Importantly, our findings do not imply that human–AI collaboration is ineffective. Previous studies, including our own, have nonetheless found that collaboration can improve diagnosis (26-28) with the largest benefits when human and model errors are complementary (26), when less experienced clinicians are supported by reliable AI (27) and when the model engages first to gather and organise information while the clinician reviews its output (28). Our workflow differed from these settings: physicians first interpreted the complete case and committed to an assessment before viewing RareLens output, meaning that the model primarily provided post hoc advice rather than actively shaping information acquisition or hypothesis formation.

How assistance is delivered may therefore matter as much as model accuracy. Future clinical AI systems should be evaluated not only by whether they outperform clinicians or improve average accuracy, but by whether their integration into clinical workflows produces complementary strengths between human judgement and machine inference. Achieving this goal will require task-specific collaboration strategies tailored to the clinician, the model and the decision environment, rather than assuming that human and AI performance simply add together.

**Comparison with other studies**

Prior artificial intelligence approaches for rare disease have primarily addressed diagnosis after clinical suspicion has already been established. Facial phenotype systems such as GestaltMatcher (29), genomic prioritisation tools such as Exomiser (30), and electronic health record-based detectors (31) have demonstrated value by extracting disease-specific signals from particular data modalities or diagnostic contexts. More recent agentic systems have advanced traceable reasoning and differential diagnosis generation, but remain focused on the diagnostic stage (13). RareLens differs from these approaches in three important ways. First, it operates upstream of diagnostic AI by addressing the earlier challenge of identifying which patients should enter the rare disease diagnostic pathway. Second, it supports the complete clinical trajectory from risk screening through diagnosis, treatment planning and prognosis rather than a single isolated decision point. Third, it introduces a different strategy for AI reasoning by explicitly modelling and aligning inter-model diversity as a source of information. RareLens is therefore complementary to existing diagnostic tools, providing an upstream and longitudinal decision framework within which specialised diagnostic approaches can be further integrated.

## Limitations

Several limitations warrant consideration. First, although RareLens was evaluated in an external physician study, the current evaluation remains based on retrospective clinical cases. Whether improved decision-making translates into shorter diagnostic journeys, more appropriate referrals, improved treatment selection and better patient outcomes requires prospective validation in routine clinical workflows across diverse healthcare settings. Such studies will also be essential to determine how RareLens should be integrated into existing clinical pathways. Second, RareLens currently relies primarily on textual clinical information, including structured clinical features and reports derived from multimodal investigations, rather than directly interpreting raw imaging, pathology or genomic data. Extending this framework to incorporate heterogeneous multimodal evidence represents an important future direction, but will require addressing challenges in data availability, disease-specific representation and rigorous evaluation across thousands of rare conditions. Third, RareLens inherits the strengths and limitations of its underlying foundation models. This dependence is partly intentional, allowing the framework to benefit from continued advances in foundation models; however, changes in model behaviour over time may introduce performance drift and require periodic monitoring, recalibration and re-evaluation. In addition, our human–AI evaluation examined one specific interaction paradigm, in which physicians first formed their assessments before receiving RareLens outputs. Alternative collaboration strategies, such as AI-assisted information gathering or iterative dialogue, may produce different outcomes and should be systematically investigated to identify workflows in which human and AI capabilities are truly complementary.

In conclusion, our findings suggest that reliable clinical artificial intelligence may emerge not only from developing increasingly capable individual models, but also from learning how to harness the diversity among imperfect systems. Rather than treating disagreement among foundation models as uncertainty to be eliminated, AI systems can exploit divergent reasoning trajectories as complementary sources of clinical information and integrate them into more reliable decisions. Although demonstrated in rare disease care, this principle may extend to other high-uncertainty clinical settings where evidence is incomplete and multiple perspectives are required. In rare disease specifically, such approaches have the potential to shift care upstream by identifying at-risk patients before diagnostic pathways are activated, while supporting coordinated decisions across diagnosis, treatment and prognosis. Realising this potential will ultimately depend on embedding these systems into clinical workflows in ways that allow human expertise and machine intelligence to complement each other, with the shared goal of reducing diagnostic delay and improving patient care.

## Methods

### Study design and overview

The objective of this study was to develop an artificial intelligence system, RareLens, capable of supporting clinical decision-making across the full continuum of rare disease care, from early risk identification through diagnosis, treatment planning, and prognosis assessment. Unlike prior approaches that focus on isolated tasks, RareLens was designed as a longitudinal decision-support system aligned with real-world clinical workflows. The system comprises four coordinated modules, RareAlert, RareDiagnosis, RareTreatment, and RarePrognosis, corresponding to sequential stages of patient management. These modules operate as an integrated system rather than as independent components: outputs generated at each stage provide structured context for subsequent stages, enabling information to propagate across the clinical trajectory.

All modules were developed and evaluated on RarelensBench, a real-world dataset assembled for this study. Internal evaluation used held-out test sets that were excluded from all model development and selection. External evaluation used an independent multi-reader, multi-case cohort in which RareLens was compared with physicians and with physicians assisted by RareLens.

**RarelensBench construction**

Existing rare disease benchmarks primarily assess diagnostic capability by testing whether language models can correctly diagnose rare diseases when complete patient information is provided. These benchmarks did not meet the requirements of the present study in three respects. First, because RareLens supports the full continuum of rare disease care, the dataset should include longitudinal clinical information across multiple stages rather than diagnosis alone, and should incorporate treatment and prognostic information. Second, because the system targets universal rare disease management, the dataset requires broad coverage across all disease categories with a large number of rare disease cases. Third, to support rare disease screening, the dataset should include non-rare disease cases to enable accurate estimation of false-positive rates and overall system development. We therefore developed RarelensBench, a dataset comprising 157,525 real-world clinical cases, including 42,836 cases of rare diseases spanning all 33 rare disease categories and more than 7,000 distinct rare conditions, together with 114,689 non-rare disease cases.

**Data sources and acquisition**

Cases were drawn from three real-world sources. Rare disease case reports were collected using the Orphanet database, a resource co-funded by the European Commission that catalogues more than 7,000 rare diseases across 33 categories (3). After obtaining the complete disease list and corresponding classifications, we conducted systematic searches for relevant case reports in PubMed using a standardized query format for each disease: "("disease"[MeSH Terms] OR "disease"[All Fields]) AND ("case reports"[Publication Type] OR "case reports"[All Fields])". For example, the search for Down syndrome cases used "("down syndrome"[MeSH Terms] OR ("down"[All Fields] AND "syndrome"[All Fields]) OR "down syndrome"[All Fields] OR "trisomy 21"[All Fields]) AND ("case reports"[Publication Type] OR "case reports"[All Fields])". When a search returned more than 100 results, only the top 100 articles ranked by relevance were retrieved, maintaining manageable data volume while preserving relevance. This process yielded 267,369 cases from published case reports.

Additional cases were obtained from two MIMIC-IV-Ext resources: 159,334 cases from MIMIC-IV-

Ext Brief Hospital Course (MIMIC-IV-Ext-BHC), which contains longitudinal multi-specialty records from a single medical center, and 2,400 cases from the MIMIC-IV-Ext Clinical Decision Making dataset (MIMIC-IV-Ext-CDM), which focuses on emergency department visits for abdominal pathologies. Across all sources, 429,103 cases were collected.

**Quality control**

All collected cases underwent a two-stage screening workflow. First, duplicate cases were removed using case identifiers; where applicable, publications were additionally assessed for eligibility as English-language case reports. Second, each case was evaluated against four inclusion criteria: (1) presence of essential clinical documentation; (2) adequacy of clinical detail within each section to support diagnostic reasoning; (3) clinical plausibility with respect to established medical knowledge and guidelines; (4) diagnostic evidence supporting the reported diagnosis.

Screening was implemented using DeepSeek-V3 via API calls. To develop the automated screening pipeline, 500 cases were randomly sampled from the candidate pool and used for iterative prompt development and parameter tuning. Through repeated testing, the screening prompts were finalized, DeepSeek-V3 was selected as the screening model, and the temperature parameter was fixed at 0. Following screening, 157,525 cases met all inclusion criteria.

**Rare disease identification**

Rare disease status was assigned to all screened cases using a three-step procedure. First, the final diagnosis was extracted from each case using DeepSeek-V3 via API calls; the full case report text was provided as input and the model was instructed to identify the final diagnosis and return it in a standardized JSON format. Second, the extracted diagnosis was matched against disease entries in the Orphanet database in two stages. Exact matching was applied first by comparing extracted diagnoses with Orphanet disease names. For cases without an exact match, semantic matching was conducted using a retrieval-augmented generation approach: all Orphanet disease entries were embedded using the text-embedding-3-large model and indexed with FAISS for efficient nearest-neighbour retrieval. Case-level diagnoses were embedded in the same vector space and used as query vectors; candidate diseases with cosine similarity greater than 0.8 were retrieved, with up to the top three matches retained. These candidates, together with the original diagnosis and their similarity scores, were organized into a structured textual context and provided to GPT-5 nano, which determined whether the diagnosis was semantically equivalent to any retrieved Orphanet disease. Cases identified by either exact or semantic matching were labelled as rare disease cases. Third, cases not classified as rare diseases after the first two steps underwent additional review. This step targeted diagnoses corresponding to rare diseases with low textual similarity or incomplete or non-standard naming. In this step, we used GPT-5 to further determine whether each remaining case corresponded to a rare disease. To ensure accurate identification, we first randomly selected 1,000 cases and used them, with the involvement of human experts, to develop this labelling procedure, including model selection, prompt optimization and temperature setting. After the procedure had been finalized, we validated its effectiveness on an additional, independently sampled set of 1,000 cases. Agreement between AI and human judgments yielded a Cohen's κ of 0.98, with an observed agreement of 99.2%, indicating almost perfect agreement.

Using this procedure, 42,836 cases were identified as rare disease cases and 114,689 cases were classified as non-rare disease cases.

**Data curation and trajectory structuring**

All included cases were curated using an automated pipeline for information extraction, normalization, and structured representation of clinical content. Rare disease cases were mapped to Orphanet disease categories and further assigned to clinical specialties using a predefined specialty taxonomy. Each rare disease case was structured as a longitudinal clinical trajectory consisting of four stages with increasing information content and distinct clinical tasks. Non-rare disease cases were structured at the primary visit stage only, consistent with their role as negative controls for evaluating rare disease risk screening at the initial clinical encounter.

The four clinical stages were defined as follows. (1) Primary visit: this stage represents the initial patient encounter at a primary healthcare facility and includes demographic information, medical history, and physical examination findings; the clinical task is to assess the likelihood of a rare disease, generate candidate diagnoses, and recommend further diagnostic evaluations. (2) Follow-up visit: this stage extends the primary visit by incorporating results from additional investigations, including laboratory tests, radiographic studies, and other relevant evaluations; the task is to interpret accumulated evidence to refine and establish the diagnosis. (3) Treatment visit: this stage extends the follow-up visit by adding confirmed diagnostic information and represents the post-diagnostic phase; the task is to formulate and prioritize an appropriate treatment strategy tailored to the identified disease. (4) Prognosis visit: this stage extends the treatment visit by incorporating information on implemented therapeutic interventions and subsequent clinical data, representing the post-treatment phase; the task is to assess and predict patient prognosis, including anticipated disease course, and long-term outcomes. Each stage was constructed to include only information available at the corresponding point in clinical care, ensuring temporal consistency and preventing information leakage across stages.

Data curation followed an iterative development process similar to that used for screening. A random sample of 500 cases was used to develop the pipeline, in which we determined the language models, temperature settings, prompt designs, and the number of API calls per case. DeepSeek-V3 was selected with the temperature set to 0. Each case was curated using six sequential API calls, with outputs constrained to a stable, predefined JSON schema. Clinical specialty was assigned using GPT-5, which selected the most appropriate category from a predefined specialty list based on the disease name. The full specialty taxonomy is provided in the Supplementary Information.

**Rare disease treatment knowledge base**

To ground treatment generation in real-world evidence, we constructed a rare disease treatment knowledge base from the Orphanet database and real-world case records. Only diseases for which Orphanet provided treatment-related information were eligible for inclusion; diseases without such information were excluded. For each eligible disease, we compiled a standardized entry comprising the disease name, its Orphacode, the disease category and clinical specialty, the general management and treatment information, and, where applicable, emergency-care guidance, together with derived question-answer pairs summarizing this content. This procedure yielded 1,021 diseases.

We then enriched the knowledge base with real cases drawn from RarelensBench. Within the list of included diseases, cases were matched to each disease, and a case was eligible if it both matched the disease and contained complete treatment information. For each disease, up to two eligible cases were randomly selected (one case if fewer than two were available), giving 371 diseases supported by one to two real cases (621 cases in total). Cases incorporated into the knowledge base were excluded from all subsequent evaluation to prevent data leakage. This knowledge base was subsequently used for retrieval-augmented generation by RareTreatment, and representative examples are available in our GitHub repository.

### The aligned-reasoning framework

RareLens converts the divergent reasoning of heterogeneous LLMs into a convergent, task-specific decision through a shared three-component framework: independent multi-model reasoning generation, quantitative characterization of inter-model divergence, and supervised reasoning alignment (with distillation for screening). The framework is instantiated per module with task-appropriate output schemas, divergence metrics, and alignment models, as detailed in Section 4.

### Multi-model reasoning generation

For each case, a panel of heterogeneous LLMs independently produced structured output appropriate to the task. Each model received identical clinical input for the relevant stage and processed each case independently, without access to the responses of other models. Outputs comprised a stage-specific prediction (a risk score, a ranked differential diagnosis, or a ranked treatment plan), the key clinical factors supporting the prediction (with relative importance weights where applicable), and an explanatory clinical rationale. To ensure comparability across models and across cases, all generations used fixed prompts and a standardized structured output schema. These independently generated outputs represent multiple alternative clinical interpretations of the same patient presentation. The exact prompts and code are shown in our Github repository. The exact panel composition and the number of models differed by module and are specified in Section 4.

### Characterisation of inter-model divergence

Divergence across models was quantified per task using metrics appropriate to the output type. For continuous risk scores (RareAlert), we compared predicted risk scores across models for each case, quantified the frequency of exact score agreement between model pairs, and calculated mean absolute pairwise differences in assigned risk estimates. For ranked diagnoses (RareDiagnosis), we computed the number of unique diagnoses generated across all models per case (breadth of the hypothesis space), pairwise Jaccard similarity between models' top-5 diagnosis sets (intersection divided by union, summarized across all model pairs and cases), the absolute difference in within-model rank position for diagnoses proposed by more than one model (rank disagreement), and top-1 disagreement, defined as the number of distinct top-ranked diagnoses generated across the panel for each case, with higher values indicating greater divergence in the most strongly prioritized diagnosis. For treatment plans (RareTreatment), diversity was characterised at the level of intervention content: we quantified, for each model, the distribution of treatment categories across all of its recommendations (modality distribution), capturing differences in the mix of intervention types selected, and, for the aligned output, the number of distinct models supporting each top-ranked recommendation. For categorical prognostic predictions (RarePrognosis), divergence was quantified as the pairwise agreement rate between models, defined as the proportion of cases for which two models predicted the same class, computed separately for each of the three ordinal dimensions (overall outcome, functional status and symptom burden), with lower agreement indicating greater divergence. The exact details of each evaluation metric are provided in Supplementary Information.

For RareDiagnosis, RareTreatment and RarePrognosis, the model outputs themselves represent distinct reasoning paths, as different diagnoses, treatment plans or prognostic judgements directly reflect different underlying reasoning. For RareAlert, however, the output is a single numeric risk score, so identical or differing scores do not by themselves reveal whether the underlying medical reasoning is the same. We therefore analysed the structure of clinical reasoning directly. From a stratified random subset of 10,000 cases drawn without replacement (7,500 non-rare and 2,500 rare disease cases), we collected, for each case, the risk scores, weighted key clinical factors and explanatory text from all models. GPT-5 served as the analysis model under fixed prompting and deterministic decoding (temperature 0); the analysis framework was developed iteratively on a separate subset of 1,000 cases, refining the prompt and output schema until model-generated analyses agreed with clinician-reviewed interpretations. For each case, the analysis model grouped the reasoning outputs into distinct reasoning chains and returned results in a standardized JSON format. Chains were considered distinct when they differed in diagnostic focus, causal logic, prioritisation of clinical features, diagnostic style, or the scope and certainty of conclusions; each chain was summarised by its core pathophysiological logic and interpretation of key features, emphasising reasoning structure rather than final diagnoses, and the number and identities of the models assigned to each chain were recorded. An example is shown in supplementary fig. 11.

Reasoning diversity was then quantified in three categories. Diversity was captured by the number of reasoning chains per case (within-case diversity) and the number of models per chain (shared interpretation across models). Structural similarity was assessed by the probability of co-occurrence (P_same), the probability that two models fell within the same chain. For each model, a consensus score was computed as the average number of models supporting the chains in which it participated.

**Reasoning alignment and distillation**

To convert divergent reasoning into a single decision, a supervised alignment step was trained on the development set using ground-truth labels, learning task-specific mappings from multi-model reasoning signals to calibrated predictions or ranked outputs rather than relying on simple aggregation. The specific alignment model differed by task: a score-integration classifier for screening, case-level learning-to-rank models for diagnosis and treatment, and a stacking ensemble for prognosis (see module sections). For screening, which should operate universally and at scale, the aligned reasoning was additionally distilled into a single locally deployable open-source model. All model selection, feature engineering, and hyperparameter tuning were performed within the development set using cross-validation, and held-out test sets were used only for final evaluation.

## Module development

### RareAlert

RareAlert is an early screening and alert module that addresses a fundamental clinical challenge in rare disease care: the inability to reliably recognize patients at high risk for rare diseases during primary clinical encounters, when only limited information is available. It operates exclusively on information available at primary visits, including demographic data, free-text medical history, and physical examination findings, without reliance on laboratory results, imaging studies, or specialized tests. RareAlert was developed to satisfy four core requirements: effective identification of high-risk patients using limited clinical information; comprehensive risk assessment across the full spectrum of rare diseases rather than a narrow subset; calibrated quantitative risk scores accompanied by clinically meaningful explanations; and suitability for large-scale global deployment, including in resource-limited settings, which necessitates efficient inference and local, privacy-preserving deployment.

RareAlert was developed through a two-stage supervised pipeline built upon a single, frozen data partition that was shared identically across both stages. The full RarelensBench dataset of 157,525 cases (42,836 rare; 114,689 non-rare) was allocated by stratified random sampling in an 8:2 ratio to a development set and a held-out test set, performed separately within the rare and non-rare groups so that disease prevalence was preserved in both partitions. Each case originated from a distinct patient, and no patient contributed cases to more than one partition, precluding patient-level leakage between the development and test sets. For each case, the information used comprised primary-visit clinical information (demographics, medical history, and physical examination findings), a binary rare/non-rare label, and, for rare disease cases, the specific rare disease category and associated medical specialty. In Stage 1, an ensemble of large language models (LLMs) independently scored each case and a supervised meta-learner fused these scores into a single calibrated reference risk score. In Stage 2, this reference risk score served as the supervision target for instruction fine-tuning of an open-source LLM, yielding RareAlert. To guard against data leakage, the test set was excluded from every component of model development: it was never used for training in both stages, and was evaluated only once, at the end of each stage, against the ground-truth labels.

In Stage 1, ten LLMs independently evaluated each case to capture variability in clinical interpretation under limited information: Claude-3.5-Haiku, DeepSeek-V3, Gemini-2.0-Flash, GPT-3.5-Turbo, GPT-4.1, GPT-4o-mini, LLaMA 3.1-8B, Qwen3-8B, Qwen3-14B, and Qwen3-32B. These models were selected to span diverse model families, parameter scales, and both proprietary (closed-source) and open-source systems. Each model received identical primary-visit information and, with temperature set to 0 and outputs constrained to a standardized structured schema, generated a structured output with three components: a quantitative clinical risk estimate on a 0-100 scale representing the likelihood of a rare disease; a set of key clinical factors, each assigned a relative importance weight normalized to sum to one; and a clinical explanation synthesizing the relationship between the identified factors and the estimated risk. The discriminative performance of each model's risk estimates was characterized using the area under the receiver operating characteristic curve (AUC), with an operating point defined by the Youden index (the threshold maximizing the sum of sensitivity and specificity); at this threshold, accuracy, sensitivity, specificity, false-negative rate, false-positive rate, positive predictive value, and negative predictive value were computed. Inter-model variation, the structural reasoning analysis on the 10,000-case subset, and the reasoning-diversity metrics described in Section 3.2 were all conducted within this module, with results summarized in Fig. 3, Supplementary Fig. 5, and 12. The ten model-derived risk scores were then treated as structured inputs to a supervised meta-learner, with the ground-truth disease status as the binary outcome, that learned a mapping from the set of scores to a single calibrated prediction. Candidate learning algorithms comprised logistic regression, random forest, extremely randomized trees, AdaBoost, gradient boosting methods (XGBoost and LightGBM), and CatBoost. Candidate feature-algorithm combinations were first screened by AUC under stratified five-fold cross-validation (Supplementary Fig. 13), and the top-ranked candidates were further optimized through nested cross-validation, with a five-fold outer loop for performance estimation and a three-fold inner loop for hyperparameter optimization. All screening, model selection, hyperparameter optimization, cross-validation, and determination of the Youden operating point were conducted exclusively within the development set; the selected configuration was then trained on the full development set and evaluated once on the held-out test set. The output of the final meta-learner, a probability between 0 and 1, was rescaled to a 0-100 range to yield the optimized reference risk score for every case.

In Stage 2, RareAlert was trained to map primary-visit clinical information directly to the optimized reference risk score while learning from the associated multi-model reasoning outputs. Each case was represented by both the set of independently generated reasoning outputs and its optimized reference risk score, and training was implemented through instruction fine-tuning of the open-source large language model Qwen3-32B. Parameter-efficient fine-tuning was performed using Low-Rank Adaptation (LoRA), with the rank set to 8. Optimization used the AdamW optimizer with a cosine learning-rate schedule and an initial learning rate of $1 \times 10^{-5}$. Training used sequence-level supervision, in which each input consisted of primary-visit clinical information and each output comprised both the reasoning content and the reference risk score. Fine-tuning was restricted to the development set, and RareAlert was evaluated a single time on the held-out test set against the ground-truth rare/non-rare labels, reporting AUC together with the metrics defined at the Youden operating point. We additionally performed an ablation analysis varying the number of reasoning paths included during RareAlert training and assessed its effect on test-set AUC (Supplementary Fig. 14a). After training, RareAlert directly processes primary-visit clinical information and outputs a calibrated rare disease risk score, identifies key contributing clinical features, and provides structured clinical explanations, and is designed for efficient local, privacy-preserving deployment.

**RareDiagnosis**

RareDiagnosis is a diagnostic decision-support module for patients suspected of having rare diseases.

It operates at two sequential stages. At the primary visit, available information is limited to demographic data, medical history, and physical examination findings, and the task is to generate a differential diagnosis and identify investigations necessary to establish a definitive diagnosis; RareDiagnosis outputs a ranked list of candidate diagnoses together with recommended confirmatory tests. At the follow-up visit, additional evidence (laboratory, biopsy, genetic, and imaging results) becomes available, and the task shifts to refinement and confirmation, with the ranking of candidate diagnoses updated on the integrated evidence. To emulate the way clinicians evaluate multiple diagnostic hypotheses in parallel, RareDiagnosis uses multiple LLMs as independent reasoning agents, preserves their heterogeneity to capture the full range of plausible diagnoses, and aligns these heterogeneous signals into a unified, clinically actionable diagnostic representation.

RareDiagnosis was developed on 42,836 RarelensBench rare disease cases with diagnosis-stage documentation. The dataset was randomly partitioned into training and test sets at an 8:2 ratio, with all development procedures performed on the training set only. To prevent leakage, data were partitioned at the case level before model development, keeping all records from the same case, including primary-visit and follow-up records, in the same partition.

In the generation stage, eleven LLMs (GPT-5, o3-mini, GPT-3.5-Turbo, DeepSeek-R1, Gemini-2.5-Flash, Claude-4.5-Haiku, Qwen3-235B, Qwen3-32B, Qwen3-14B, GPT-4o-mini and Qwen3-8B) independently evaluated each case under fixed prompts and standardized output requirements with temperature set to 0. These models were selected to span diverse model families, parameter scales, and both proprietary (closed-source) and open-source systems. At the primary visit, with input limited to demographics, medical history, and physical examination findings, each model generated (i) the five most likely diagnoses, each accompanied by brief diagnostic reasoning and a confidence score on a 0 to 10 scale, ranked from highest to lowest confidence, and (ii) the five most crucial diagnostic tests required to establish the diagnosis, each accompanied by a brief rationale and a necessity score on a 0 to 10 scale, ranked from highest to lowest necessity. At the follow-up visit, with input additionally including subsequent diagnostic test results, each model generated the five most likely diagnoses, each with brief diagnostic reasoning and a confidence score on a 0 to 10 scale, ranked from highest to lowest confidence. All generated diagnoses were mapped to Orphacode entries from the Orphanet database using the same retrieval-augmented matching procedure described for RarelensBench, unifying semantically equivalent diagnoses across models and enabling consistent quantification of inter-model heterogeneity (number of unique diagnoses per case, pairwise Jaccard similarity of top-5 sets, rank disagreement for shared diagnoses, and top-1 disagreement, as defined in Section 3.2).

In the alignment stage, candidate diagnoses generated by multiple models were consolidated into a unified diagnostic space using Orphacode normalization, with diagnoses mapped to the same Orphacode merged into a single candidate. For each candidate, multi-source attributes were retained (contributing models, within-model ranks, confidence scores, and associated reasoning), from which aggregated features were constructed (cross-model support counts, summary statistics of confidence scores, and rank-based features), preserving both consensus and variability. A learning-based alignment step estimated the relative priority of candidate diagnoses within each case, trained using ground-truth diagnoses as supervision with the objective of assigning higher scores to correct diagnoses than to competing candidates; all candidates within a case were jointly considered and ranking was constrained at the case level. Diagnostic signals from the primary and follow-up visits were jointly incorporated within a unified ranking space, enabling diagnoses supported by follow-up findings to be preferentially ranked while retaining relevant primary-stage hypotheses, thereby capturing the longitudinal diagnostic trajectory. At the primary visit, diagnostic test recommendations were derived from source-model outputs for the top-K reranked diagnoses and were taken from the diagnosis with the broadest model support.

The diagnostic alignment model was developed by iterative optimisation. We first compared gradient-boosting ranking models (LightGBM, CatBoost and XGBoost) under pointwise, pairwise and listwise formulations; the listwise formulation resulted in the most stable performance across top-k metrics, and the final model used XGBoost with a listwise objective (rank:ndcg) (Supplementary Fig. 15a,b). All development steps were performed within a GroupKFold cross-validation scheme grouped by case identifier, so that every candidate diagnosis from a given case remained in a single fold and could not leak across folds. Within this scheme, features derived from the multi-model reasoning signals were organised into semantically related groups and assessed by group-wise ablation, with non-contributory groups removed (Supplementary Fig. 15c-e); different subsets of contributing LLMs were compared and stable configurations retained (Supplementary Fig. 16a-d); and hyperparameters were tuned by Bayesian optimisation. After alignment, distributions of alignment-model scores, case-level separation between ground-truth and competing candidates, and source-model support for top-ranked candidates were analysed to characterize the learned diagnostic ranking space (Supplementary Fig. 17). The selected model was then retrained on the full development set and evaluated once on the independent test set.

**RareTreatment**

RareTreatment is a treatment decision-support module for use after a rare disease diagnosis has been established. It operates at the treatment decision stage and takes as input the confirmed diagnosis together with relevant clinical context available at that time, without access to downstream outcome information, preserving temporal consistency with real-world care. Given a case, RareTreatment outputs a strictly ordered list of candidate treatment plans, each represented in a structured format including intervention details, implementation considerations, monitoring strategies and a concise clinical rationale.

RareTreatment was developed on a curated subset of RarelensBench. Building on the cases that satisfied the inclusion criteria described above, we further required each case to document a complete treatment intervention, operationalised as the presence of at least one valid, non-placeholder entry in each of four core fields: treatment type, specific treatment, treatment response, and final patient status. Entries consisting of empty strings, template placeholders, or "Not specified" (in any capitalisation) were treated as missing. After quality control, 2,131 cases were retained. These cases were split at an 8:2 ratio, with one partition used to construct the treatment knowledge base and the remainder used to develop and evaluate RareTreatment. To prevent leakage, data were partitioned at the case level, keeping all records, LLM outputs, features, and labels from a given case within the same partition. For each case, model inputs were restricted to information available at the treatment decision point, including the diagnosis and relevant patient context; outcome data were used only for curation and evaluation and were never provided to the model.

To capture heterogeneity in clinical reasoning, 12 LLMs independently generated candidate treatment plans for each case: Claude-Haiku-4.5, DeepSeek-R1, DeepSeek-V3.2-Exp, Gemini-2.5-Flash, GPT-3.5-Turbo, GPT-4o-mini, GPT-5, o3-mini, Qwen3-8B, Qwen3-14B, Qwen3-32B and Qwen3-235B. These models were selected to span diverse model families, parameter scales, and both proprietary (closed-source) and open-source systems. To ground generation in curated evidence, the confirmed diagnosis was used to retrieve the corresponding disease entry from the rare disease treatment knowledge base (Section 2.5), and the retrieved general management and, where available, emergency-care information was supplied to every model as additional context (retrieval-augmented generation). Conditioned on the case information and this retrieved evidence, each model produced a single structured treatment plan, comprising intervention content, implementation details, monitoring strategies, safety considerations and a clinical rationale, with temperature set to 0.

To construct the training targets, we assessed each candidate plan on whether it had been implemented in the documented clinical record and on three quality dimensions: completeness (whether essential implementation elements were present), helpfulness (expected clinical benefit in context) and safety (risk profile given patient-specific constraints). These were combined into a single binary judgment of appropriateness indicating whether the plan was clinically suitable for the case. The evaluation procedure is described in Section 5.2 (Metrics). Detailed distributions of source-model treatment recommendations, quality scores and confidence-quality associations are provided in Supplementary Figs. 18 and 19.

RareTreatment learned a case-level re-ranker that assigns each candidate treatment option a scalar priority score and orders treatment options within each case. Candidate plans were represented using multi-source attributes retained from the generation stage, including model provenance, within-model ranks, model-reported importance and associated clinical rationale. Aggregated features captured cross-model support, rank-based evidence, semantic fit to the case context, rationale-text characteristics, and variability in model support and candidate rankings.

The treatment re-ranker was developed through iterative optimisation of gradient-boosting ranking models, including LightGBM, CatBoost and XGBoost, under pointwise, pairwise and listwise formulations (Supplementary Fig. 20a). The final model used XGBoost with a listwise NDCG ranking objective. Model development was performed within a GroupKFold cross-validation scheme grouped by case identifier, ensuring that all candidate treatments from the same case remained within a single fold. Within this scheme, feature groups derived from multi-model treatment reasoning signals were assessed by group-wise ablation (Supplementary Fig. 20b,c), different subsets of contributing LLMs were compared and stable configurations retained (Supplementary Fig. 20d,e), and hyperparameters were tuned within the development set. We further examined post-alignment treatment priority scores, including score assignment patterns, label-stratified score distributions, and within-case separation between appropriate and inappropriate options (Supplementary Fig. 21).

Supervision was anchored by the binary appropriateness label from the structured treatment evaluation pipeline. At inference, candidates were generated using the same multi-model evidence-grounded procedure, pooled, featurized and ordered by the trained re-ranker. For ranking evaluation, any-hit@k was defined as a case-level success if at least one appropriate treatment appeared among the top-k recommendations; all-hit@k, or strict@k, required all top-k recommendations to be appropriate.

**RarePrognosis**

RarePrognosis assesses prognosis for rare disease patients who have started treatment and entered follow-up. Using only prognosis-stage inputs (medical history, physical examination, diagnosis and treatment received), it estimates the likely post-treatment trajectory. Given the multidimensional nature of rare disease prognosis, the task is framed as case-level multi-task classification, predicting patient outcomes across three targets: Overall outcome captures the global course of the condition and comprises five categories: complete recovery (full resolution with return to pre-illness baseline and no residual symptoms or limitations), partial recovery (substantial improvement with persistent symptoms, functional limitations or sequelae despite appropriate treatment), stabilization (a plateau in which the condition neither meaningfully improves nor deteriorates under current management), progression (ongoing deterioration despite therapeutic intervention, with worsening symptoms or advancing disease burden) and terminal outcome (an irreversible condition expected to result in death). Functional status reflects the patient's ability to perform daily or age-appropriate activities and is graded as none (no limitation), mild (minor limitation with activities largely preserved), moderate (clear limitation affecting normal activities) or severe (major impairment or dependence). Symptom burden reflects the persistence and impact of symptoms and is graded as none, occasional, persistent mild or persistent severe.

RarePrognosis was developed on a curated subset of RarelensBench. Building on the cases that satisfied the inclusion criteria described above, we further required each case to document complete prognosis information, defined as valid content for three structured fields: the overall outcome and the two quality-of-life dimensions of functional status and symptom burden. Cases in which any of these fields was absent or fell outside its permitted vocabulary were excluded. Applying this filter yielded 18,668 RarelensBench cases with adequate prognosis-stage documentation. The dataset was randomly partitioned into training and test sets at an 8:2 ratio, with all development procedures performed on the training set only. To prevent leakage, data were partitioned at the case level, keeping all records, LLM outputs, features, and labels from a case in the same partition.

To obtain diverse prognosis evaluations across model families, 12 LLMs independently evaluated each case: Claude-Haiku-4.5, DeepSeek-R1, DeepSeek-V3.2-Exp, Gemini-2.5-Flash, GPT-3.5-Turbo, GPT-4o-mini, GPT-5, o3-mini, Qwen3-8B, Qwen3-14B, Qwen3-32B, and Qwen3-235B. These models were selected to represent diverse model families, parameter scales, and both proprietary (closed-source) and open-source systems. All models received the same structured clinical input from the prognosis stage and generated outputs without access to the responses of other models. For each case, each model produced a structured prognosis output with three task-level predictions: overall outcome, functional status, and symptom burden. Each prediction included a category, a confidence score on a 0 to 10 scale, and a brief clinical explanation. Models also generated prognosis-related clinical event signals, including symptom improvement, functional improvement, deterioration, readmission, re-intervention, major complication, and death, with the expected timing categorized as short-term (<3 months), mid-term (3 months to 1 year), long-term (>1 year), or unknown. All generations used fixed prompts, standardized output requirements, and temperature set to 0.

The LLM-derived outputs were represented as ensemble features for supervised aggregation. For each task, categorical predictions from the LLMs were encoded as base-model features, and model explanations were summarized using text-derived features capturing explanation length, negation, numeric content, and prognosis-related terms. These features were standardized using development-set statistics and combined into a multi-model feature representation. Training targets were obtained from upstream RarelensBench structured prognosis annotations, aligned by case identifier and used as supervision for the three task-specific classifiers. Prognosis annotations were extracted from documented clinical outcomes. Task-specific label normalization was applied to merge equivalent labels, and cases without valid split assignment or usable reference labels were excluded.

Model development compared multiple ensemble strategies, including majority voting, weighted voting, calibrated confidence aggregation, dynamic ensemble selection, linear stacking, and tree-based ensemble models (Supplementary Fig. 22). Feature groups derived from multi-model prognosis signals were assessed by group-wise ablation (Supplementary Fig. 23a,b), and different subsets of contributing LLMs were compared to identify stable configurations (Supplementary Fig. 23c,d). The final aligned model was selected according to cross-task performance and stability, and used gradient-boosted decision-tree stacking implemented as a GradientBoostingClassifier for all three prognosis tasks. For each task, a separate classifier was trained using stratified five-fold cross-validation within the development set. After model selection, the selected model was retrained on the full development set for each task and evaluated once on the independent test set.

**Evaluation**

**Comparators**

At each stage, RareLens was compared with individual LLMs evaluated independently under the same settings and using the same case inputs, including both the models used during development and additional frontier models. For RareAlert, comparisons were conducted with the development LLMs and with additional models Claude-3.7-Sonnet, DeepSeek-R1, Gemini-2.5-Pro, GPT-5, o3-mini, and Qwen3-235B; RareAlert was additionally compared with the machine-learning integration models developed during alignment. For RareDiagnosis, comparator models were Claude-Haiku-4.5, DeepSeek-R1, Gemini-2.5-Flash, GPT-3.5-Turbo, GPT-4o-mini, GPT-5, o3-mini, Qwen3-8B, Qwen3-14B, Qwen3-32B, and Qwen3-235B. For RareTreatment and RarePrognosis, the comparator models were GPT-5, o3-mini, Gemini-2.5-Flash, Claude-Haiku-4.5, Qwen3-235B, GPT-3.5-Turbo, DeepSeek-R1, DeepSeek-V3.2-exp, Qwen3-14B, Qwen3-32B, Qwen3-8B, and GPT-4o-mini.

For every module, we additionally compared RareLens with a simple aggregation baseline that combined the panel's outputs without learned alignment. For RareAlert, whose outputs are continuous risk scores, the model-generated scores were aggregated by their mean, median and mode, with the aggregated value treated as the final prediction. For RareDiagnosis, RareTreatment and RarePrognosis, whose outputs are discrete, the most frequently produced output across models (majority vote) was taken as the final prediction.

**Metrics**

For RareAlert, screening discrimination was summarised by the area under the receiver operating characteristic curve (AUC), which measures the ability to separate rare from non-rare cases across all decision thresholds. At the threshold defined by the Youden index, we computed sensitivity (the proportion of true rare disease cases correctly flagged as high risk), specificity (the proportion of non-rare cases correctly classified as low risk), positive predictive value (the proportion of high-risk predictions that were true rare disease cases) and negative predictive value (the corresponding proportion for low-risk predictions), together with accuracy (the proportion of all cases classified correctly), the false-negative and false-positive rates, balanced accuracy (the mean of sensitivity and specificity), the F1 and F2 scores (the harmonic mean of precision and sensitivity, with F2 weighting sensitivity more heavily) and the Matthews correlation coefficient (a correlation between predicted and true labels that accounts for all four cells of the confusion matrix).

For RareDiagnosis and RareTreatment, ranking quality was summarised by top-k accuracy (Acc@1, Acc@3 and Acc@5; whether a correct answer appeared within the top 1, 3 or 5 ranked candidates), the mean reciprocal rank (MRR; the reciprocal of the rank at which the first correct answer appeared, averaged over cases, so that earlier correct answers score higher) and the normalised discounted cumulative gain (NDCG@1, NDCG@3 and NDCG@5; a measure of overall ranking quality that gives greater weight to correct answers placed nearer the top). For RareTreatment, where several treatments may be appropriate, Acc@k was additionally reported under an any-hit criterion (at least one appropriate treatment within the top k) and an all-hit criterion (all top-k treatments appropriate), and we reported the mean number of appropriate and inappropriate treatments per case, the proportion of high-risk recommendations (treatments with a low safety rating), and clinician-relevant ratings of completeness, helpfulness and safety. We further analysed RareDiagnosis by specialty (computing Acc@1, Acc@3, Acc@5 and MRR within each specialty using the labels assigned during RarelensBench curation). Longitudinal refinement was assessed within paired cases by the change in accuracy between stages (delta-Acc@k, follow-up minus primary), the recovery rate (Recovery@k, the proportion of cases incorrect at the primary visit that became correct at follow-up) and the degradation rate (Degradation@k, the proportion correct at the primary visit that became incorrect).

Diagnostic and treatment outputs were scored by model-based adjudication (an LLM-as-judge approach), in which a judge model was provided with the reference standard and the model-generated output and determined whether the output was correct or clinically appropriate. All adjudication used fixed prompts and deterministic decoding (temperature 0).

For RareDiagnosis, GPT-5 nano served as the judge. Each candidate diagnosis was rated on a five-point scale adapted from Bond et al (32, 33). (5, exact diagnosis; 4, closely related; 3, partially relevant; 2, weakly related; 0, no relevant suggestion), and only a score of 5 was counted as correct for the calculation of accuracy and ranking metrics. Recommended diagnostic tests at the primary visit were rated on a five-point Likert scale of clinical usefulness, with tests scoring above 3 considered useful.

For RareTreatment, GPT-5 served as the judge, drawing on three sources of clinical evidence for each case-candidate pair: the patient's clinical information, disease-specific treatment knowledge from curated resources, and the real-world treatments the patient actually received. Each candidate plan was assessed on whether it had been implemented in the documented clinical record and on three quality dimensions, completeness (whether essential implementation elements were present), helpfulness (expected clinical benefit in context) and safety (risk profile given patient-specific constraints), which were combined into a single binary judgment of appropriateness indicating whether the plan was clinically suitable for the case.

The adjudication pipeline was first developed on 500 randomly sampled cases, on which the judge model, prompts and scoring procedure were determined through iterative refinement. To further confirm reliability, an additional 500 cases were randomly sampled for each evaluation task (diagnosis and treatment) and assessed independently by the judge model and by human reviewers. The LLM-based evaluation pipeline showed almost perfect agreement with expert humans in diagnosis assessment(agreement rate 97.6%; Cohen's $\kappa = 0.95$) and substantial agreement (agreement rate 86.0%; Cohen's $\kappa = 0.63$) in treatment assessment.

For RarePrognosis, performance was summarised by the accuracy and Matthews correlation coefficient of each of the three ordinal dimensions, a joint criterion requiring all three dimensions to be correct for the same patient, the severe-outcome recall (the proportion of truly high-risk outcomes correctly identified), the false-optimism rate (the proportion of high-risk outcomes misclassified as less severe, equal to one minus the severe-outcome recall) and the ordinal error (the mean absolute difference between the predicted and true ordinal levels).

The per-model rescue rate (Rescue Rate@k) was the proportion of cases failed by a given comparator at top-k that RareLens nonetheless answered correctly, and the multi-model rescue rate (Multi-model Rescue Rate@k(n)) was the proportion answered correctly among cases failed by at least n comparator models.

Exact definitions and formulas for all metrics are provided in the supplementary table 1.

**External human-AI evaluation**

To examine the clinical potential of RareLens and how it might be used in practice, we conducted a human–AI study comparing three groups: RareLens alone, physicians alone, and physicians working together with RareLens(collaboration group). The study used an independent test set of 1,287 cases (371 rare disease and 916 non-rare disease cases) that had not been used in any training or development and that met the same quality-control criteria. Unlike our earlier evaluations, in which each stage was tested separately, here each case was assessed through the entire clinical pathway in sequence, advancing to the next stage only if the previous stage was completed correctly. This end-to-end design more closely reflects how decisions actually unfold in clinical care.

For the RareLens group, RareLens processed each case directly. For the Physicians group and Collaboration group, we used a crossed design so that every case was evaluated under both conditions, each time by a different physician; this allowed the two conditions to be compared on the same cases while ensuring that no physician saw the same case twice. A total of 23 practicing physicians from 10 hospitals, all of whom had completed standardized residency training, participated in the study. They all worked under both conditions. The number of physicians was determined by the available workforce and the overall workload; because most cases were non-rare and therefore did not proceed to diagnosis, treatment or prognosis, the per-physician workload was reduced.

In the RareLens group and Physicians group, the agent or the physician completed each case independently, without any other input. In the Collaboration group, the physician first reviewed the patient information and formed an initial impression, could then choose to view the corresponding RareLens output, and finally submitted a decision that served as the group's output. An online demonstration of the evaluation system is available at https://eval.rarelens.org/.

Every case was processed in sequence from the primary visit through the full pathway, with each stage receiving only the information that would actually be available at that decision point. A case advanced to the next stage only if it met the current stage's criterion; once it failed a stage, or did not reach it, the case was counted as failed at that and all subsequent stages. Cumulative success at every stage was therefore measured against the full set of cases, not only against those that happened to reach that stage, allowing the three groups to be compared on the same basis. For a rare disease case, the criteria were: flagged as high risk at the alert stage; correct diagnosis within the top 5 candidates at the primary visit; correct diagnosis ranked first at follow-up; an appropriate treatment within the top 5 at the treatment stage; and at least one of the three prognostic dimensions correct at the prognosis stage. A rare disease case was only counted as an end-to-end success if it met all five criteria in order. For a non-rare case, success was defined as being correctly identified as low risk at the alert stage.

We report three outcomes: the overall end-to-end success rate, cumulative accuracy along the care pathway, and the sensitivity and specificity in screening rare disease patients. For the Collaboration Group, we additionally report rescue and regression rates relative to physicians working unaided.

**Comparison with other ensemble techniques**

To test whether the supervised RareLens aggregators could be replaced by training-free combination of the constituent model outputs, each module was additionally evaluated against a panel of ensemble baselines that use no learned parameters, with the ensemble family matched to the output type of the module and every baseline scored on the same cases and metrics as RareLens. For the candidate-ranking modules (primary and follow-up diagnosis, and treatment), the candidate set for each case was the union of the model nominations, with every candidate carrying a confidence score (0–10; 0 when not nominated), a rank (999 when not nominated ) and a binary nomination flag; we evaluated score fusion (CombSUM, mean, CombMNZ and CombANZ), rank fusion (reciprocal rank fusion with K = 60, Borda, and a length-normalized Borda that removes the bias toward models nominating longer lists), voting (hit-sum) and central tendency (median and mode over the nominating models). Missing entries were set to zero for score fusion and voting and skipped for central tendency. For rank-fusion methods, scores were accumulated only over nominated items, so non-nominated items contributed zero to the rank-fusion score. Candidates were then ranked by the resulting composite score and evaluated at top-k. For the ordinal prognosis module (functional status, symptom burden and overall outcome), each of the 12 models returned a single ordinal category and a scalar confidence (0–10), which precluded strict soft voting; we therefore compared mode, confidence-weighted vote, ordinal mean, ordinal median and confidence-weighted median (the prespecified ordinal baseline, which uses both order and confidence without assuming equal spacing between grades), breaking voting ties by summed confidence rather than by defaulting to the more severe grade so as not to bias the severity-sensitive metrics. For the binary risk-screening module, each of the 16 models returned a continuous risk score (0–100) and a binary prediction, and we compared majority voting (the positive-vote fraction at a threshold of 0.5), the mean and median of the per-model min–max normalized scores, and the mean, median and mode of the raw scores as scale-sensitivity controls. All ensemble outputs were evaluated with the same metrics as the corresponding RareLens method, and full per-method results are reported in the Supplementary Information.

**5.5 Reliability analysis**

To assess the stability of RareLens performance, each module was run three times on a random subset of 500 cases, and agreement across the three runs was quantified. Continuous scores were assessed with the absolute-agreement intraclass correlation coefficient (ICC[2,1]). Categorical outputs were assessed with Cohen's kappa (mean of the three pairwise values) and Fleiss' kappa (three raters); ordinal prognostic outputs were assessed with linearly weighted Cohen's kappa; and Gwet's AC1 was additionally reported for categorical and hit@k agreement analyses as a prevalence-robust complement to kappa, which can be deflated when agreement is high and class distributions are skewed. For diagnosis and treatment, agreement was computed on the hit@k indicator of whether a correct diagnosis or appropriate treatment appeared within the top k. Binary screening agreement used the Youden-index operating threshold. Full per-field statistics are reported in the Supplementary Information.

### Efficiency and Cost

API cost was estimated from token usage across 371 test cases that completed the full RareLens workflow (diagnosis, prognosis, treatment). For each metered call, cost was derived from prompt and completion tokens using provider-listed prices at analysis time, then summed per case. RareAlert was excluded, as it ran on a self-hosted Qwen3-32B endpoint rather than a metered API, incurring local compute but no provider fee.

Runtime was benchmarked on two boundary cases spanning the lowest and highest observed token counts, each run three times through the sequential workflow (risk screening, primary diagnosis, follow-up diagnosis, prognosis, treatment). Runtime was summarized overall and by stage as approximate mean wall-clock time with ranges. Because ensemble calls within a stage ran concurrently, stage time reflected the slowest model response plus synthesis and provider latency, not the sum of individual latencies.

### Statistical Analysis

Continuous variables are summarised as mean and standard deviation, and categorical variables as counts and percentages. Results are reported descriptively without statistical hypothesis testing. GGiven the large held-out test sets, in which even negligible differences reach nominal statistical significance, we report results descriptively and emphasise the magnitude and clinical relevance of performance differences rather than p-values (34).

### Ethics

This study used published literature and publicly available databases and did not involve any intervention on patients; ethical approval was therefore not required. The study did not involve any human genetic samples.

### Data and Code Availability

A subset of RarelensBench comprising 500 randomly selected cases is released to illustrate the dataset structure and to support reproducibility, and is available in our GitHub repository (https://github.com/geteff1/RareLens). The full dataset cannot be released without restriction because it incorporates data governed by third-party agreements. The MIMIC-IV-Ext data are subject to controlled access and can be obtained from PhysioNet (https://physionet.org/) after registration, completion of the required training, and acceptance of the data use agreement. The published case reports are publicly available and can be retrieved using the procedure described in this paper (Section 2.2). The code for RareLens, including the multi-model reasoning generation, and the alignment and distillation pipelines for all modules is available at https://github.com/geteff1/RareLens, which also contains a demonstration video. An interactive demonstration of the system is available on the project website (https://rarelens.org/).

# Main Figure Legends

**Fig. 1 | The RareLens framework.**

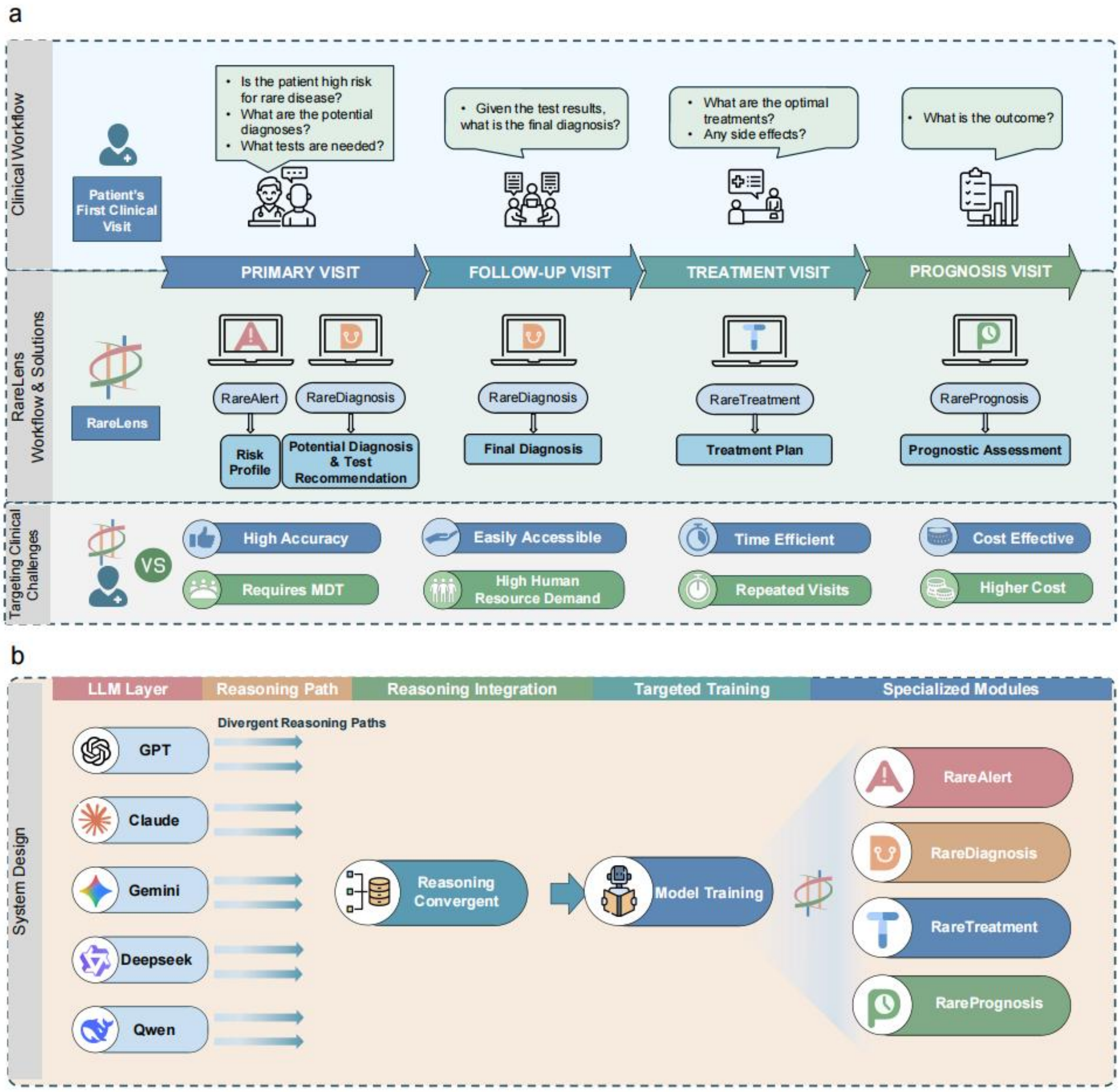

**a,** Schematic of the clinical trajectory from initial presentation through primary and follow-up visits to treatment and prognosis. RareLens supports sequential clinical decisions through four specialized modules: RareAlert estimates rare-disease risk; RareDiagnosis generates candidate diagnoses and test recommendations before determining the final diagnosis; RareTreatment prioritizes treatment options; and RarePrognosis provides prognostic assessment. The lower section summarizes the intended advantages of the RareLens workflow over conventional care.

**b,** Shared technical architecture underlying the four modules. A heterogeneous panel of large language models (LLMs) generates parallel reasoning paths for each case. These outputs are aligned, aggregated and calibrated by a case-level machine-learning integration layer and subsequently used for targeted training of locally deployable, task-specific RareAlert, RareDiagnosis, RareTreatment and RarePrognosis models.

**Fig. 2 | Construction of RarelensBench.**

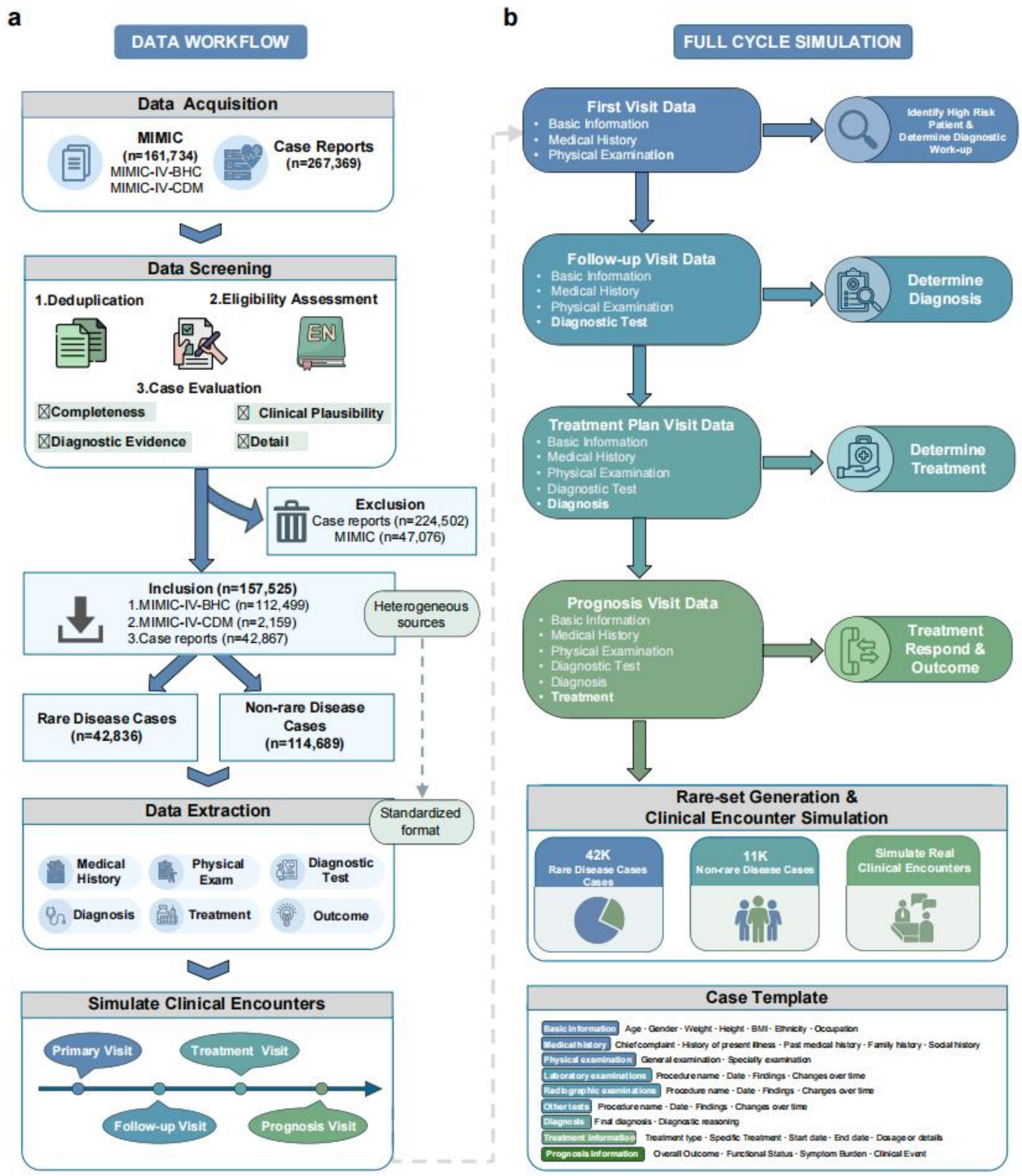

**a,** Data acquisition, screening and standardized extraction workflow. Records from the MIMIC-IV-Ext-BHC and MIMIC-IV-Ext-CDM databases and published case reports underwent deduplication, eligibility assessment and quality review. Eligible cases were classified as rare or non-rare disease and standardized into six clinical fields: medical history, physical examination, diagnostic tests, diagnosis, treatment and outcome. Source-specific inclusion and exclusion counts are shown. The resulting cases were organized into longitudinal clinical encounters comprising primary, follow-up, treatment and prognosis visits.

**b,** Full-cycle simulation of longitudinal clinical encounters. Clinical information was released progressively across the four visits to support diagnostic-work-up planning, diagnosis, treatment selection and prognostic assessment, respectively. Information revealed at later stages was withheld from earlier stages to prevent information leakage.

**Figure 3 | RareAlert workflow, alerting performance and rescue analysis.**

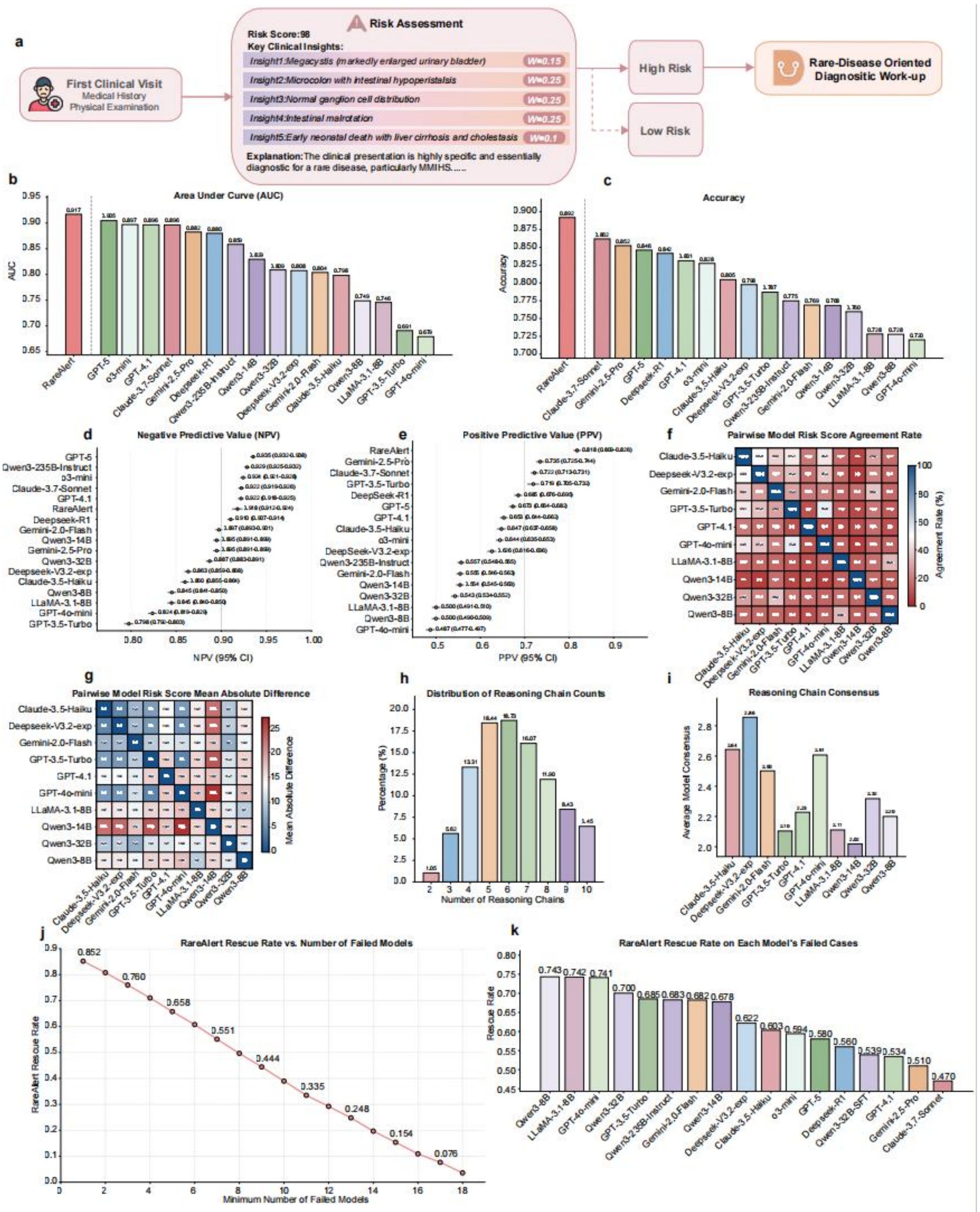

**a,** RareAlert workflow, showing how patient-level clinical information is integrated to generate rare-disease alert scores and reasoning outputs.

**b,** Area under the receiver operating characteristic curve (AUC) for RareAlert and baseline LLMs.

**c,** Overall accuracy, defined as the proportion of cases correctly classified as rare-disease-positive or negative.

**d,** Negative predictive value (NPV), the proportion of predicted-negative cases that are true negatives; points show estimates and horizontal lines show confidence intervals.

**e,** Positive predictive value (PPV), the proportion of predicted-positive cases that are true rare-disease-positive cases.

**f,** Pairwise agreement heat map; rows and columns denote models and cells show the proportion of cases assigned the same binary alert label.

**g,** Mean absolute difference (MAD) heat map; cells show the average absolute difference in alert scores between model pairs for the same cases.

**h,** Distribution of reasoning-chain counts or classes in rare-disease alerting.

**i,** Reasoning-chain consensus across models.

**j,** Rescue rate as a function of the number of failed baseline models, defined as the proportion of missed cases correctly recovered by RareAlert.

**k,** Model-specific rescue rate among failed baseline cases.

**Figure 4 | RareDiagnosis workflow and diagnostic ranking performance.**

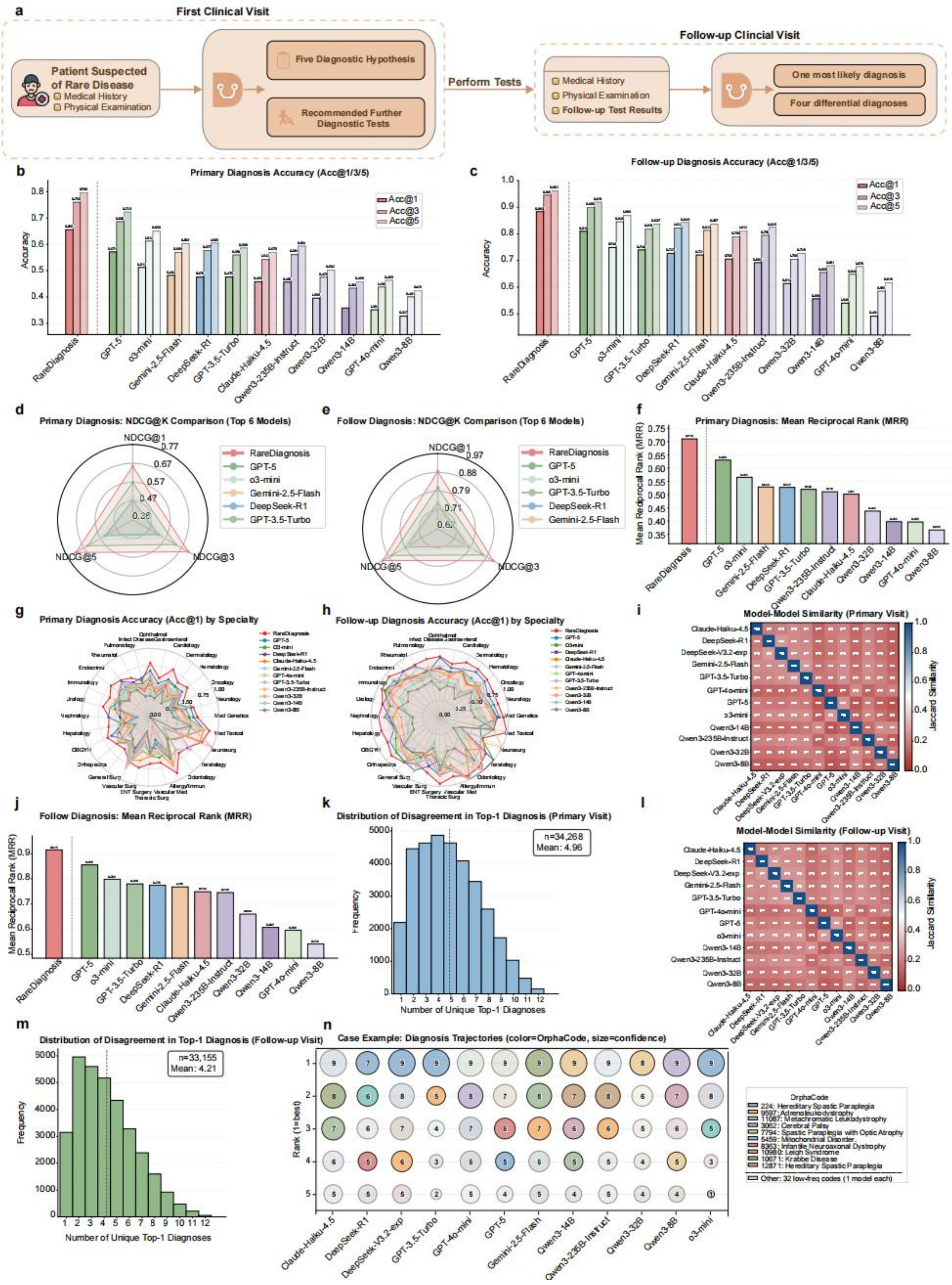

**a,** RareDiagnosis workflow from initial presentation to follow-up. Initial patient information is used to generate five preliminary disease hypotheses and recommend additional examinations; the resulting clinical findings are then incorporated to update the ranked candidates and identify the most likely follow-up diagnosis.

**b,** Primary-diagnosis Acc@K for RareDiagnosis and baseline LLMs. Acc@K denotes the proportion of cases in which the reference diagnosis appears within the top K ranked candidates (K = 1, 3 and 5). The divider separates RareDiagnosis from single-model baselines.

**c,** Follow-up Acc@K after incorporation of additional clinical information, using the same top-k definition as in b.

**d,** Primary-diagnosis NDCG@K for RareDiagnosis and baseline LLMs (K = 1, 3 and 5). NDCG is a rank-sensitive measure that gives higher credit when the reference diagnosis is placed closer to the top of the ranked candidate list.

**e,** Follow-up NDCG@K after incorporation of additional clinical information (K = 1, 3 and 5), using the same rank-sensitive definition as in d.

**f,** Primary-diagnosis mean reciprocal rank (MRR), defined as the reciprocal rank of the reference diagnosis averaged across cases.

**g,** Primary-diagnosis Acc@1 stratified by clinical specialty, evaluating performance when only the leading candidate is considered.

**h,** Follow-up Acc@1 stratified by clinical specialty after incorporation of additional clinical information.

**i,** Pairwise model-similarity heat map for primary diagnosis. Rows and columns denote models, and the colour scale indicates pairwise Jaccard similarity between candidate diagnosis sets.

**j,** Follow-up MRR, defined analogously to f for the updated ranked diagnosis list.

**k,** Distribution of top-1 disagreement in primary diagnosis. The x axis shows the number of distinct first-ranked diagnoses produced across models for each case.

**l,** Pairwise model-similarity heat map for follow-up diagnosis, using the same model ordering and Jaccard similarity scale as in i.

**m,** Distribution of top-1 disagreement after follow-up information is incorporated, summarizing residual diagnostic heterogeneity across models.

**n,** Representative case example illustrating diagnostic trajectories through the RareDiagnosis workflow.

**Figure 5 | RareTreatment workflow and treatment recommendation performance.**

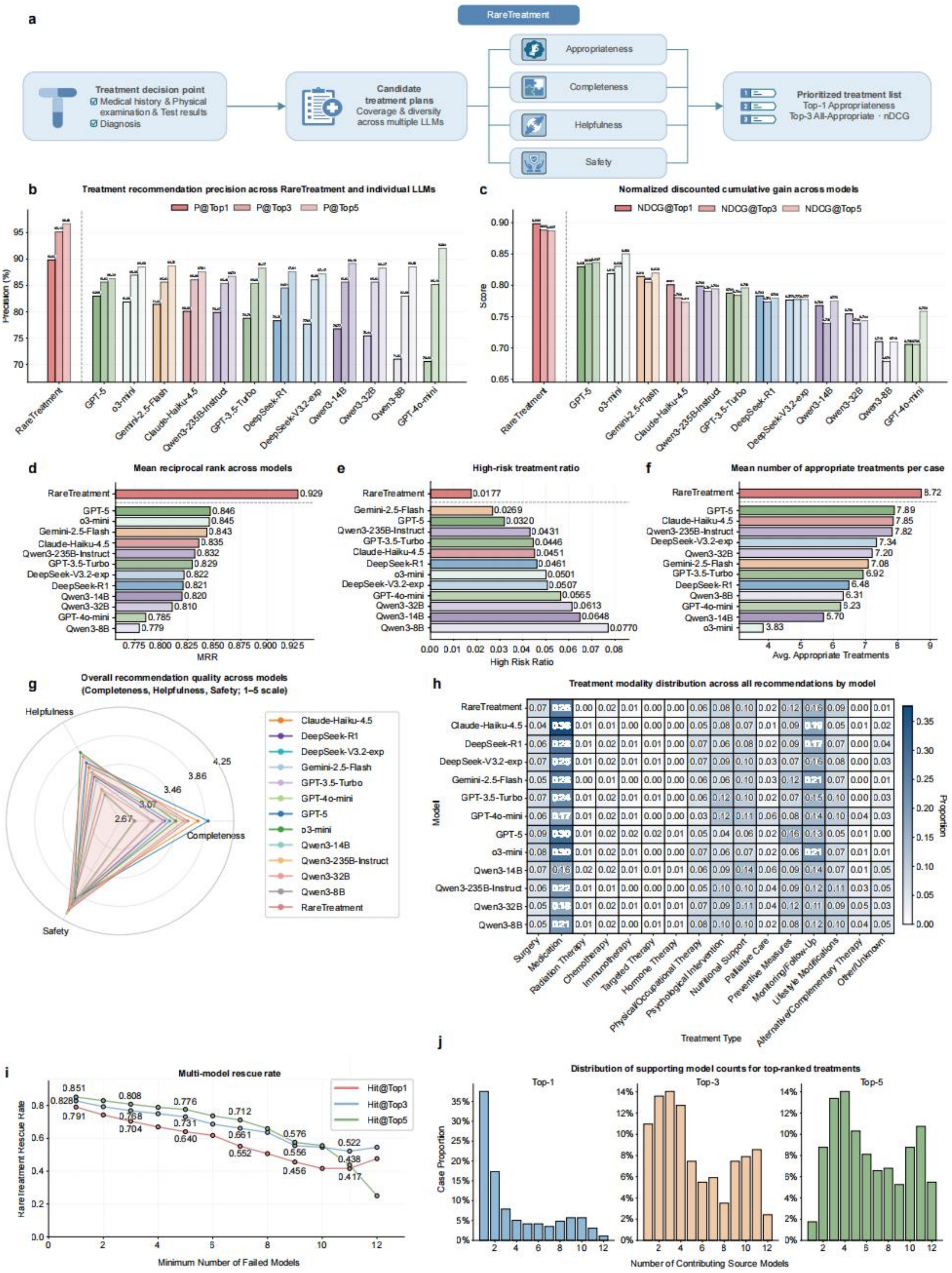

**a,** RareTreatment workflow from treatment decision point to prioritized treatment recommendation. Patient information is used to assemble candidate treatment plans with broad coverage and diversity; RareTreatment then generate a prioritized treatment list.

**b,** Top-k treatment recommendation hit rate across n = 452 rare-disease test cases. A case is counted as a hit if at least one appropriate treatment appears among the top k ranked recommendations (k = 1, 3 and 5; denoted P@T1, P@T3 and P@T5).

**c,** Normalized discounted cumulative gain (NDCG@K; K = 1, 3 and 5), a rank-sensitive measure that gives higher credit when appropriate treatments are placed closer to the top of the ranked list.

**d,** Mean reciprocal rank (MRR), defined as the reciprocal rank of the first appropriate treatment averaged across cases.

e, High-risk treatment ratio, defined as the proportion of scored recommendations with a safety score of 2 or lower on a 1-5 scale; unrated recommendations are excluded from the denominator.

**f,** Mean number of appropriate treatments per case, where appropriate denotes clinical suitability for the patient and concordance with reference knowledge.

**g,** Overall treatment-recommendation quality across models. Radar plots show the mean completeness, helpfulness and safety scores of the generated recommendations on a 1–5 scale. Each line represents one model, and higher scores indicate better recommendation quality.

**h,** Treatment-modality distribution heat map across all recommendations. Rows denote models, columns denote treatment categories and colour intensity indicates the proportion of recommendations in each category.

**i,** Multi-model rescue analysis. Rescue rate is defined as the proportion of cases in which RareTreatment recovers at least one appropriate treatment among its top k recommendations among cases where at least t baseline LLMs fail at the same top-k threshold.

**j,** Distribution of supporting-model counts for top-ranked RareTreatment recommendations at Top-1, Top-3 and Top-5, showing how often prioritized recommendations are supported by multiple baseline models.

**Figure 6 | RarePrognosis workflow and prognostic prediction performance.**

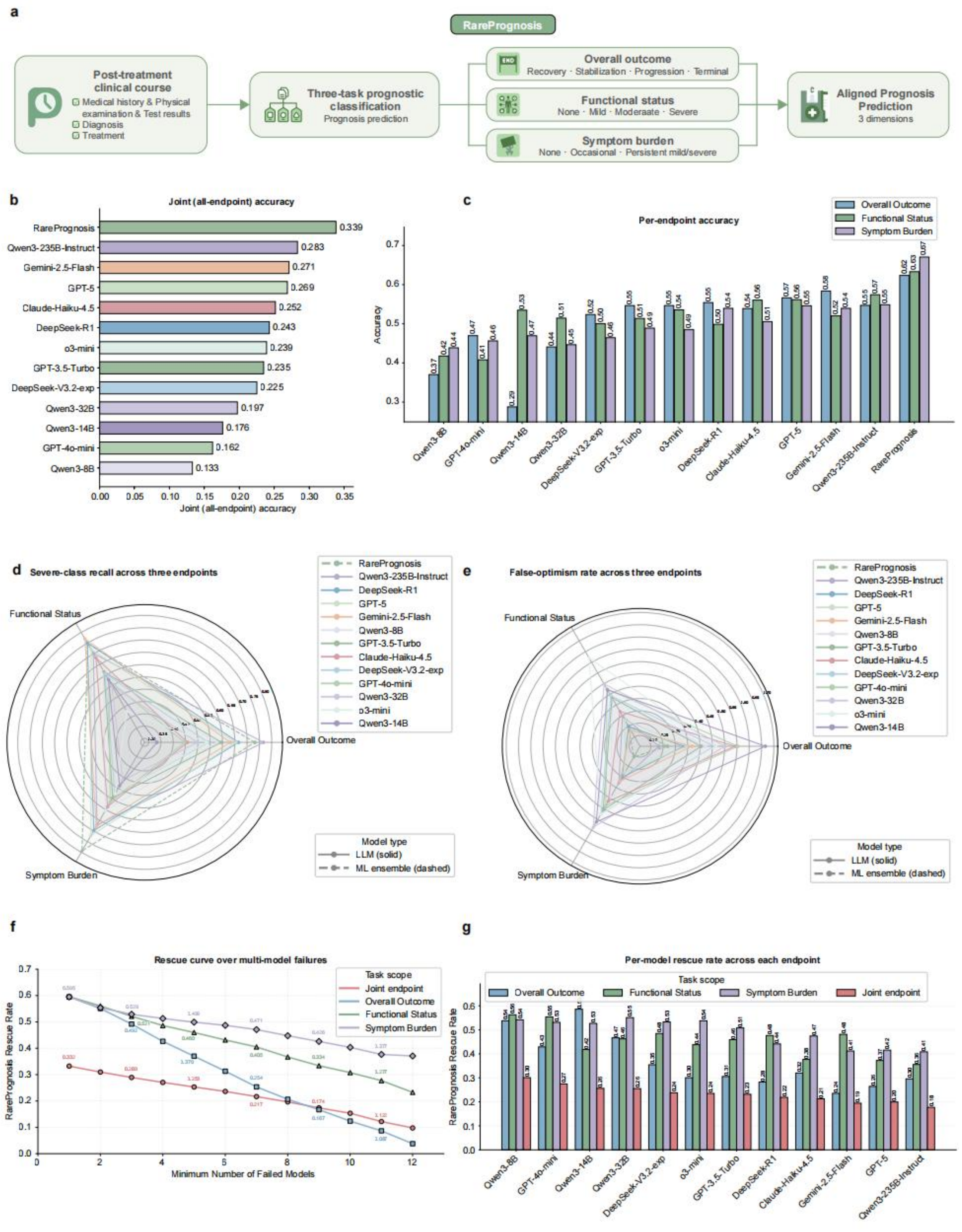

**a,** RarePrognosis workflow from post-treatment clinical course to aligned prognosis prediction. Medical history, physical examination findings, test results, diagnosis and treatment information are used to perform three-task prognostic classification across overall outcome, functional status and symptom burden.

**b,** All-three-correct accuracy across independent rare-disease test cases. A case is counted as correct only when overall outcome, functional status and symptom burden are all predicted correctly for the same patient.

**c,** Per-task exact-match accuracy for each prognostic endpoint. Accuracy is defined as the proportion of cases in which the predicted category matches the reference category for overall outcome, functional status or symptom burden.

**d,** Severe-class recall across the three prognostic endpoints. High-risk classes are progression or terminal for overall outcome, severe for functional status and persistent severe for symptom burden. Larger values indicate better recognition of high-risk cases; solid and dashed lines denote baseline LLMs and ML ensembles, respectively.

**e,** False-optimism rate across the three prognostic endpoints. False optimism is defined as the proportion of truly high-risk cases predicted as non-high-risk categories, equivalent to 1 minus severe-class recall. Lower values indicate fewer missed high-risk cases; solid and dashed lines denote baseline LLMs and ML ensembles, respectively.

**f,** Rescue curve over multi-model failures. For each task scope, rescue rate is defined as the proportion of cases correctly predicted by RarePrognosis among cases missed by at least t baseline LLMs, plotted as a function of the minimum number of failed models.

**g,** Per-model rescue rate across prognostic endpoints. For each baseline LLM and task scope, rescue rate is defined as the proportion of cases missed by that baseline model that are correctly predicted by RarePrognosis; bars show rescue rates for overall outcome, functional status, symptom burden and the joint endpoint.

**Figure 7 | External comparison of RareLens, physicians and RareLens-assisted physicians.**

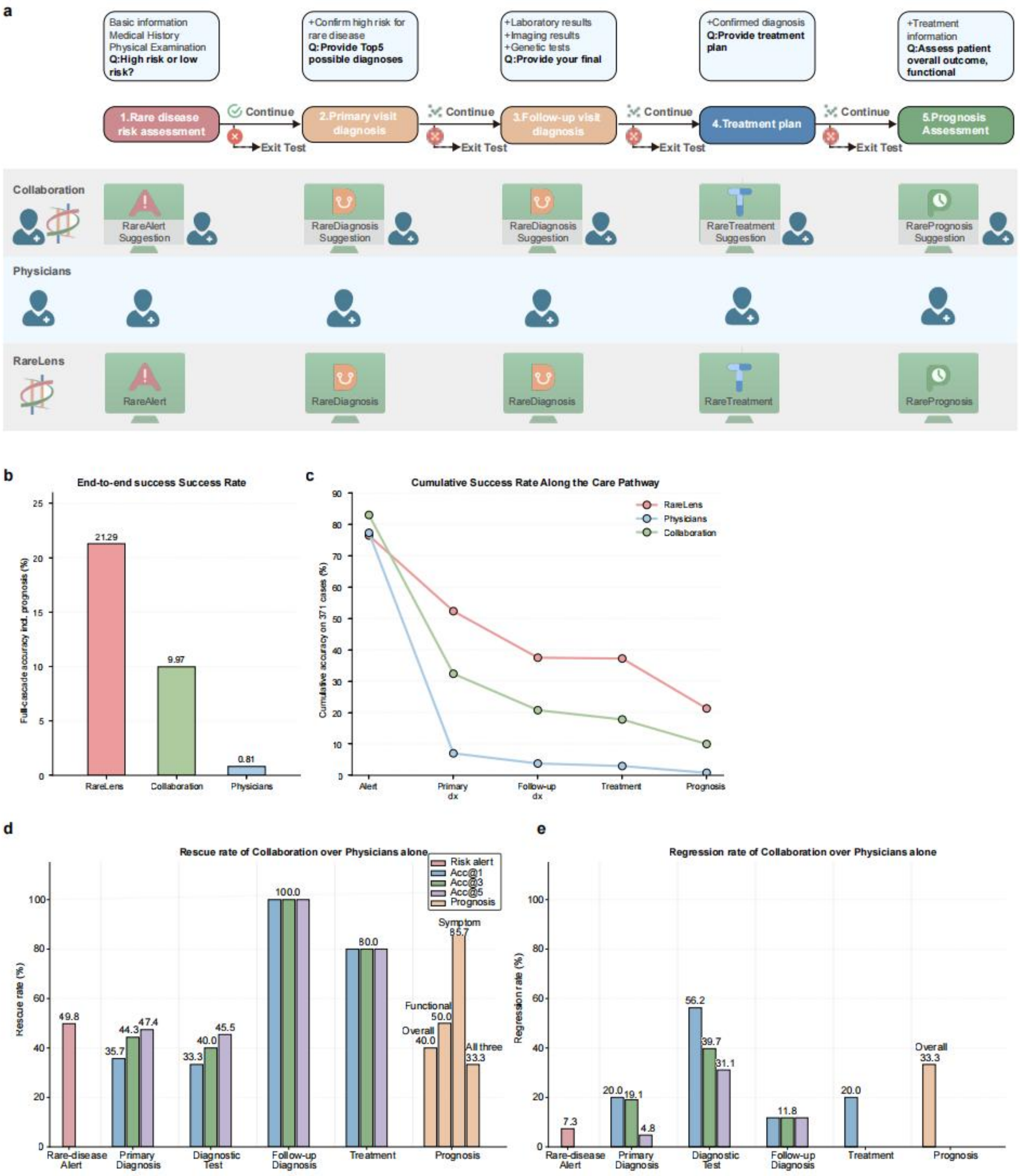

**a,** Evaluation workflow for RareLens alone, physicians unaided and physicians assisted by RareLens on an independent cohort of 371 rare-disease and 916 non-rare-disease cases. In the assisted condition, physicians could consult the RareLens output before submitting their final decisions. Cases advanced through alert, primary diagnosis, follow-up diagnosis, treatment and prognosis only after passing the preceding stage; failure terminated the cascade. Non-rare-disease cases were evaluated at the alert stage only.

**b,** End-to-end success rates among the rare-disease cases. Success required correct risk identification, a top-five primary diagnosis, a top-one follow-up diagnosis, an appropriate treatment within the top five recommendations and at least one correct prognostic assessment.

**c,** Cumulative success rates using a fixed denominator of rare-disease cases. Cases failing or not reaching a stage were counted as unsuccessful at that and all subsequent stages.

**d,** Case-level rescue and regression rates with RareLens assistance. Rescue denotes an incorrect unaided result that was correct under assistance; regression denotes a correct unaided result that was incorrect under assistance.

# Supplementary Information

*Supplementary Notes, Figures and Tables*

This document contains 1 Supplementary Note, 23 Supplementary Figures and 3 Supplementary Tables.

# Contents

# Supplementary Figures

## Supplementary Figure 1 | Candidate-set diversity and confidence heterogeneity across source LLMs in RareDiagnosis.

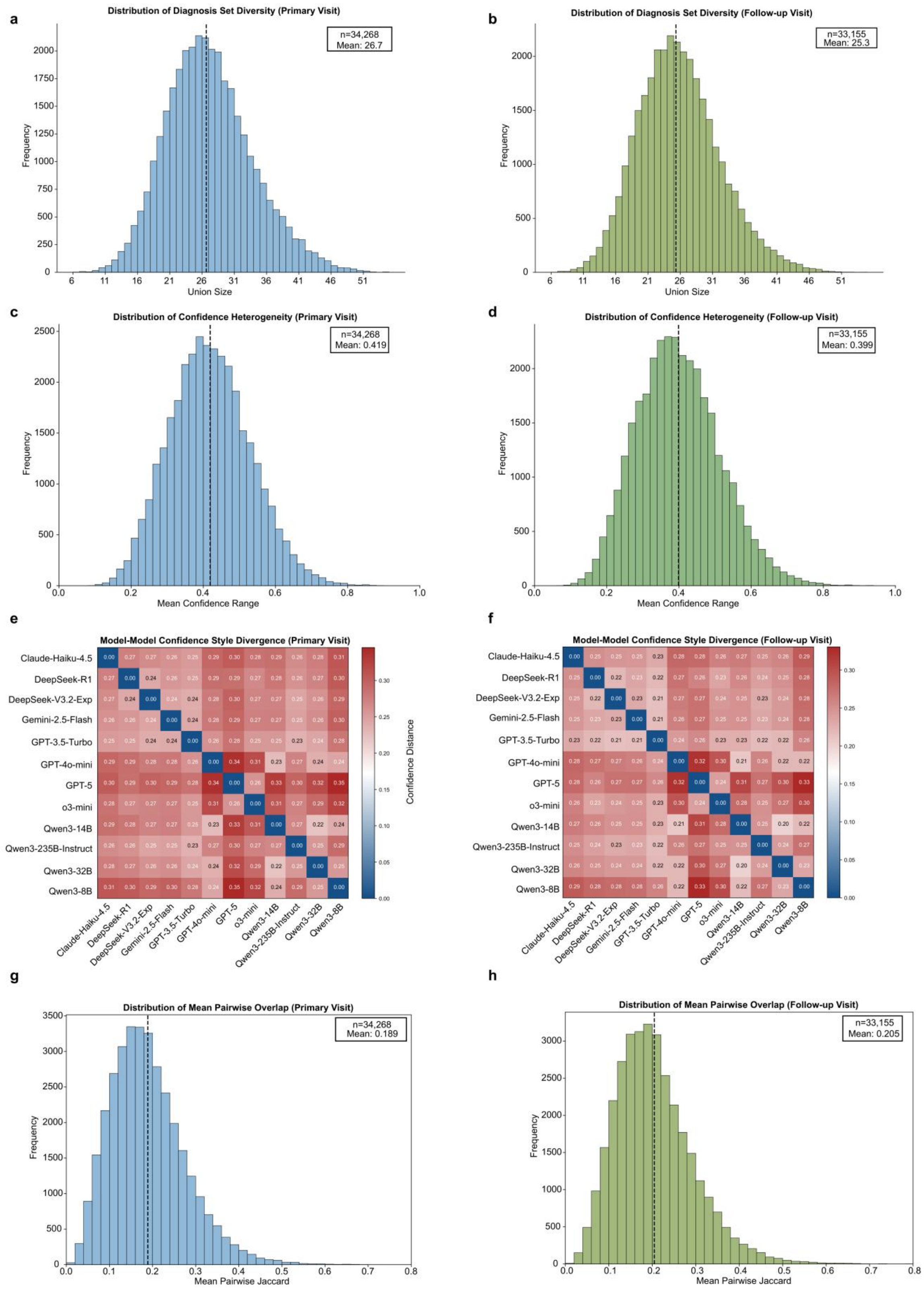

Union size was defined as the number of distinct diagnoses across the source-LLM candidate lists for each case. Confidence heterogeneity summarizes variation in normalized confidence scores for diagnoses shared by multiple models, with larger values indicating greater inter-model variation. Pairwise candidate-set overlap was quantified using the Jaccard index and averaged across model pairs within each case.

a, Candidate-set union size at the primary visit.

b, Candidate-set union size at the follow-up visit.

c, Mean confidence range at the primary visit.

d, Mean confidence range at the follow-up visit. In a–d, dashed lines indicate means, and annotations report sample sizes and mean values.

e, Pairwise confidence divergence between source LLMs at the follow-up visit.

f, Pairwise confidence divergence between source LLMs at the primary visit. In e and f, redder cells indicate greater confidence divergence and bluer cells indicate lower divergence; annotations show pairwise values.

g, Mean pairwise Jaccard overlap among source-LLM candidate sets at the primary visit.

h, Mean pairwise Jaccard overlap among source-LLM candidate sets at the follow-up visit. In g and h, dashed lines indicate means, and annotations report sample sizes and mean values.

**Supplementary Figure 2 | RareDiagnosis consensus strength and ranking heterogeneity.**

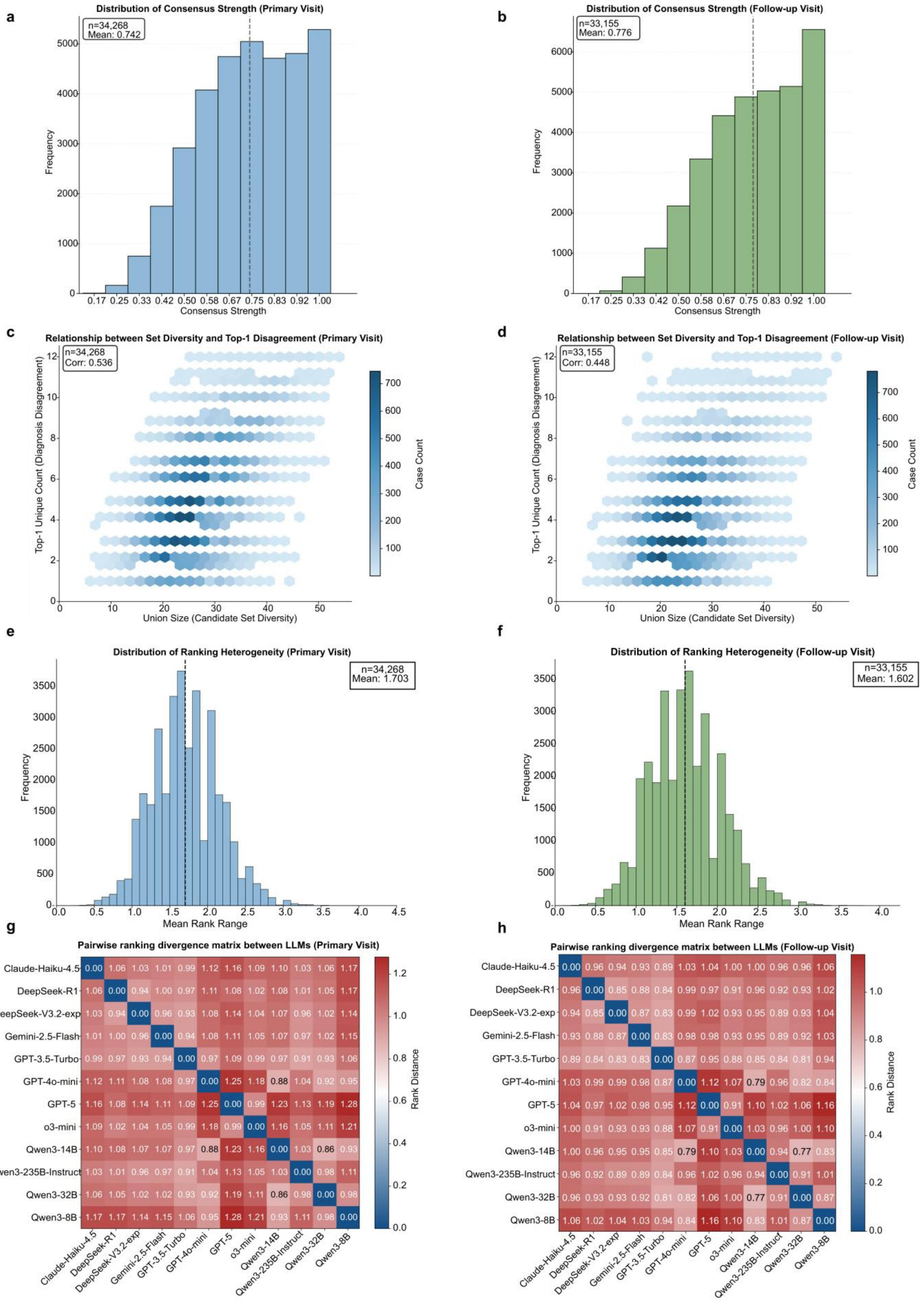

Consensus strength was defined for each case as the proportion of source large language models (LLMs) that included the most frequently predicted diagnosis in their candidate lists. Candidate-set union size was the number of distinct diagnoses across all source-LLM candidate lists, and top-1 unique count was the number of distinct diagnoses ranked first by the source LLMs. For each diagnosis predicted by at least two LLMs, rank range was calculated as the highest rank minus the lowest rank assigned across LLMs.

a, Distribution of consensus strength at the primary visit.

b, Distribution of consensus strength at the follow-up visit. In a and b, dashed lines indicate means, and annotations report sample sizes and mean values.

c, Relationship between candidate-set union size and top-1 unique count at the primary visit.

d, Relationship between candidate-set union size and top-1 unique count at the follow-up visit. In c and d, colour intensity indicates the number of cases in each hexagonal bin, and annotations report sample sizes and correlation coefficients.

e, Distribution of mean rank range at the primary visit.

f, Distribution of mean rank range at the follow-up visit. For each case, rank ranges were calculated for diagnoses predicted by at least two source LLMs and then averaged across these shared diagnoses. Larger values indicate less consistent ranking. In e and f, dashed lines indicate means, and annotations report sample sizes and mean values.

g, Pairwise ranking divergence between source LLMs at the primary visit.

h, Pairwise ranking divergence between source LLMs at the follow-up visit. For each model pair, ranking divergence was calculated as the mean absolute rank difference for diagnoses predicted by both models within each case and then averaged across cases. Redder cells indicate greater ranking divergence and bluer cells indicate lower divergence; annotations show the corresponding pairwise values.

**Supplementary Figure 3 | Treatment modality composition across RareTreatment and LLMs' recommendations.**

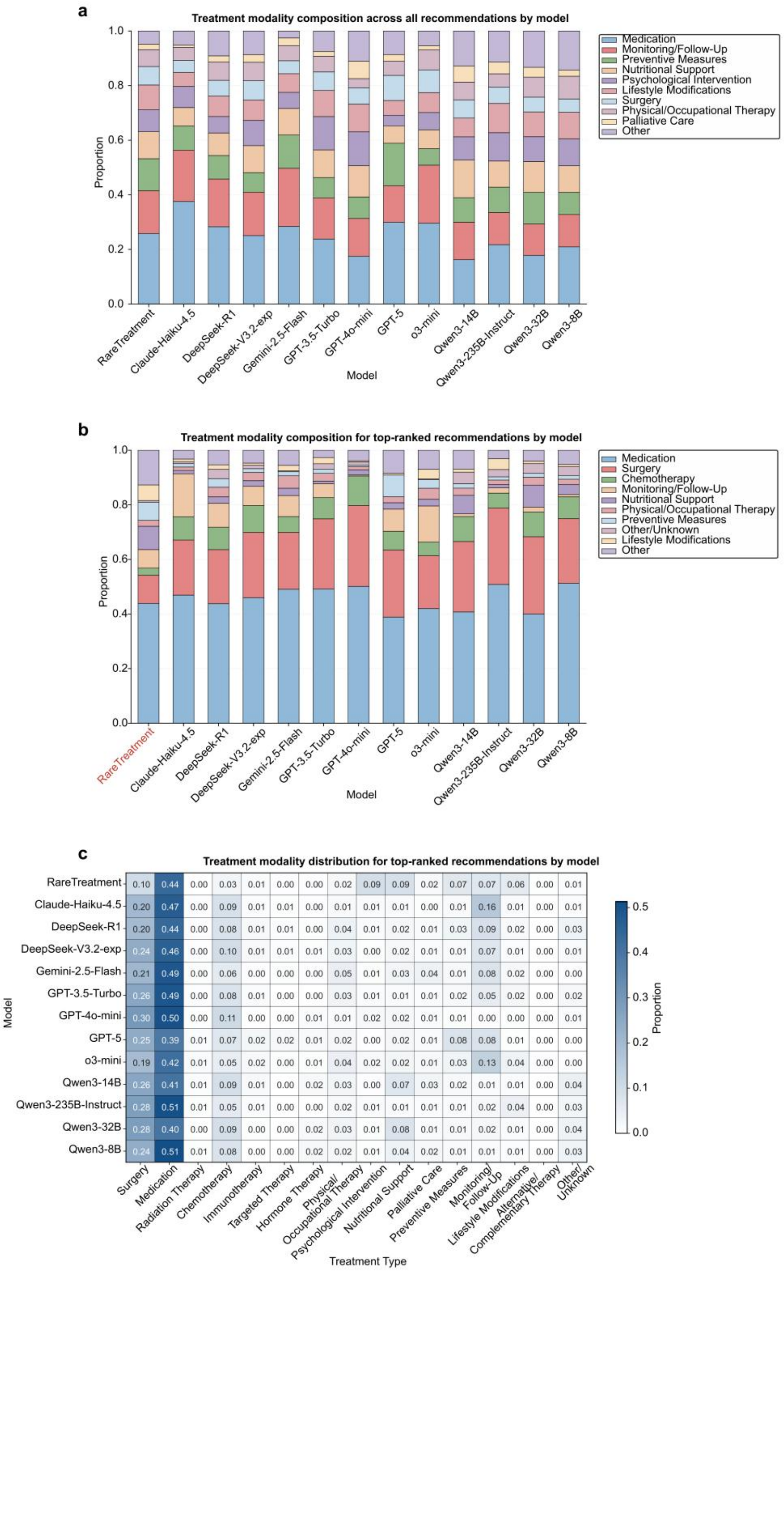

Proportions were calculated separately within each model. Top-ranked recommendations refer to the first-ranked treatment for each case.

a, Composition of all treatment recommendations by treatment category. The nine most frequent categories across all recommendations and models are shown separately; the remaining categories are grouped as ‘Other’.

b, Composition of top-ranked treatment recommendations by treatment category. The nine most frequent categories across top-ranked recommendations and models; the remaining categories are grouped as ‘Other’.

c, Distribution of detailed treatment types among top-ranked recommendations. Rows represent models and columns represent treatment types. Cell values and colour intensity indicate the within-model proportion assigned to each treatment type (range, 0–1), with darker shading indicating a higher proportion.

**Supplementary Figure 4 | Prediction heterogeneity among LLMs in RarePrognosis.**

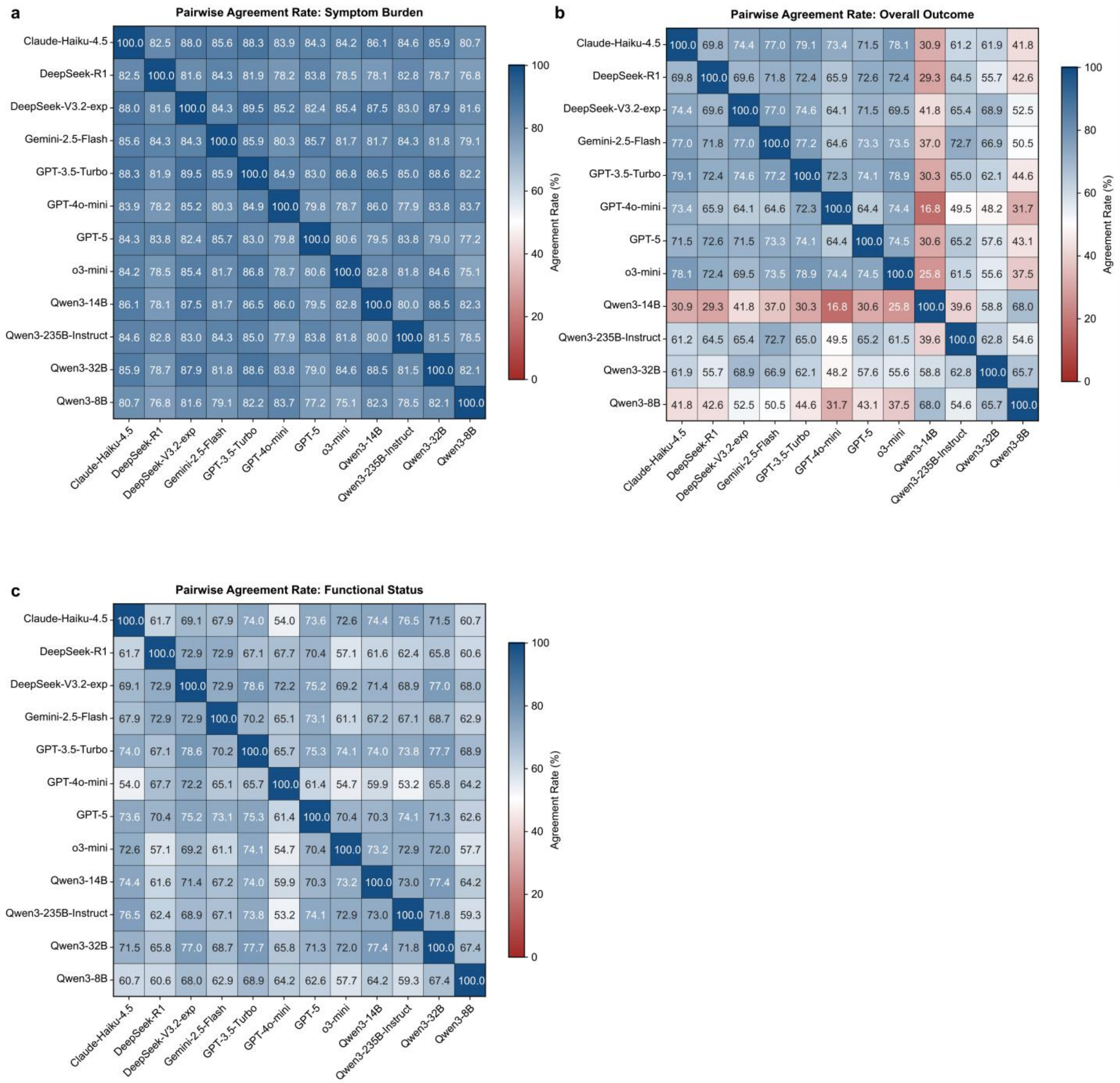

Pairwise agreement was used to quantify heterogeneity among the selected LLMs integrated within RarePrognosis and was defined as the percentage of cases for which two models predicted the same categorical prognosis label. Rows and columns represent LLMs. Cell values and colour intensity indicate agreement rates (0–100%), with redder cells indicating lower agreement and greater prediction heterogeneity, and bluer cells indicating higher agreement. Diagonal values denote 100% self-agreement.

a, Pairwise agreement for symptom-burden predictions.

b, Pairwise agreement for overall-outcome predictions.

c, Pairwise agreement for functional-status predictions.

**Supplementary Figure 5 | RareAlert performance across medical specialties.**

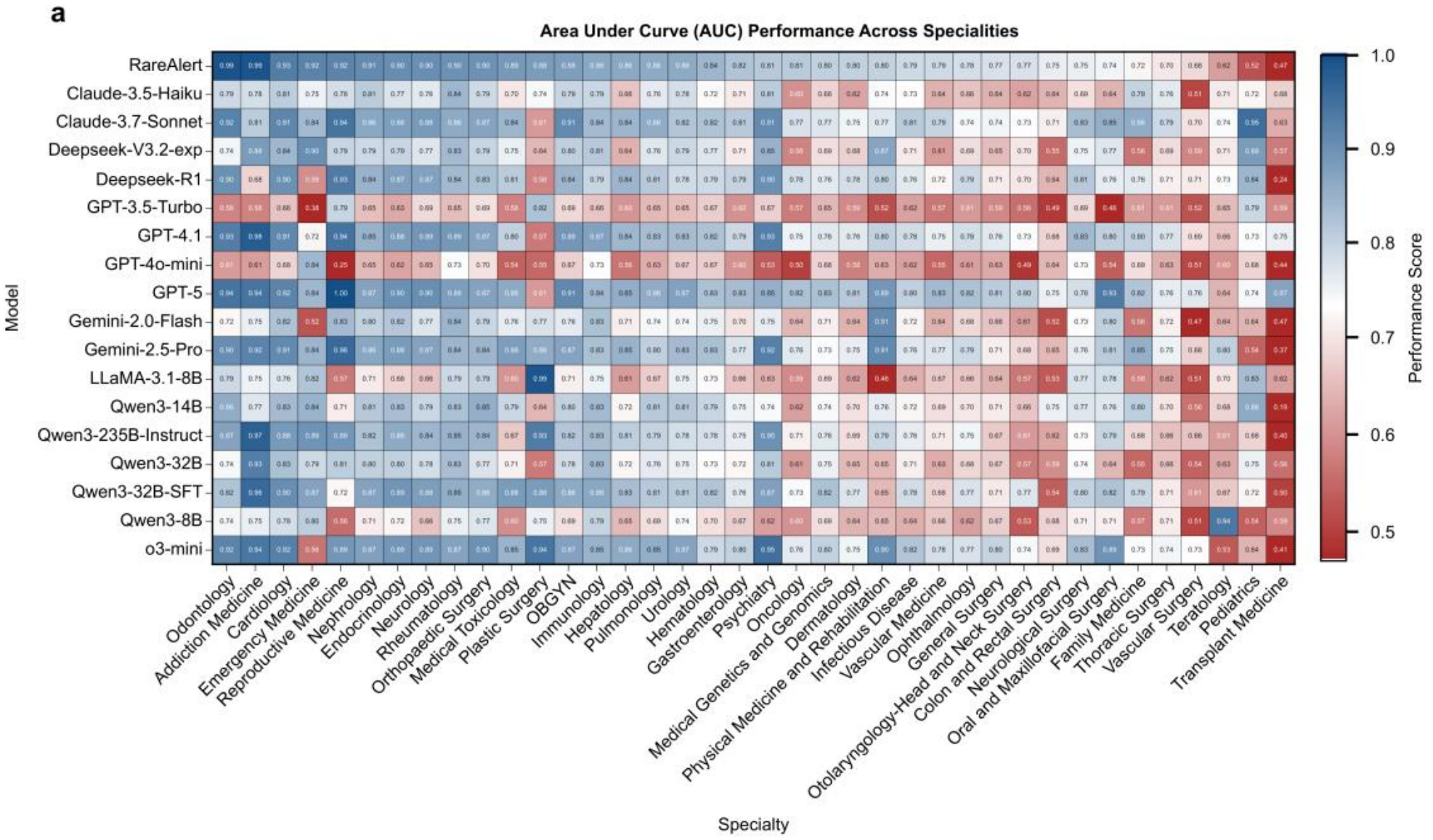

The area under the receiver operating characteristic curve (AUC) was calculated separately within each clinical speciality subgroup. The heatmap shows speciality-specific AUC values for RareAlert and individual large language models (LLMs). Bluer cells indicate higher AUC values and redder cells indicate lower AUC values; annotations show the corresponding AUC for each model–speciality pair

**Supplementary Figure 6 |Performance comparison of RareAlert, individual large language models for early rare disease risk screening.**

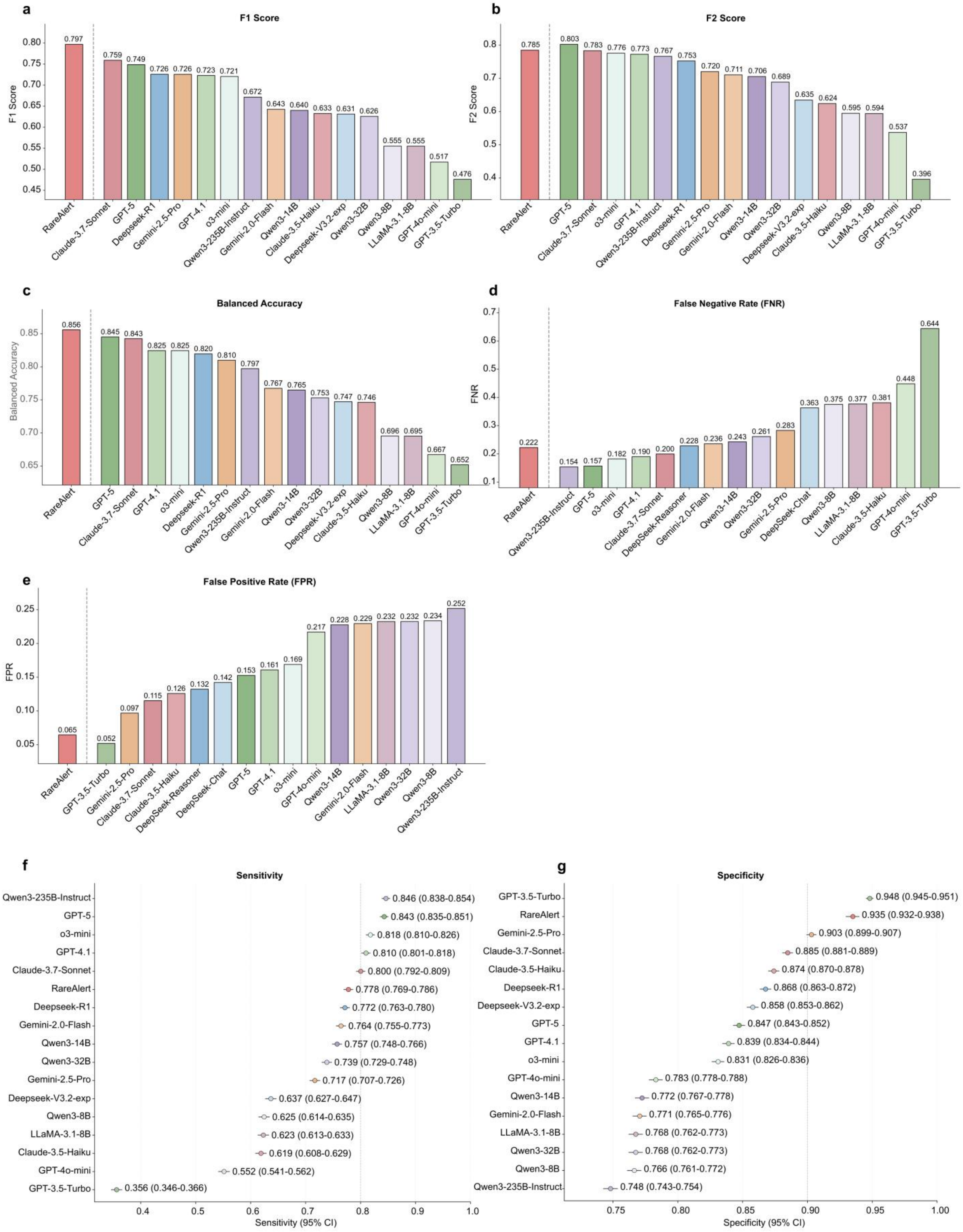

F1 and F2 scores summarize the balance between precision and recall, with the F2 score assigning greater weight to sensitivity. Balanced accuracy is the mean of sensitivity and specificity. The false-negative rate (FNR) is the proportion of rare-disease-positive cases classified as negative, and the false-positive rate (FPR) is the proportion of negative cases classified as positive.

a, F1 score for RareAlert and the individual large language models (LLMs).

b, F2 score.

c, Balanced accuracy.

d, FNR at the selected operating threshold.

e, FPR at the selected operating threshold.

f, Sensitivity was calculated at the operating threshold defined by Youden's index, with 95% confidence intervals (CIs) estimated using the Wald (normal-approximation) method.

g, Specificity was calculated at the operating threshold defined by Youden's index, with 95% confidence intervals (CIs) estimated using the Wald (normal-approximation) method.

**Supplementary Figure 7 | Changes in diagnostic accuracy between primary and follow-up visits and rescue by RareDiagnosis.**

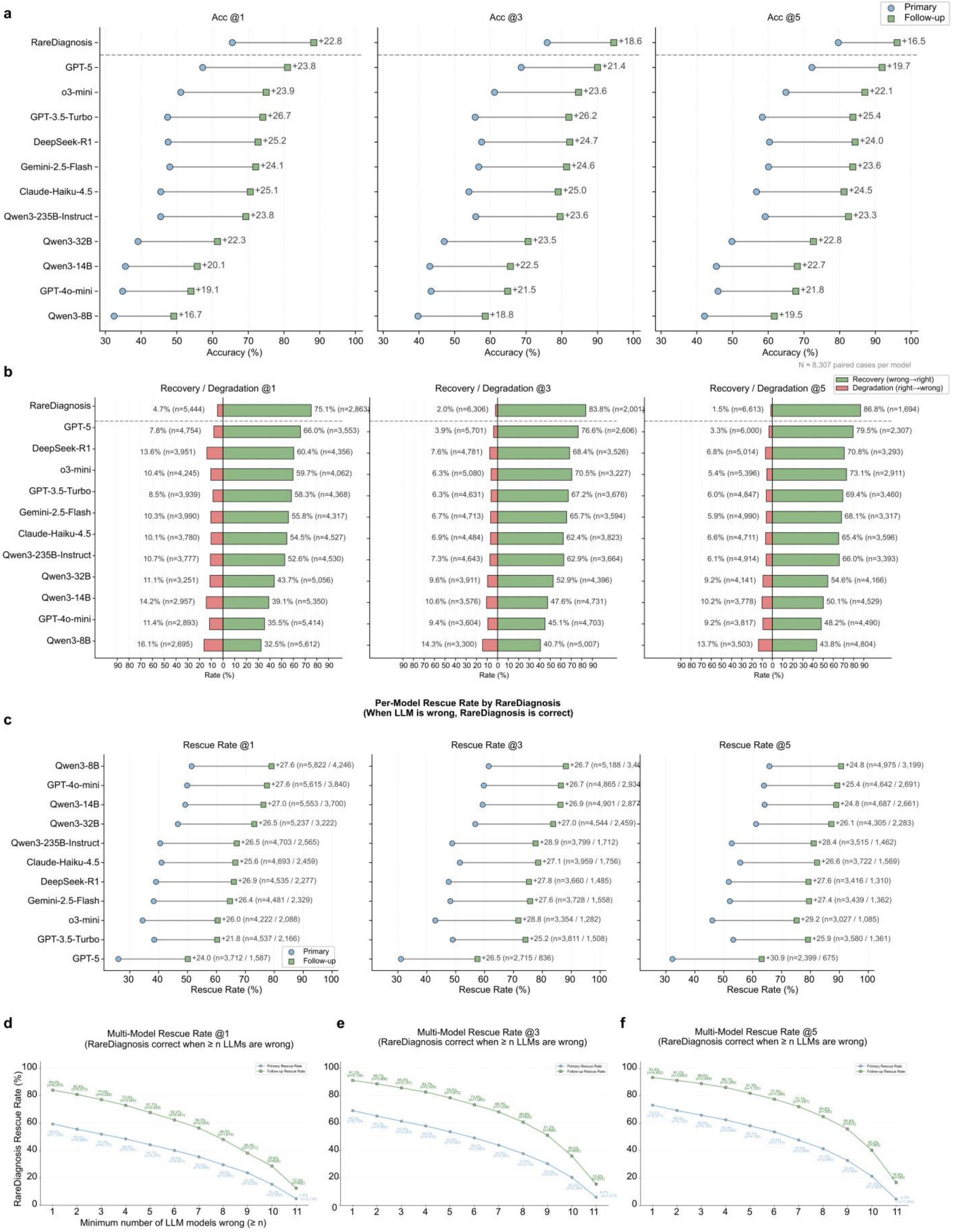

Acc@K denotes the proportion of cases for which at least one of the top K ranked diagnoses matches the reference diagnosis, where K = 1, 3 or 5. Analyses in a,b were restricted to paired cases with data available at both visits (n = 8,307), enabling within-case comparisons. Recovery rate was defined as the proportion of cases missed at the primary visit that were correctly diagnosed at follow-up, whereas degradation rate was defined as the proportion of cases correctly diagnosed at the primary visit that were missed at follow-up. Per-model rescue rate was defined as the proportion of cases missed by an individual source large language model (LLM) that were correctly diagnosed by RareDiagnosis at the same visit. Multi-model rescue rate at threshold n was defined as the proportion of cases missed by at least n source LLMs that were correctly diagnosed by RareDiagnosis at the same visit.

a, Primary- and follow-up-visit Acc@1, Acc@3 and Acc@5 for each model. Connected points represent results from the same model, and annotations show the follow-up-minus-primary difference in percentage points.

b, Recovery and degradation rates at Acc@1, Acc@3 and Acc@5 for each model. Recovery rates use cases missed at the primary visit as the denominator, whereas degradation rates use cases correctly diagnosed at the primary visit as the denominator. Annotations report rates and the corresponding numbers of recovery or degradation cases.

c, Per-model rescue rates of RareDiagnosis at Acc@1, Acc@3 and Acc@5 among cases missed by each source LLM, shown separately for primary and follow-up visits. Annotations report the rate difference and the missed-case denominators (primary/follow-up).

d, Multi-model rescue rate at Acc@1 according to the minimum number of source LLMs that missed the case.

e, Multi-model rescue rate at Acc@3 according to the minimum number of source LLMs that missed the case.

f, Multi-model rescue rate at Acc@5 according to the minimum number of source LLMs that missed the case. In d–f, threshold n includes cases missed by at least n source LLMs. Parenthetical N values indicate the number of cases in the corresponding denominator group. Results are shown separately for primary and follow-up visits.

**Supplementary Figure 8 | Further-test recommendation performance of RareDiagnosis and individual LLMs.**

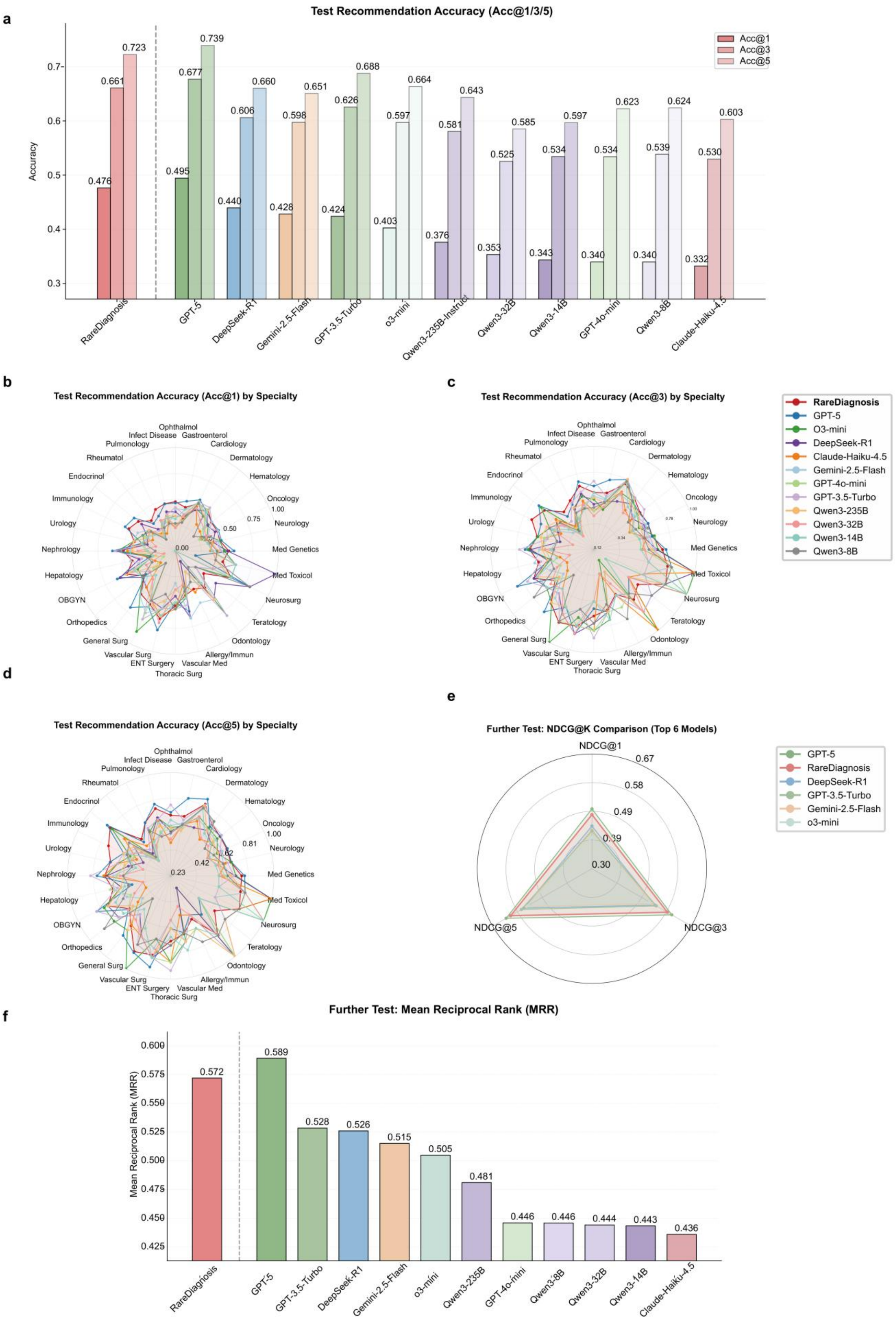

Acc@K denotes the proportion of cases for which at least one of the top K recommended tests matches a reference test, where K = 1, 3 or 5. Normalized discounted cumulative gain at K (NDCG@K) applies logarithmic rank discounting to assign greater weight to relevant tests appearing near the top of the ranked list. Mean reciprocal rank (MRR) is the mean reciprocal rank of the first correctly recommended test.

a, Acc@1, Acc@3 and Acc@5 for further-test recommendations generated by RareDiagnosis and the individual LLMs. Values above the bars indicate the corresponding Acc@K values.

b, Radar plot of speciality-specific Acc@1.

c, Radar plot of speciality-specific Acc@3.

d, Radar plot of speciality-specific Acc@5. In b–d, axes represent clinical specialities and coloured lines represent RareDiagnosis and the individual LLMs.

e, Radar plot comparing NDCG@1, NDCG@3 and NDCG@5 among the six top-performing models.

f, MRR for further-test recommendation by RareDiagnosis and the individual LLMs. Values above the bars indicate the corresponding MRR values.

**Supplementary Figure 9 | Additional RareTreatment clinical evaluation metrics.**

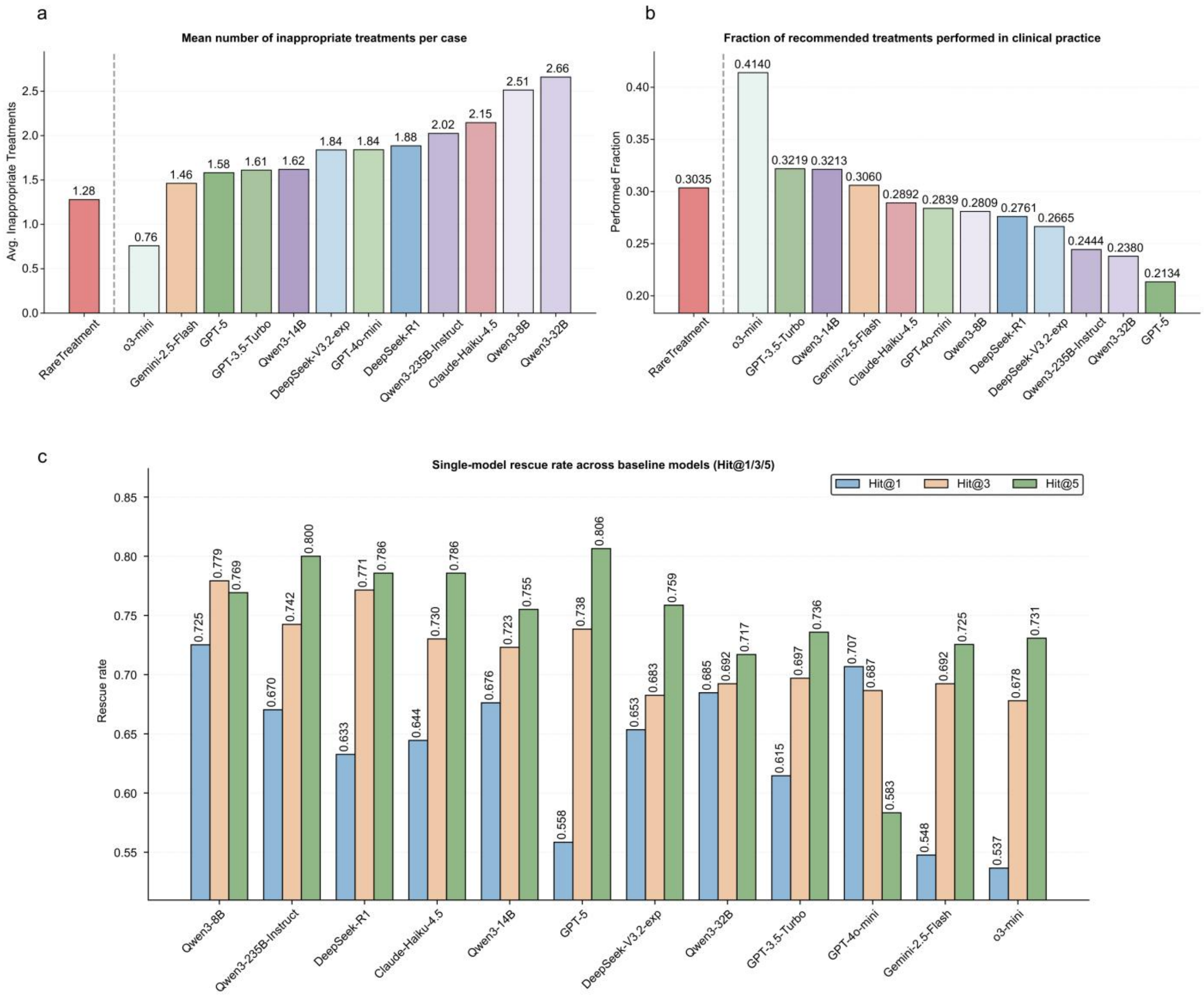

A hit at K was defined as the presence of at least one appropriate treatment among the first K ranked recommendations.

a, Mean number of inappropriate treatment recommendations per case for RareTreatment and each selected LLM.

b, Mean within-case fraction of recommended treatments that matched treatments documented as performed in clinical practice. For each case, the number of matched recommendations was divided by the total number of treatments recommended by that model, and the resulting fractions were averaged across cases.

c, Rescue rates for individual baseline LLMs at Hit@1, Hit@3 and Hit@5. Rescue rate was defined as the proportion of cases without a baseline-model hit at K for which RareTreatment achieved a hit at the same value of K.

In a and b, the vertical dashed line separates RareTreatment from the baseline LLMs.

**Supplementary Figure 10 | Absolute ordinal error and directional bias across RarePrognosis prediction tasks.**

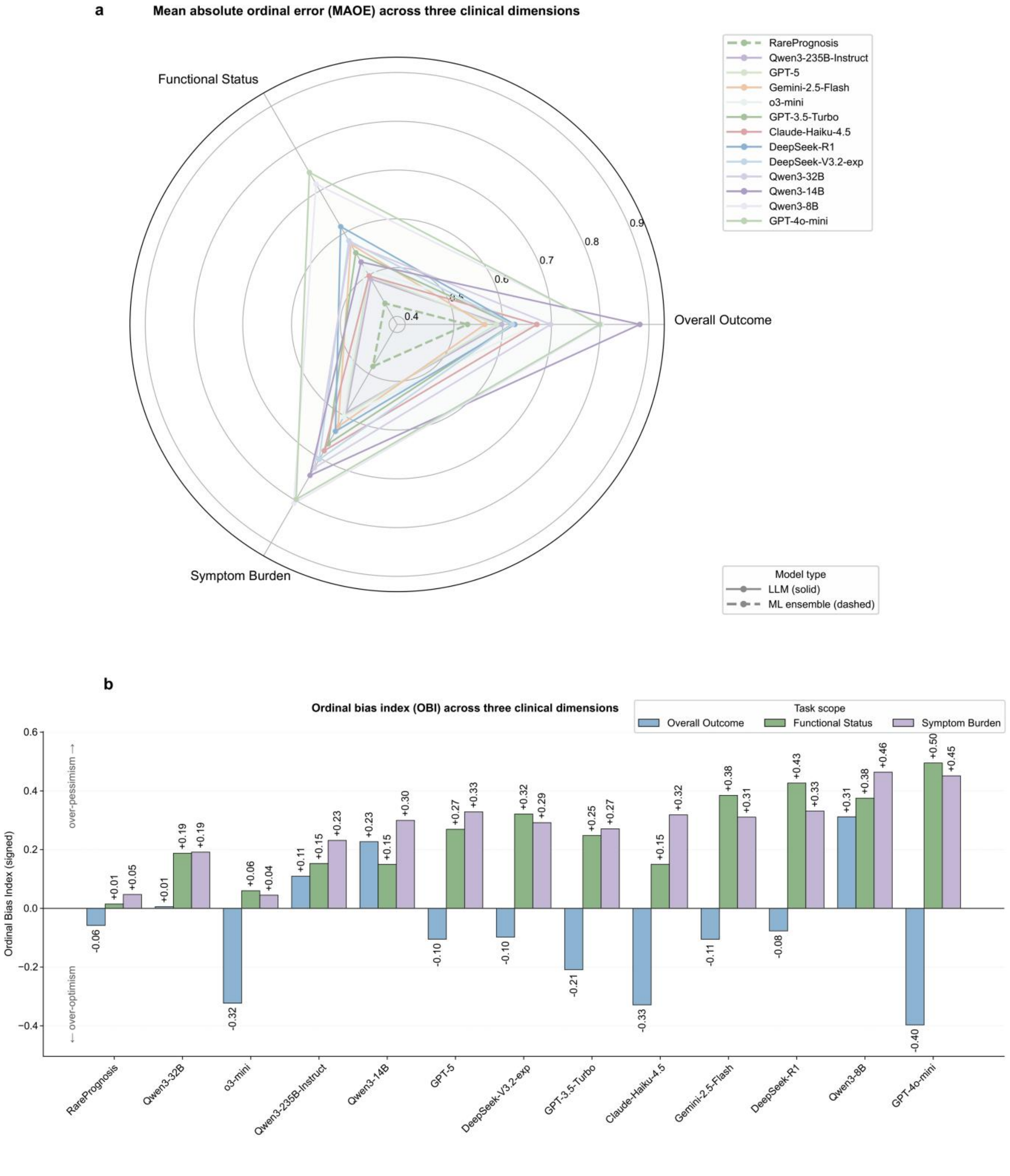

Ordered outcome categories were mapped to numeric ranks of increasing clinical severity. Mean absolute ordinal error (MAOE) is the mean absolute difference between predicted and observed ranks, with lower values indicating better ordinal agreement. The ordinal bias index (OBI) is the mean signed difference between predicted and observed ranks; positive values indicate predictions biased towards greater severity or pessimism, negative values indicate bias towards lower severity or optimism, and zero indicates no directional bias.

a, MAOE for overall outcome, functional status and symptom burden. Each line represents one model; solid lines represent individual LLMs and the dashed line represents RarePrognosis.

b, OBI for the same three clinical dimensions. Coloured bars represent the three tasks, and the horizontal line at zero indicates no directional bias.

**Supplementary Figure 11 | Case-level RareAlert reasoning-chain integration.**

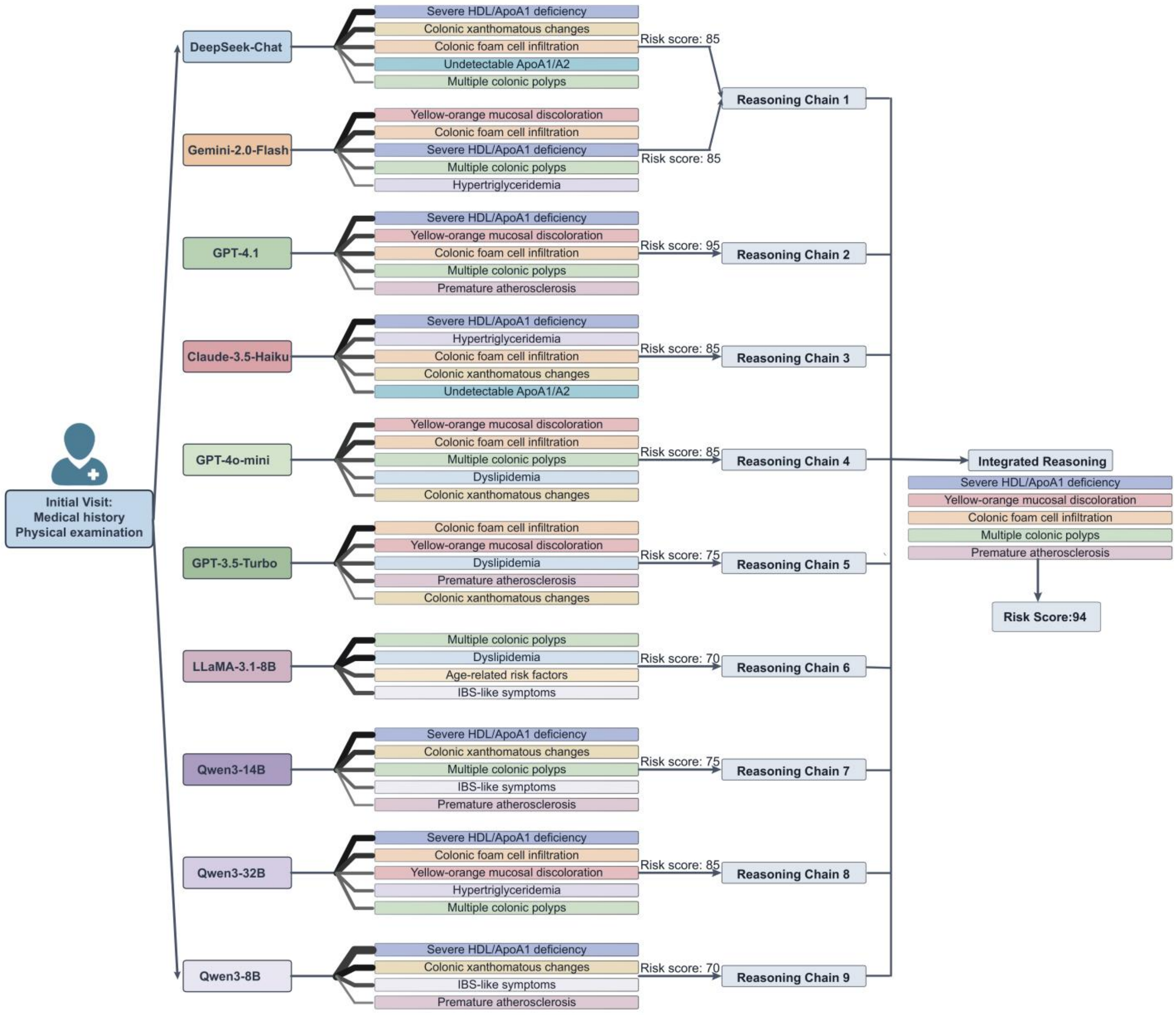

An example of how RareALert aligned medical reasoning from multiple LLMs and provided the final risk estimate.

**Supplementary Figure 12 |Risk-score and reasoning-chain characteristics in rare and non-rare disease cases**

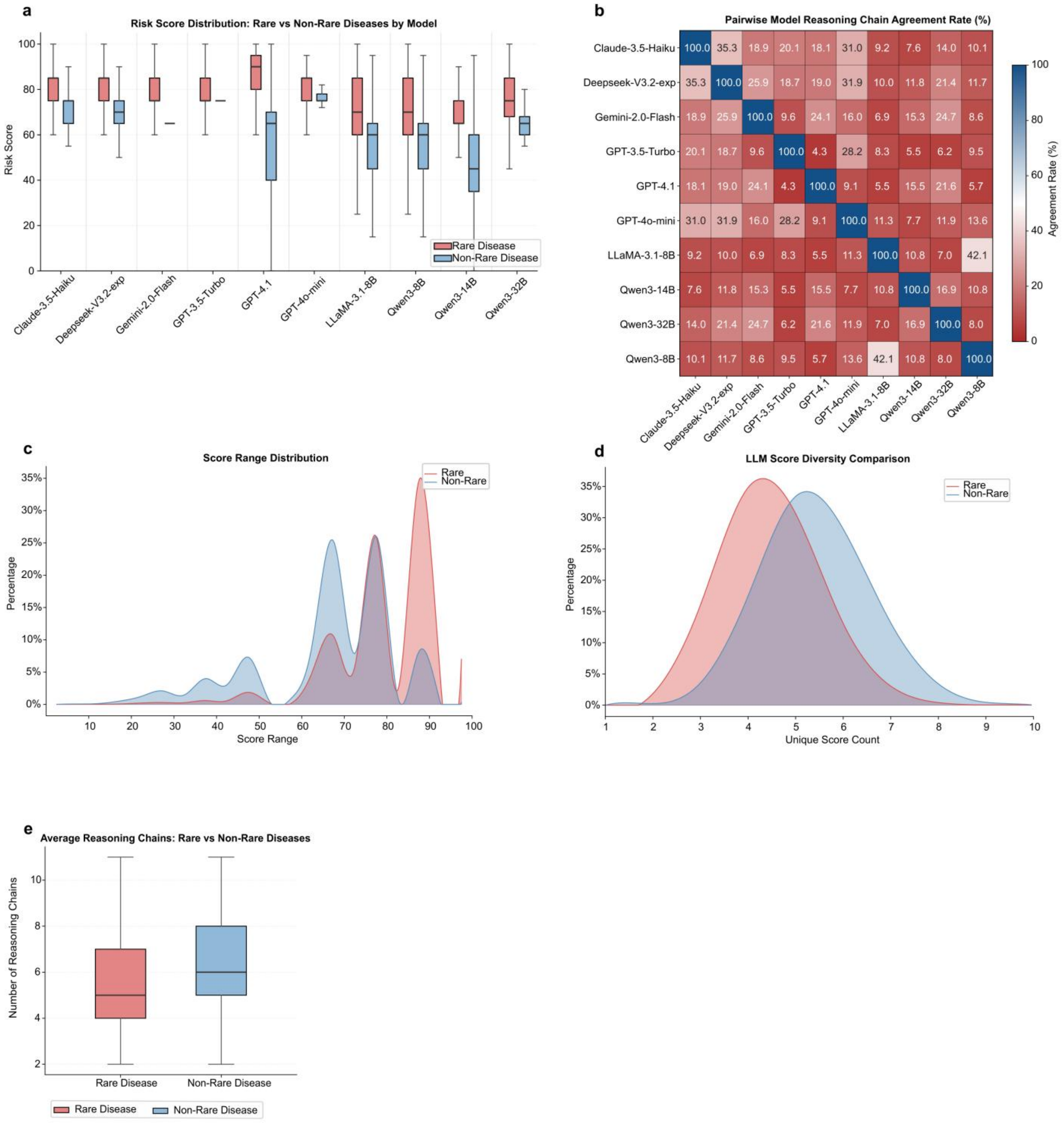

a, Model-wise distributions of risk scores assigned by the source LLMs to rare and non-rare disease cases, shown as box plots.

b, Pairwise agreement rates for medical reasoning chains generated by the source LLMs. Bluer cells indicate higher agreement and redder cells indicate lower agreement; annotations show the corresponding percentages.

c, Distributions of risk-score from all tested LLMs for rare and non-rare disease cases.

d, Distribution of the number of unique risk scores produced by LLMs per case, expressed as the percentage of cases at each unique-score count, for rare (red) and non-rare (blue) disease cases.

e, Comparison of number of reasoning chains generated by LLMs for rare and non-rare disease cases a, Mean AUC of candidate machine-learning algorithms evaluated for all LLMs configurations b, Best AUC achieved by each candidate algorithm as the number of included LLMs increased from 1 to 10. For a given number of LLMs (k), many different LLM combinations are possible; each combination was evaluated separately, and the highest AUC among all combinations of that size is reported.

**Supplementary Figure 13 |Comparison of machine-learning algorithms and the effect of the number of included LLMs on RareAlert performance.**

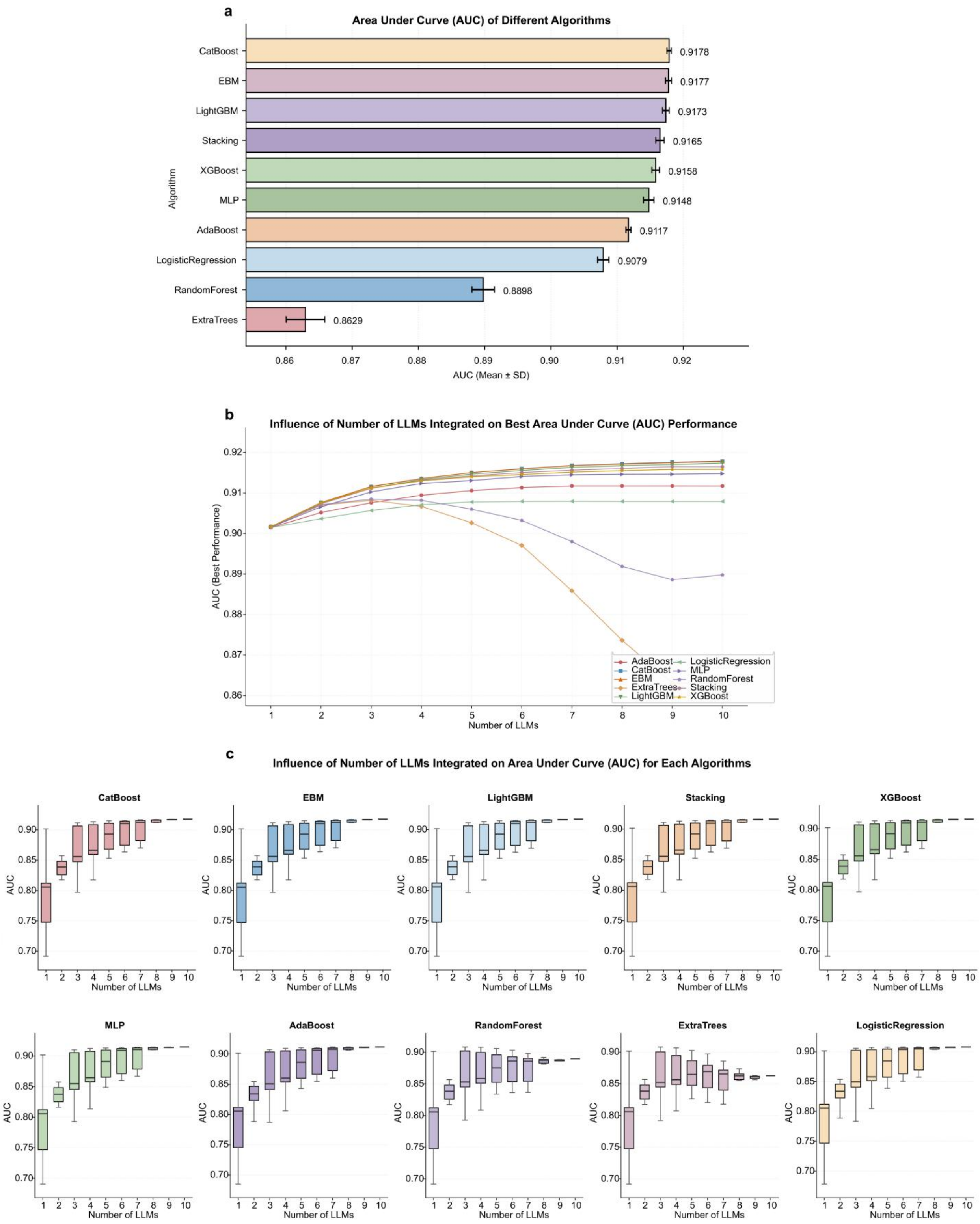

c, Distributions of AUC values obtained from different LLM combinations for each machine-learning algorithm, stratified by the number of included LLMs. For a given number of LLMs (k), many different LLM combinations are possible; each combination was evaluated and reported.

**Supplementary Figure 14 | Sensitivity analyses of the number of reasoning paths and alternative ensemble strategies for RareAlert.**

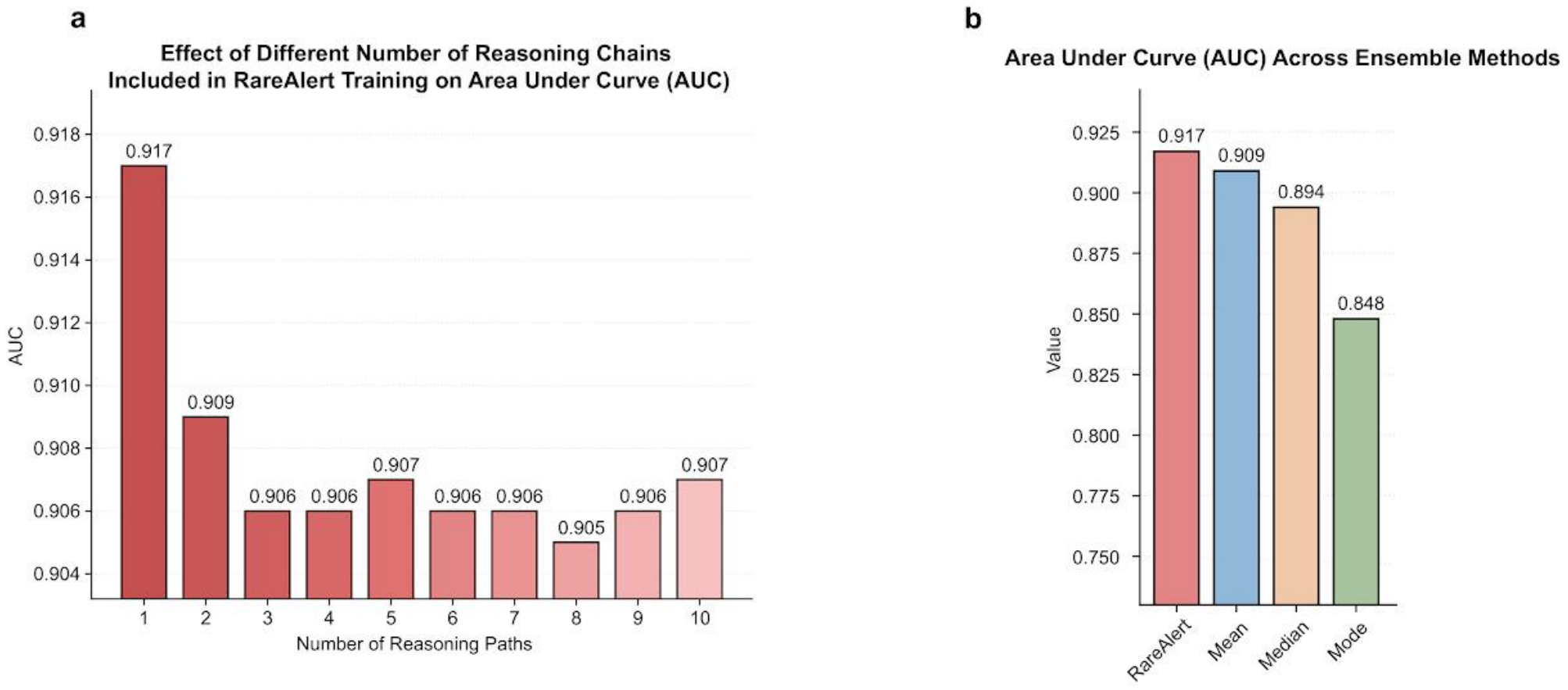

AUC denotes the area under the receiver operating characteristic curve.

a, AUC of RareAlert models trained using between 1 and 10 reasoning paths per case.

b, Comparison of AUC between RareAlert and alternative ensemble strategies, including mean, median and mode aggregation.

**Supplementary Figure 15 | RareDiagnosis feature sensitivity and learning-to-rank model selection.**

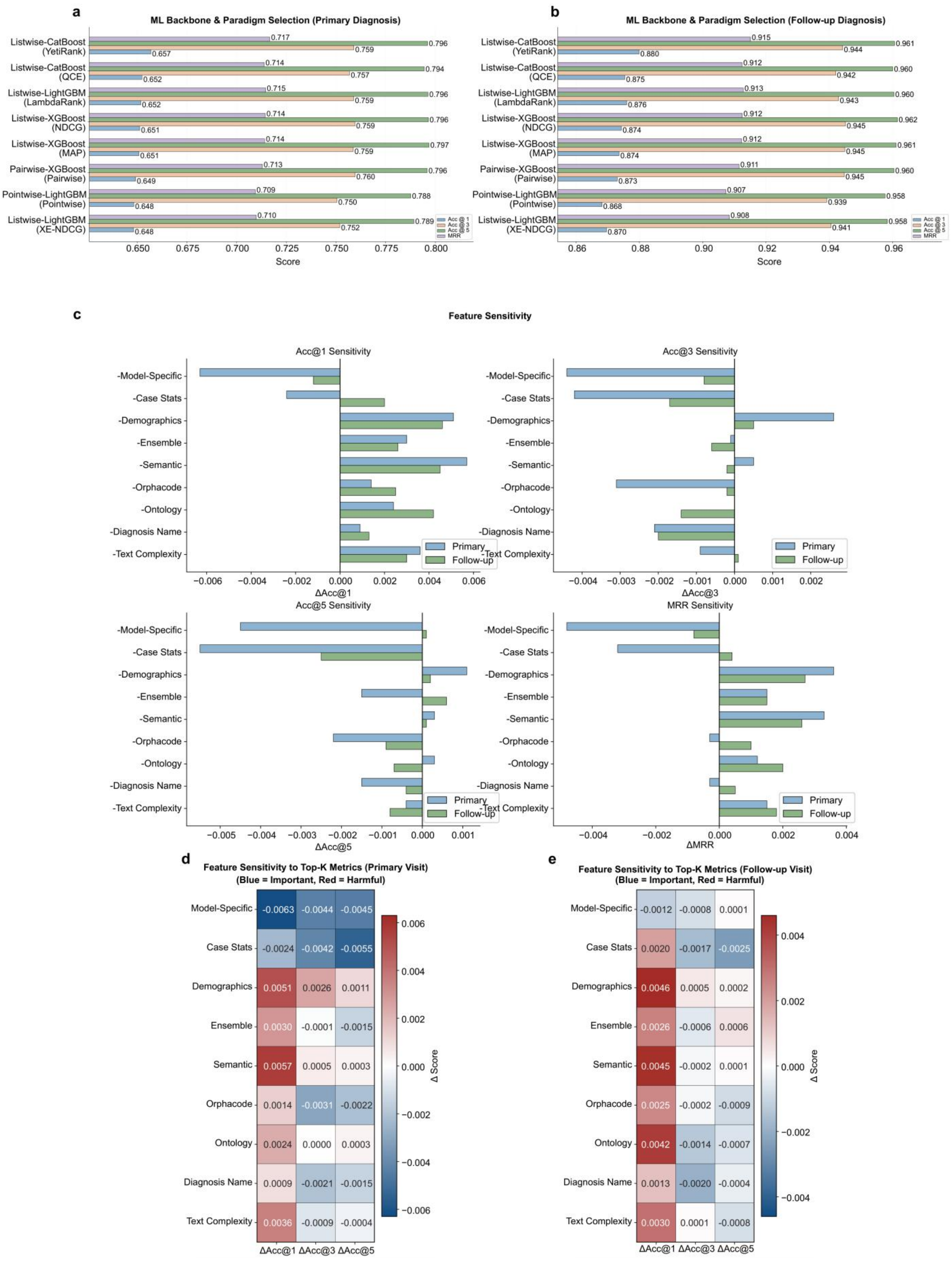

Candidate gradient-boosted decision-tree frameworks and learning-to-rank objectives were evaluated using the full feature set. Feature sensitivity was defined as the performance after removing one predefined feature group minus the corresponding full-feature performance. Negative values therefore indicate a decrease in performance after feature removal and suggest a positive contribution of the removed feature group, whereas positive values indicate improved performance after removal and suggest that the feature group may be detrimental. MRR denotes the mean reciprocal rank of the reference diagnosis. Acc@K denotes the proportion of cases for which at least one of the top K ranked candidates matches the reference diagnosis, where K = 1, 3 or 5.

a, Primary-visit Acc@1, Acc@3 and Acc@5 for candidate machine-learning backbones and learning-to-rank objectives using the full feature set.

b, Follow-up-visit Acc@1, Acc@3 and Acc@5 for the same candidate models using the full feature set.

c, Changes in Acc@1, Acc@3, Acc@5 and MRR after removing each feature group, shown separately for primary and follow-up visits.

d, Heatmap of changes in primary-visit Acc@1, Acc@3 and Acc@5 after feature-group removal.

e, Heatmap of changes in follow-up-visit Acc@1, Acc@3 and Acc@5 after feature-group removal. In d and e, bluer cells indicate decreases in performance after feature removal, whereas redder cells indicate improvements; annotations show the corresponding changes in score.

**Supplementary Figure 16 | Training-set analysis of the effect of the number of included LLMs on RareDiagnosis ranking performance.**

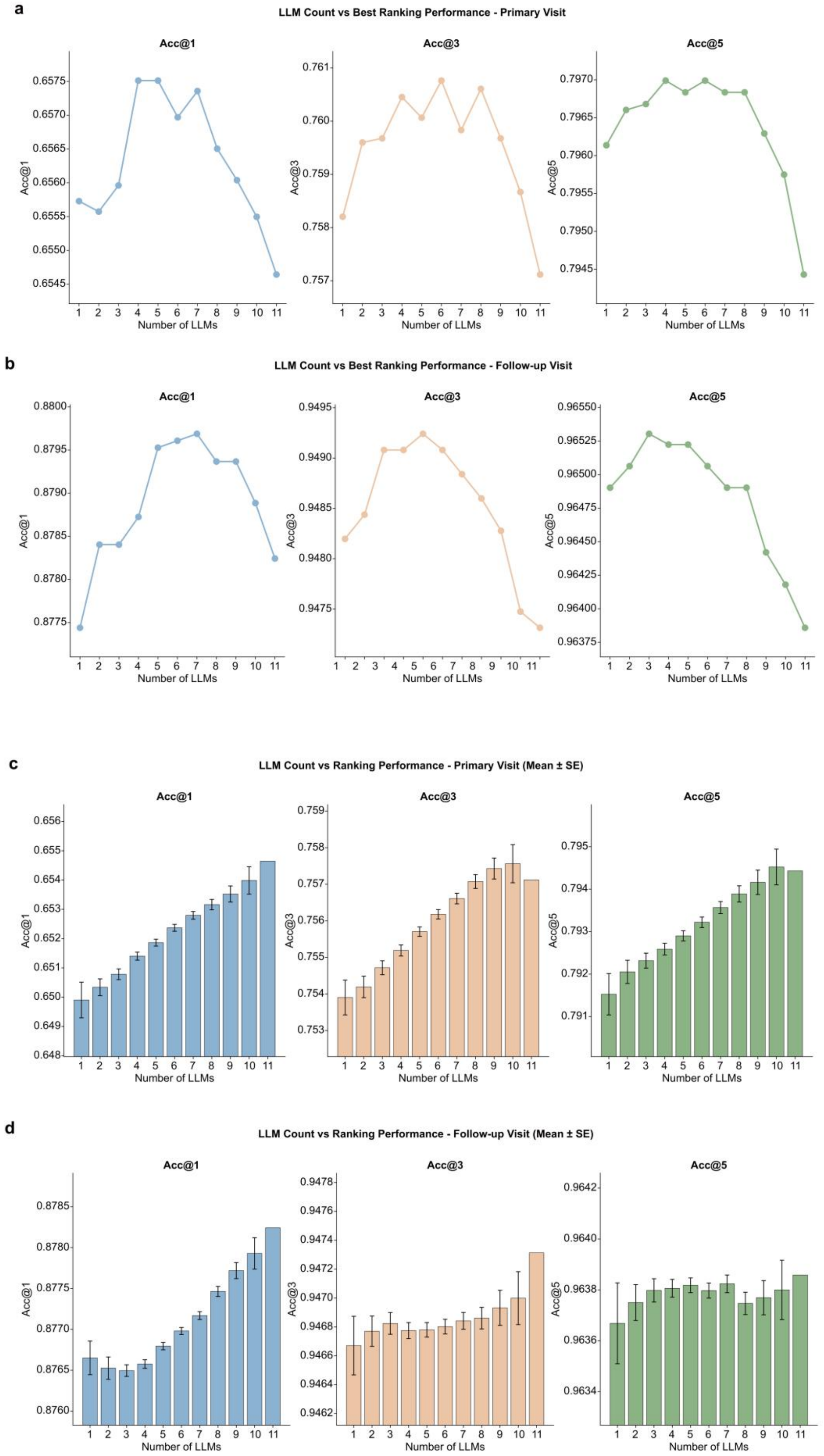

Acc@K denotes the proportion of cases for which at least one of the top K ranked candidates matches the reference diagnosis, where K = 1, 3 or 5. For each number of included large language models (LLMs), the best-performing result for each Acc@K metric was the highest value obtained across all evaluated LLM combinations. Mean values and standard errors of the mean (s.e.m.) were calculated across all LLM combinations containing the same number of LLMs.

a, Best primary-visit Acc@1, Acc@3 and Acc@5 achieved for each number of included LLMs.

b, Best follow-up-visit Acc@1, Acc@3 and Acc@5 achieved for each number of included LLMs.

c, Mean primary-visit Acc@1, Acc@3 and Acc@5 across all evaluated LLM combinations for each number of included LLMs. Error bars indicate s.e.m. across LLM combinations.

d, Mean follow-up-visit Acc@1, Acc@3 and Acc@5 across all evaluated LLM combinations for each number of included LLMs. Error bars indicate s.e.m. across LLM combinations.

**Supplementary Figure 17 | Candidate-score distributions, within-case score separation and source-model contributions in RareDiagnosis.**

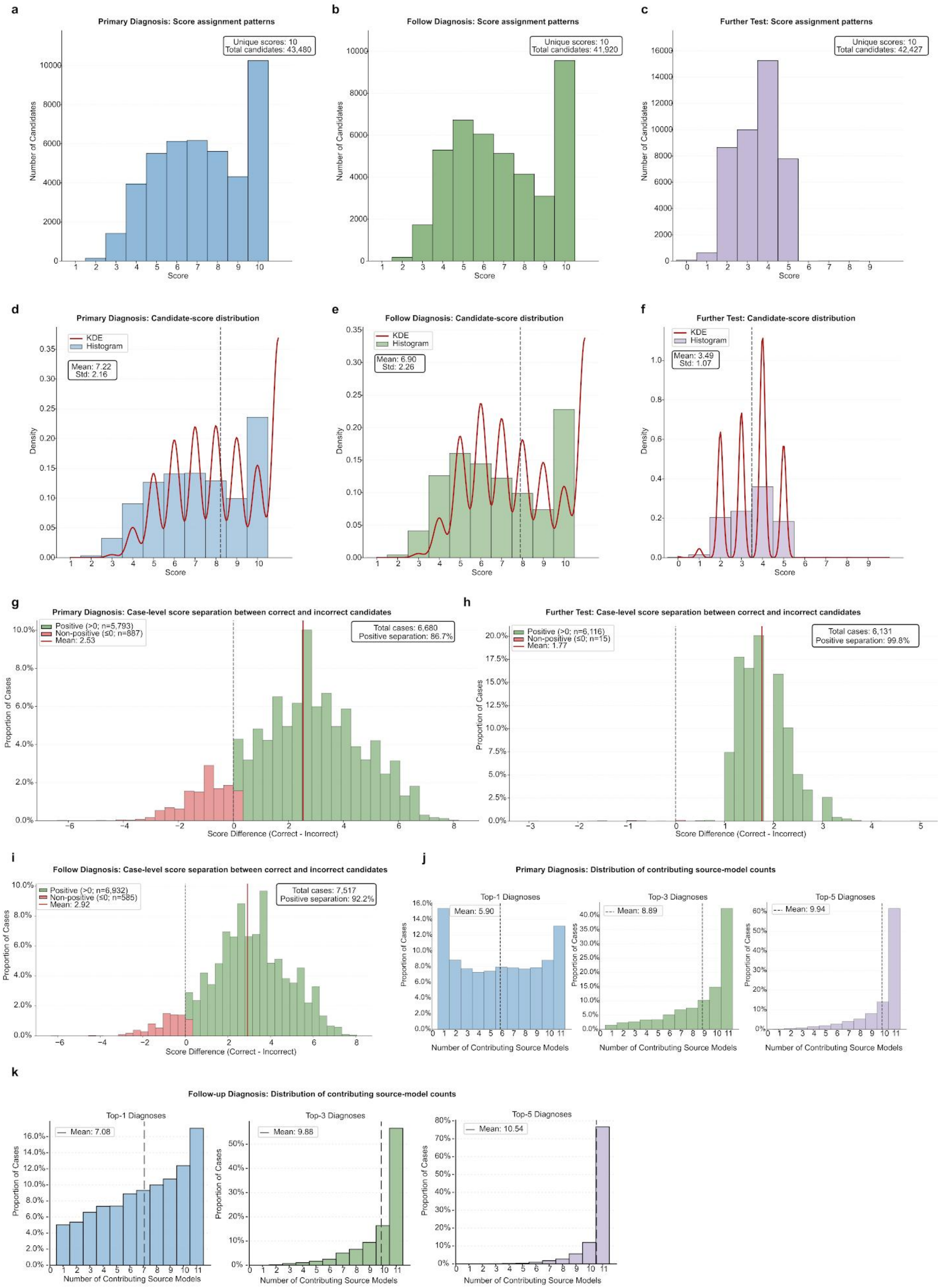

Candidate scores were analysed for primary diagnosis, follow-up diagnosis and further-test recommendation. Case-level score separation was defined as the mean score of correct candidates minus that of incorrect candidates within each case; positive values indicate higher scores for correct candidates. For each case, the contributing source-model count was defined as the number of distinct large language models (LLMs) that contributed at least one diagnosis to the top-K candidates, after combining the source-model sets for all K candidates.

a, Frequency distribution of scores assigned to primary-diagnosis candidates.

b, Frequency distribution of scores assigned to follow-up-diagnosis candidates.

c, Frequency distribution of scores assigned to further-test candidates. In a–c, annotations report the number of unique scores and the total number of candidates.

d, Score distribution for primary-diagnosis candidates.

e, Score distribution for follow-up-diagnosis candidates.

f, Score distribution for further-test candidates. In d–f, histograms show score densities, red curves show kernel density estimates (KDEs), dashed lines indicate mean scores, and annotations report means and s.d.

g, Distribution of case-level score separation for primary-diagnosis candidates.

h, Distribution of case-level score separation for further-test candidates.

i, Distribution of case-level score separation for follow-up-diagnosis candidates. In g–i, green and red bars indicate positive and negative separation, respectively. Dashed lines at zero indicate no separation, and red vertical lines indicate mean score differences. Annotations report case counts and proportions with positive separation.

j, Distributions of the numbers of distinct source LLMs contributing to the top-1, top-3 and top-5 primary-diagnosis candidates.

k, Distributions of the numbers of distinct source LLMs contributing to the top-1, top-3 and top-5 follow-up-diagnosis candidates. In j and k, the x axes show contributing LLM counts, the y axes show case proportions, and dashed lines indicate means. KDE, kernel density estimate; s.d., standard deviation.

**Supplementary Figure 18 | Cross-model heterogeneity in LLM-generated treatment recommendations in RareTreatment.**

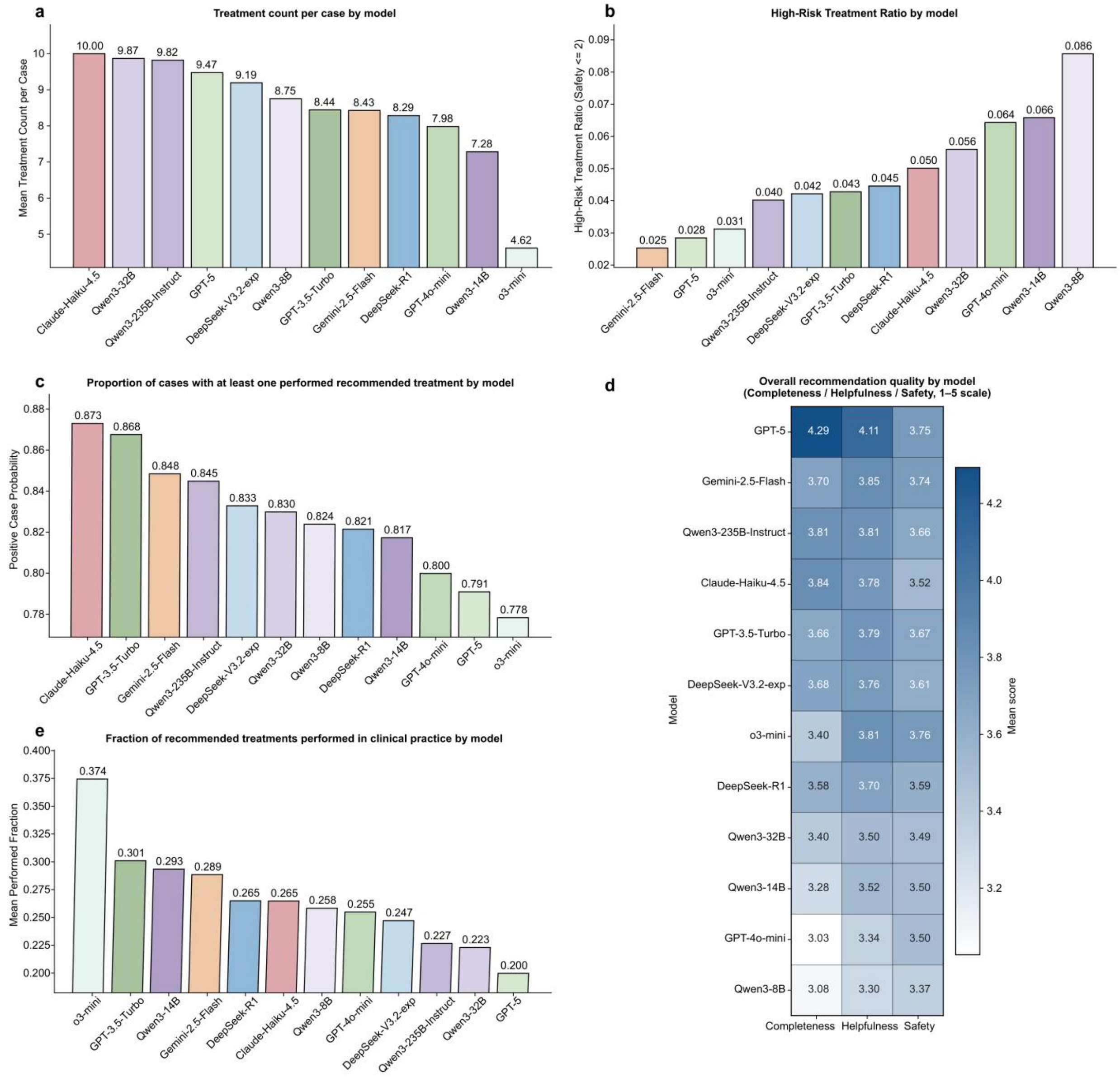

Completeness, helpfulness and safety were rated on ordinal scales from 1 to 5, with higher scores indicating better performance. High-risk recommendations were defined as those receiving a safety score of 2 or lower. Recommended treatments were compared with treatments documented as performed in clinical practice.

a, Mean number of treatment recommendations generated per case by each model.

b, Proportion of treatment recommendations classified as high risk for each model.

c, Proportion of cases for which at least one model-recommended treatment matched a treatment documented as performed in clinical practice.

d, Mean completeness, helpfulness and safety scores for treatment recommendations generated by each model. Bluer cells indicate higher mean scores.

e, Mean within-case fraction of recommended treatments that matched treatments documented as performed in clinical practice. For each case, this fraction was calculated as the number of matched recommended treatments divided by the total number of treatments recommended by the model and then averaged across cases.

**Supplementary Figure 19 | RareTreatment confidence and consistency analyses.**

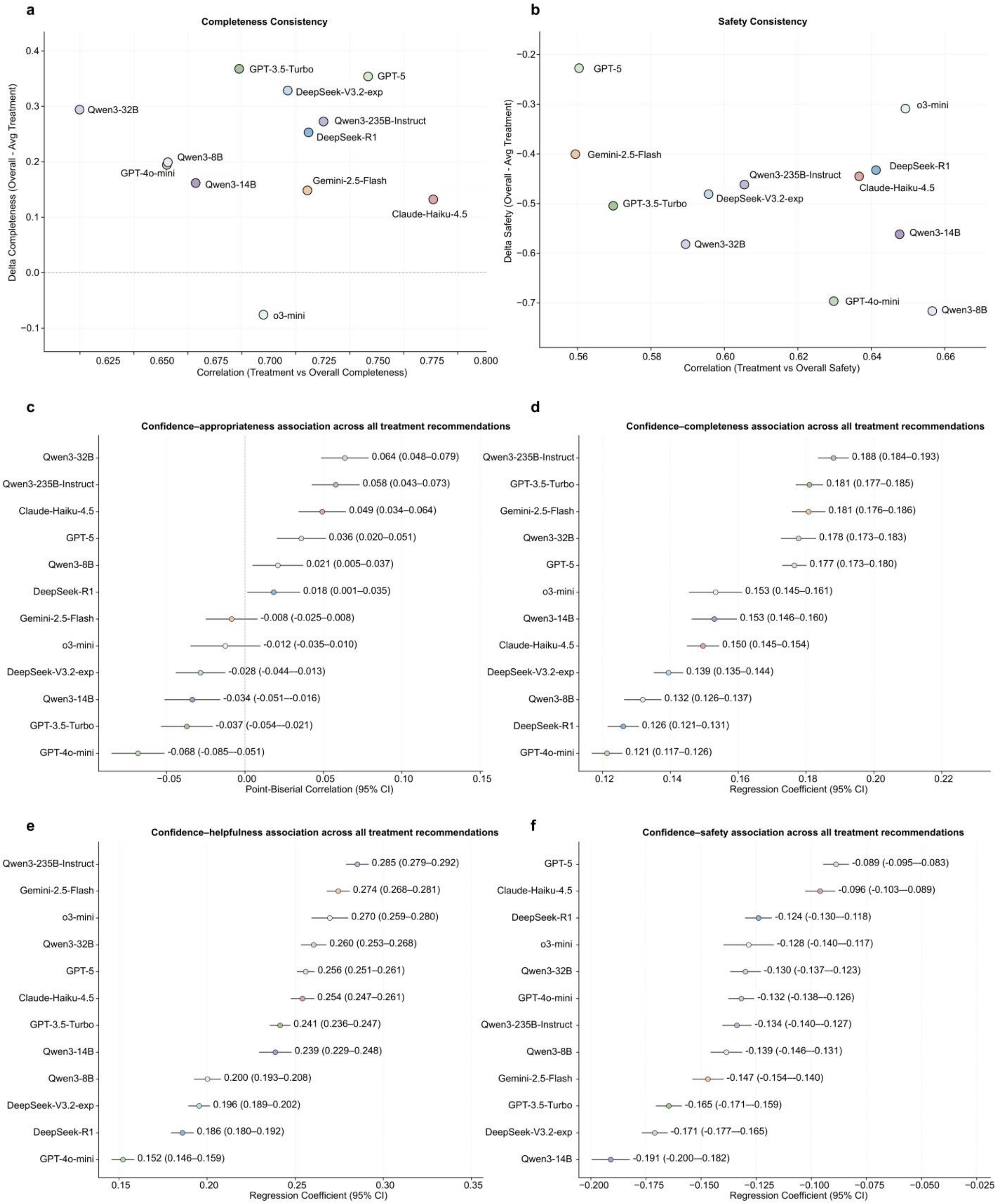

Each treatment recommendation was treated as one candidate. Completeness, helpfulness and safety were rated on 1–5 ordinal scales, with higher scores indicating better performance. Appropriateness was binary (appropriate = 1; inappropriate = 0), and importance scores ranged from 0 to 10.

a, Consistency between the overall case-level completeness rating and the within-case average of candidate-level completeness ratings. Each point represents one baseline LLM. The horizontal axis shows Pearson's correlation coefficient between the two measures across cases; the vertical axis shows their mean paired difference (case-level rating − candidate-level average).

b, Consistency between case- and candidate-level safety ratings, presented as in a. Horizontal dashed lines in a and b indicate zero mean difference.

c, Point-biserial correlation between importance score and candidate appropriateness across all candidates. Confidence intervals were calculated using Fisher's z-transformation.

d, Association between importance score and candidate-level completeness.

e, Association between importance score and candidate-level helpfulness.

f, Association between importance score and candidate-level safety.

Associations in d–f were estimated using linear mixed-effects models with importance score as a fixed effect and case ID as a random intercept. In c–f, points indicate model-specific estimates and horizontal lines indicate 95% confidence intervals. Coefficients in d–f represent the expected change in the corresponding 1–5 rating per one-point increase in importance score.

**Supplementary Figure 20 | RareTreatment feature sensitivity and learning-to-rank model selection.**

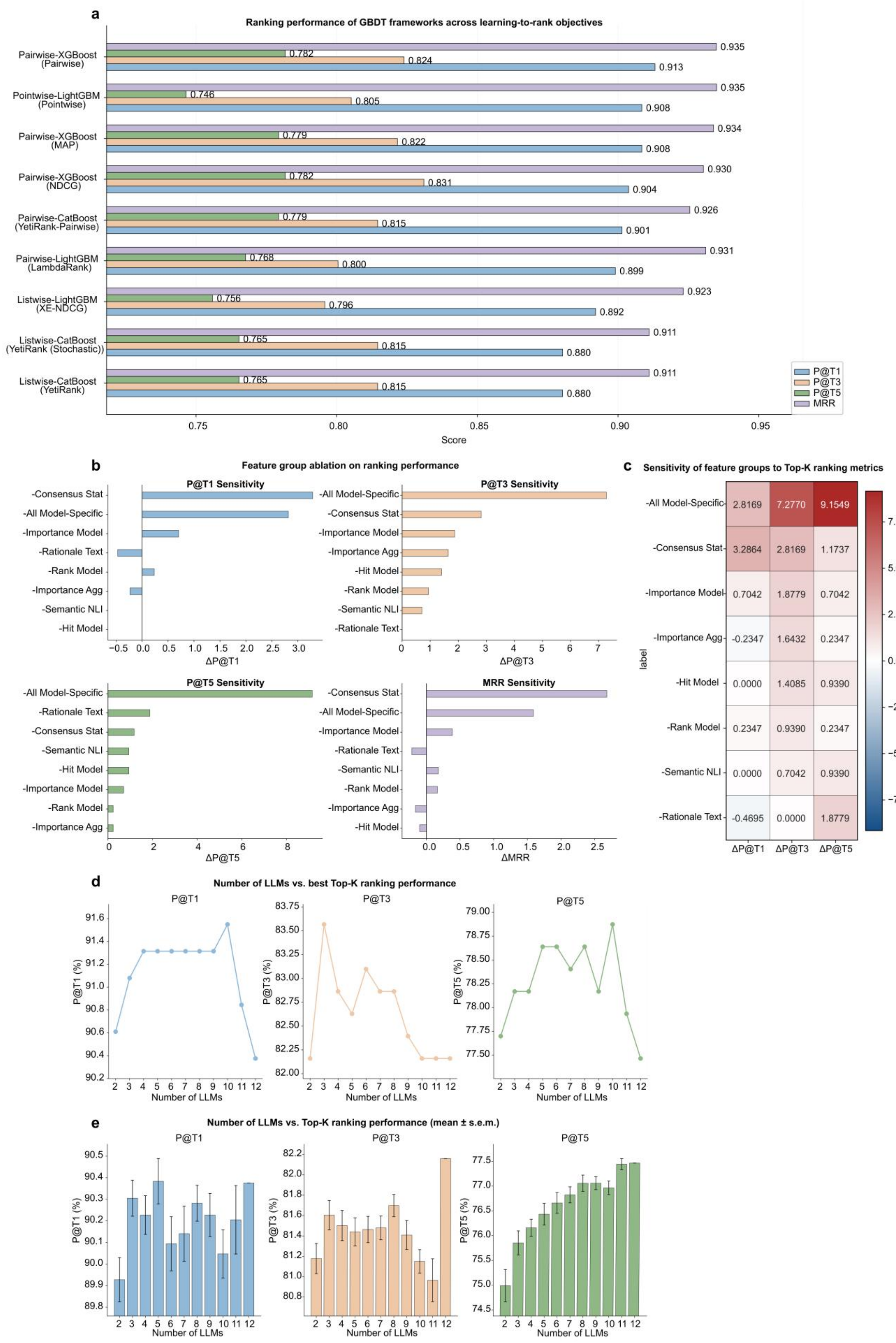

P@T(k) (hereafter referred to as All-appropriate@k) denotes the strict top-k success rate. P@T1, P@T3 and P@T5 are defined as the proportions of cases for which all of the first 1, 3 or 5 ranked treatment recommendations, respectively, were labelled appropriate. A case was considered unsuccessful if any recommendation among the first k was labelled inappropriate. MRR denotes the mean reciprocal rank of the first appropriate treatment, with a value of 0 assigned when no appropriate treatment was retrieved. Feature-ablation effects were calculated as the full-model score minus the score after feature removal; thus, larger positive values indicate greater feature importance, whereas negative values indicate improved performance after removal.

a, Ranking performance of gradient-boosted decision-tree frameworks and learning-to-rank objectives. Bars show P@T1, P@T3, P@T5 and MRR for each framework–objective configuration.

b, Leave-one-feature-group-out ablation analysis for P@T1, P@T3, P@T5 and MRR. Bars show the performance loss after removal of each feature group relative to the full reranker, expressed on a 100-point scale. Positive values indicate reduced performance, whereas negative values indicate improved performance after feature removal.

c, Heatmap summarizing the feature-ablation effects on P@T1, P@T3 and P@T5. Cell values show the performance changes defined in b, with warmer colours indicating larger performance losses.

d, Best P@T1, P@T3 and P@T5 achieved for each ensemble size. For each number of included large language models (LLMs), the plotted value is the highest score observed across all evaluated LLM combinations of that size.

e, Mean P@T1, P@T3 and P@T5 across the same LLM combinations for each ensemble size. Bars indicate means and error bars indicate the standard error of the mean (s.e.m.) across LLM combinations.

**Supplementary Figure 21 | Candidate-score distributions and case-level score separation and source-model contributions in RareTreatment.**

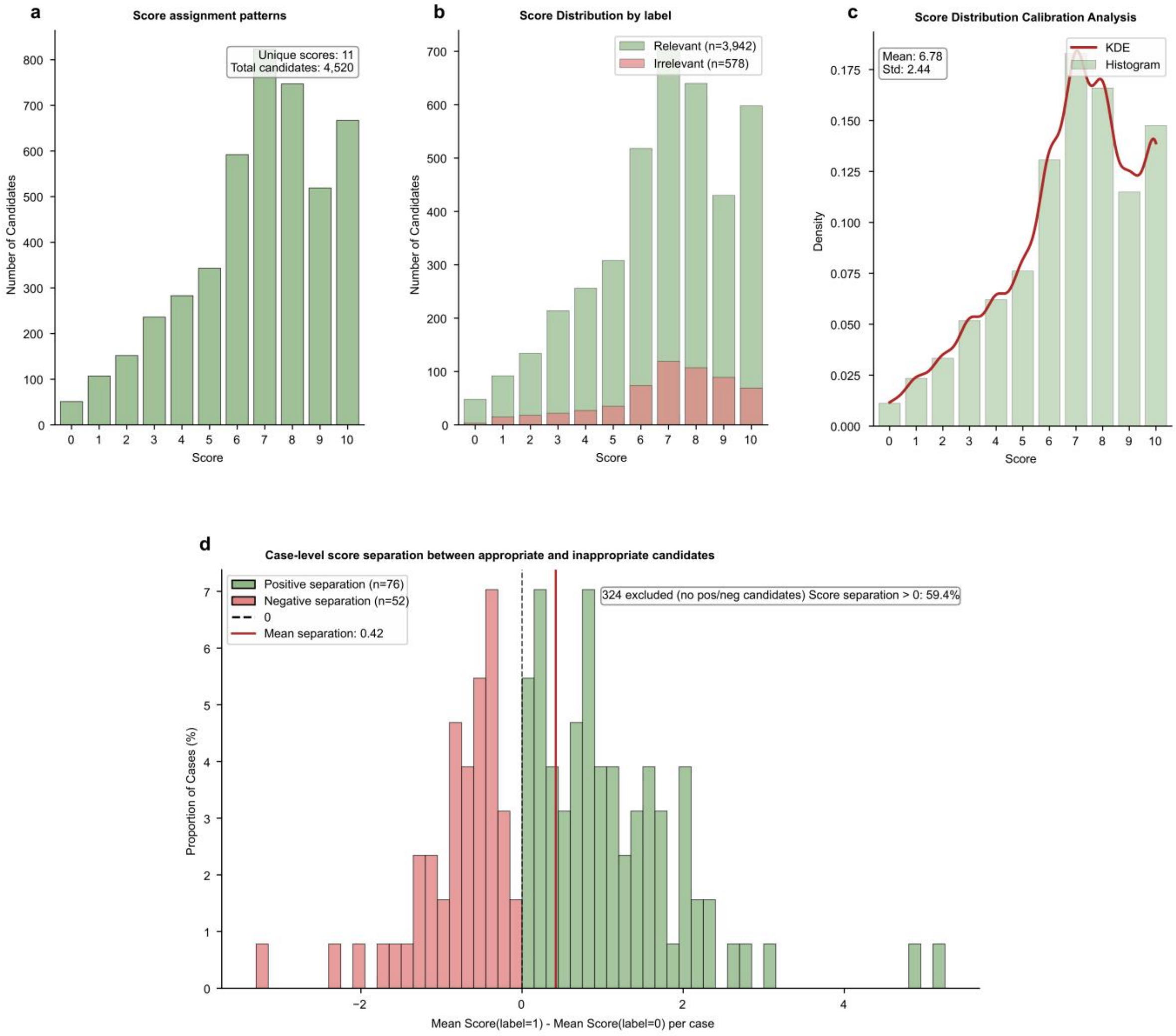

Candidate scores ranged from 0 to 10, with higher scores indicating higher-ranked treatment candidates. Case-level score separation was defined as the mean score of appropriate candidates minus that of inappropriate candidates within each case; positive values indicate higher scores for appropriate candidates.

a, Frequency distribution of scores assigned to treatment candidates. The annotation reports the number of unique scores and the total number of candidates.

b, Stacked frequency distributions of scores for appropriate and inappropriate treatment candidates.

c, Overall treatment-candidate score distribution. The histogram shows score density, the red curve shows the kernel density estimate (KDE), and the annotation reports the mean and s.d.

d, Distribution of case-level score separation. Green and red bars indicate positive and negative separation, respectively. The dashed line at zero indicates no separation, and the red vertical line indicates the mean score difference. The analysis included 128 cases containing both candidate classes; 324 of 452 cases containing only one class were excluded. Annotations report the case counts and proportion with positive separation.

**Supplementary Figure 22 | RarePrognosis model-backbone and paradigm selection.**

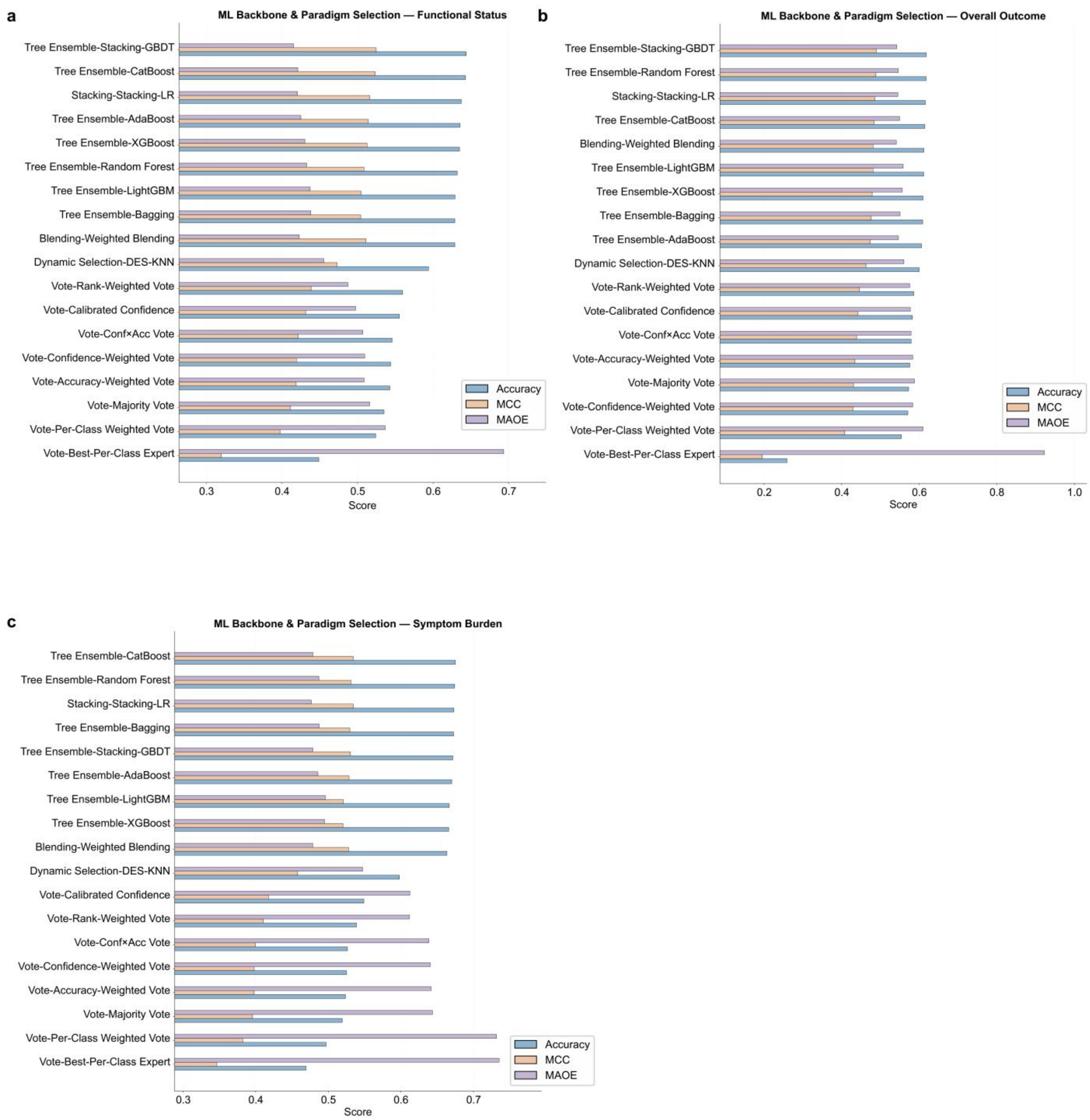

Candidate voting, blending, dynamic-selection, stacking and tree-ensemble configurations were evaluated across three clinical prognosis tasks. Accuracy denotes the proportion of cases assigned the exact outcome category, MCC denotes the multiclass Matthews correlation coefficient, and mean absolute ordinal error (MAOE) denotes the mean absolute difference between predicted and observed category ranks. Higher accuracy and MCC values indicate better performance, whereas lower MAOE values indicate better ordinal agreement.

a, Performance of candidate configurations for functional-status prediction.

b, Performance of candidate configurations for overall-outcome prediction.

c, Performance of candidate configurations for symptom-burden prediction.

In a–c, horizontal bars show accuracy, MCC and MAOE for each learning-paradigm and backbone configuration.

**Supplementary Figure 23 | RarePrognosis feature sensitivity and ensemble-size analyses.**

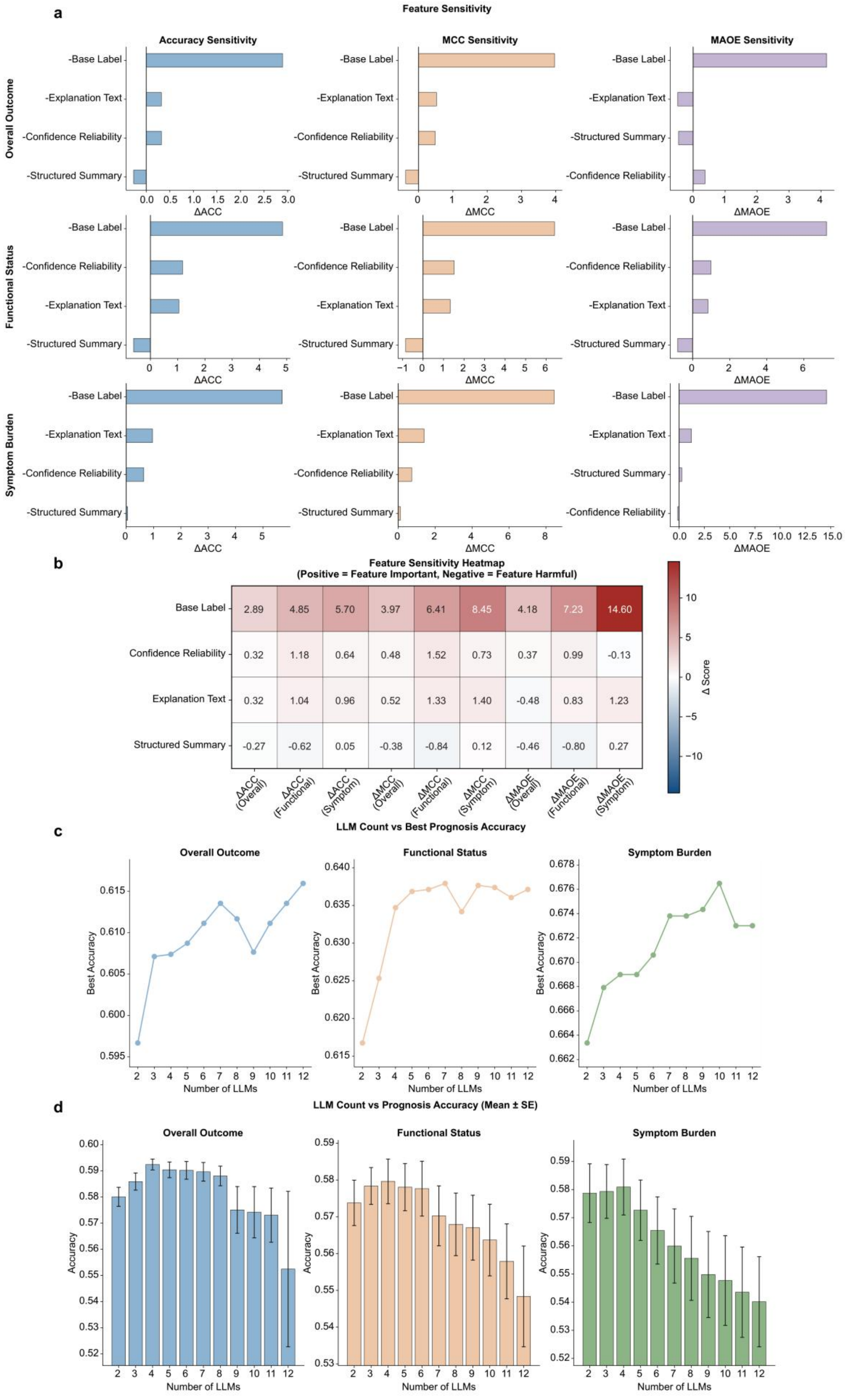

Feature sensitivity was calculated in the direction of performance loss after feature removal: decreases in accuracy or Matthews correlation coefficient (MCC), and increases in mean absolute ordinal error (MAOE), are shown as positive values. Positive values therefore indicate a beneficial contribution from the removed feature group, whereas negative values indicate improved performance after its removal. Ablation values are expressed on a 100-point scale.

a, Performance loss after removing each feature group, evaluated using accuracy, MCC and MAOE across the three prognosis tasks. Vertical lines indicate no change.

b, Heatmap summarizing feature sensitivity across tasks and metrics. Redder cells indicate greater performance loss after feature removal, whereas bluer cells indicate improved performance; annotations show the corresponding changes in score.

c, Best accuracy achieved across the same LLM combinations. For each task and ensemble size, points show the highest accuracy across all evaluated LLM combinations of that size.

d, Mean accuracy across the same LLM combinations. Bars indicate means and error bars indicate s.e.m. across combinations.

# Supplementary Tables

## Supplementary Table 1 | Definitions of primary, secondary, inter-model heterogeneity and RareLens candidate-level metrics

| Module / analysis | Metric name | Definition and calculation |
|---|---|---|
| **1. Primary performance metrics** | | |
| **RareAlert** | | |
| RareAlert | Overall accuracy | At the operating threshold selected by the Youden index, (TP + TN) / (TP + TN + FP + FN): the proportion of all cases classified correctly as rare-disease-positive or non-rare-disease. |
| **RareDiagnosis - diagnosis ranking** | | |
| RareDiagnosis - diagnosis ranking | Diagnostic Acc@k (k = 1, 3, 5) | Calculated separately at the primary and follow-up visits. For each case, Acc@k records whether at least one correct diagnosis appears among the first k ranked candidates. The reported Acc@k is the proportion of eligible cases satisfying this criterion. |
| **RareDiagnosis - further-test recommendation** | | |
| RareDiagnosis - further-test recommendation | Further-test Acc@k (k = 1, 3, 5) | For each case, Acc@k records whether at least one of the first k recommended tests is judged correct under the prespecified adjudication criterion. The reported Acc@k is the proportion of eligible cases satisfying this criterion. |
| **RareTreatment** | | |
| RareTreatment | Any-hit Acc@k (k = 1, 3, 5) | For each case, Any-hit Acc@k records whether at least one of the first k recommended treatments is judged appropriate under the prespecified adjudication criterion. The reported Any-hit Acc@k is the proportion of eligible cases satisfying this criterion. |
| **RarePrognosis** | | |
| RarePrognosis | Task-specific exact-match accuracy | Calculated separately for overall outcome, functional status and symptom burden as the proportion of cases for which the predicted ordinal category exactly matches the reference category. |
| RarePrognosis | All-three-correct (joint) accuracy | The proportion of cases for which overall outcome, functional status and symptom burden are all predicted correctly for the same patient. |
| **2. Secondary performance and clinical-quality metrics** | | |
| **RareAlert** | | |
| RareAlert | Area under the receiver operating characteristic curve (AUC-ROC) | The area under the curve plotting sensitivity against 1 - specificity across all decision thresholds; equivalently, the probability that a randomly selected rare-disease case receives a higher risk score than a randomly selected non-rare-disease case. |
| RareAlert | Youden index | J = sensitivity + specificity - 1. The candidate threshold maximizing J is used to dichotomize the risk score for all threshold-dependent RareAlert metrics. |
| RareAlert | Sensitivity | TP / (TP + FN): the proportion of true rare-disease cases classified as high risk. |
| RareAlert | Specificity | TN / (TN + FP): the proportion of non-rare-disease cases classified as low risk. |
| RareAlert | Positive predictive value (PPV) | TP / (TP + FP): the proportion of high-risk predictions that are true rare-disease cases. |
| RareAlert | Negative predictive value (NPV) | TN / (TN + FN): the proportion of low-risk predictions that are true non-rare-disease cases. |
| RareAlert | False-negative rate (FNR) | FN / (TP + FN) = 1 - sensitivity: the proportion of true rare-disease cases classified as low risk. |
| RareAlert | False-positive rate (FPR) | FP / (TN + FP) = 1 - specificity: the proportion of non-rare-disease cases classified as high risk. |
| RareAlert | Balanced accuracy | (Sensitivity + specificity) / 2. |
| RareAlert | F1 score | 2 x PPV x sensitivity / (PPV + sensitivity) = 2TP / (2TP + FP + FN). |
| RareAlert | F2 score | 5 x PPV x sensitivity / (4 x PPV + sensitivity), giving greater weight to sensitivity. |
| RareAlert | Matthews correlation coefficient (MCC) | (TP x TN - FP x FN) / sqrt[(TP + FP)(TP + FN)(TN + FP)(TN + FN)]. If the denominator is zero, the value is treated as undefined and reported according to the analysis convention. |
| **RareDiagnosis** | | |
| RareDiagnosis | NDCG@k (k = 1, 3, 5) | Calculated separately for primary diagnosis, follow-up diagnosis and further-test recommendation. DCG@k = Sum_{r=1}^k I(the item at rank r is correct) / log2(r + 1), and NDCG@k = DCG@k / IDCG@k. Correctness is determined against the reference standard. NDCG@k is set to 0 when no correct item is available for the ideal ranking. |
| RareDiagnosis | Mean reciprocal rank (MRR) | Calculated separately for primary diagnosis, follow-up diagnosis and further-test recommendation as (1/N) Sum_i 1/r_i, where r_i is the rank of the first correct candidate. A case contributes 0 when no correct candidate is retrieved. |
| RareDiagnosis | Delta-Acc@k | Acc@k_follow-up - Acc@k_primary among paired cases with data at both visits, reported in percentage points. |
| RareDiagnosis | Recovery@k | Among paired cases that are incorrect at the primary visit, the proportion that become correct at follow-up: N(primary incorrect and follow-up correct) / N(primary incorrect). |

| Module / analysis | Metric name | Definition and calculation |
|---|---|---|
| RareDiagnosis | Degradation@k | Among paired cases that are correct at the primary visit, the proportion that become incorrect at follow-up: N(primary correct and follow-up incorrect) / N(primary correct). |
| RareDiagnosis | Per-model Rescue Rate@k | For comparator model m, the proportion of cases it misses at top-k that RareDiagnosis answers correctly at the same visit and k: N(comparator_m incorrect and RareDiagnosis correct) / N(comparator_m incorrect). |
| RareDiagnosis | Multi-model Rescue Rate@k(n) | Among cases missed at top-k by at least n comparator LLMs, the proportion answered correctly by RareDiagnosis at the same visit and k. |
| **RareTreatment** | | |
| RareTreatment | P@T(k) / All-appropriate@k (k = 1, 3, 5) | P@T(k) (hereafter referred to as All-appropriate@k) = (1/N) Sum_i I(all of the first k treatment recommendations for case i are appropriate): the proportion of cases for which every recommendation within the first k is appropriate. This strict secondary metric is used in Supplementary Figure 20. |
| RareTreatment | NDCG@k (k = 1, 3, 5) | DCG@k = Sum_{r=1}^k I(the treatment at rank r is appropriate) / log2(r + 1), and NDCG@k = DCG@k / IDCG@k. Earlier appropriate treatments receive greater weight. NDCG@k is set to 0 when no appropriate treatment is available for the ideal ranking. |
| RareTreatment | Mean reciprocal rank (MRR) | (1/N) Sum_i 1/r_i, where r_i is the rank of the first appropriate treatment; the case contribution is 0 when no appropriate treatment is retrieved. |
| RareTreatment | Mean number of appropriate treatments per case | For each case, count recommendations judged clinically appropriate; report the mean (and, where shown, standard deviation) across cases for each model. |
| RareTreatment | Mean number of inappropriate treatments per case | For each case, count recommendations judged clinically inappropriate; report the mean (and, where shown, standard deviation) across cases for each model. |
| RareTreatment | High-risk treatment proportion | The number of treatment recommendations rated 2 or lower for safety divided by the number of recommendations with an available safety rating. |
| RareTreatment | Completeness score | A 1-5 ordinal rating of whether essential implementation elements are present; higher scores indicate greater completeness. Report the mean across the specified unit of analysis. |
| RareTreatment | Helpfulness score | A 1-5 ordinal rating of expected clinical benefit in context; higher scores indicate greater helpfulness. Report the mean across the specified unit of analysis. |
| RareTreatment | Safety score | A 1-5 ordinal rating of patient-specific risk; higher scores indicate greater safety. Report the mean across the specified unit of analysis. |
| RareTreatment | Case-level real-world treatment match | The proportion of cases for which at least one model-recommended treatment matches a treatment documented as performed in clinical practice. |
| RareTreatment | Within-case real-world treatment-match fraction | For each case, the number of recommended treatments matching documented performed treatments divided by the total number of recommended treatments; report the mean across cases. |
| RareTreatment | Per-model Rescue Rate@k | For comparator model m, the proportion of its no-hit cases at k for which RareTreatment achieves an any-hit at the same k. |
| RareTreatment | Multi-model Rescue Rate@k(n) | Among cases with no any-hit at k from at least n comparator LLMs, the proportion for which RareTreatment achieves an any-hit at the same k. |
| **RarePrognosis** | | |
| RarePrognosis | Multiclass Matthews correlation coefficient (MCC) | Calculated separately for overall outcome, functional status and symptom burden from the multiclass confusion matrix; higher values indicate stronger agreement between predicted and reference categories. |
| RarePrognosis | Severe-class recall | For each prognostic task, the proportion of truly high-risk cases predicted in the prespecified high-risk class. High-risk classes are progression/terminal for overall outcome, severe for functional status and persistent severe for symptom burden. |
| RarePrognosis | False-optimism rate | Among truly high-risk cases, the proportion predicted as a less severe category; equal to 1 - severe-class recall. |
| RarePrognosis | Mean absolute ordinal error (MAOE) | After mapping ordered categories to ranks of increasing severity, (1/N) Sum_i \|predicted rank_i - observed rank_i\|. |
| RarePrognosis | Ordinal bias index (OBI) | (1/N) Sum_i (predicted rank_i - observed rank_i). Positive values indicate pessimistic bias toward greater severity; negative values indicate optimistic bias. |
| RarePrognosis | Per-model Rescue Rate | For each comparator and prognostic task or joint endpoint, the proportion of cases missed by that comparator that RarePrognosis predicts correctly. |
| RarePrognosis | Multi-model Rescue Rate(n) | For each prognostic task or joint endpoint, the proportion correctly predicted by RarePrognosis among cases missed by at least n comparator LLMs. |
| **End-to-end evaluation** | | |
| End-to-end evaluation | End-to-end success rate | The proportion of rare-disease cases satisfying all prespecified pathway requirements: correct risk identification, a top-five primary diagnosis, a top-five follow-up diagnosis, an appropriate treatment within the top five recommendations and at least one correct prognostic assessment. |
| End-to-end evaluation | Cumulative success rate | At each pathway stage, the proportion of the fixed rare-disease denominator that has succeeded at that and every preceding stage; cases failing or not reaching a stage remain unsuccessful thereafter. |
| End-to-end evaluation | Assistance rescue rate | Among cases answered incorrectly without RareLens assistance, the proportion answered correctly with assistance. |

| Module / analysis | Metric name | Definition and calculation |
|---|---|---|
| End-to-end evaluation | Assistance regression rate | Among cases answered correctly without RareLens assistance, the proportion answered incorrectly with assistance. |
| **3. Inter-model heterogeneity metrics** | | |
| **RareAlert** | | |
| RareAlert | Pairwise binary-alert agreement | For each model pair, the proportion of shared cases assigned the same binary alert label. |
| RareAlert | Pairwise risk-score exact agreement | For each model pair, the proportion of evaluated cases with exactly identical numerical risk scores; missing or unequal values are treated as non-agreement. |
| RareAlert | Pairwise mean absolute risk-score difference | For each model pair, the mean absolute difference between risk scores across cases with valid scores from both models. |
| RareAlert | Reasoning-chain consensus rate | For each case, the frequency of the most common identical or prespecified-equivalent reasoning-chain category divided by the number of models with valid chains; average across cases. |
| RareAlert | Pairwise reasoning-chain agreement | For each model pair, the proportion of shared valid cases assigned the same or prespecified-equivalent reasoning-chain category. |
| RareAlert | Reasoning-chain count/class distribution | The frequency or proportion of cases assigned to each prespecified reasoning-chain count or category, summarized by model or disease-status group as shown. |
| RareAlert | Risk-score distribution by rare-disease status | Pool valid model-generated risk scores separately for rare and non-rare cases and report the within-group percentage in each prespecified score bin. |
| **RareDiagnosis** | | |
| RareDiagnosis | Candidate-set union size | For each case, the number of distinct diagnoses in the union of all source-LLM candidate lists, calculated separately for primary and follow-up visits. |
| RareDiagnosis | Mean pairwise top-5 Jaccard similarity | For model-pair candidate sets A and B, \|A intersection B\| / \|A union B\|; average across model pairs within each case and summarize across cases. |
| RareDiagnosis | Consensus strength | For each case, the maximum number of source LLMs supporting any one diagnosis divided by the number of source LLMs with valid candidate lists. |
| RareDiagnosis | Top-1 unique count (top-1 disagreement) | For each case, the number of distinct first-ranked diagnoses produced across source LLMs. |
| RareDiagnosis | Mean confidence range | Normalize confidence within each model's candidate list. For diagnoses shared by at least two models, calculate max normalized confidence - min normalized confidence; average across shared diagnoses within a case. |
| RareDiagnosis | Pairwise confidence divergence | For each model pair, the mean absolute normalized-confidence difference for diagnoses proposed by both models, averaged across eligible cases. |
| RareDiagnosis | Mean rank range | For diagnoses proposed by at least two source LLMs, calculate highest assigned rank - lowest assigned rank; average across shared diagnoses within each case. |
| RareDiagnosis | Pairwise ranking divergence | For each model pair, calculate the mean absolute rank difference for diagnoses proposed by both models within a case, then average across eligible cases. |
| **RareTreatment** | | |
| RareTreatment | Treatment-modality proportion | For each model and treatment category, the number of recommendations in that category divided by the model's total number of extracted recommendations. |
| RareTreatment | First-recommendation modality proportion | The same category proportion calculated using only the first-ranked treatment recommendation for each case. |
| RareTreatment | Mean recommendation count per case | For each model, count treatment recommendations per case and report the mean across cases. |
| RareTreatment | Cross-model clinical-quality profiles | Compare model-level high-risk proportions, real-world-treatment match measures and mean completeness, helpfulness and safety scores, each defined in Section 2. |
| RareTreatment | Case/candidate completeness correlation | For each baseline LLM, the Pearson correlation across cases between the overall case-level completeness rating and the within-case mean candidate-level completeness rating. |
| RareTreatment | Case/candidate completeness mean difference | For each baseline LLM, the mean across cases of overall case-level completeness minus the within-case mean candidate-level completeness. |
| RareTreatment | Case/candidate safety correlation | For each baseline LLM, the Pearson correlation across cases between the overall case-level safety rating and the within-case mean candidate-level safety rating. |
| RareTreatment | Case/candidate safety mean difference | For each baseline LLM, the mean across cases of overall case-level safety minus the within-case mean candidate-level safety. |
| RareTreatment | Importance-appropriateness association | For each model, the point-biserial correlation between 0-10 importance score and binary treatment appropriateness; 95% confidence intervals are obtained by Fisher z-transformation. |
| RareTreatment | Importance-completeness association | For each model, the fixed-effect coefficient for importance score from a linear mixed-effects model with candidate-level completeness as outcome and case ID as a random intercept; report a 95% confidence interval. |
| RareTreatment | Importance-helpfulness association | As above, using candidate-level helpfulness as the outcome. |
| RareTreatment | Importance-safety association | As above, using candidate-level safety as the outcome. |
| **RarePrognosis** | | |
| RarePrognosis | Pairwise categorical agreement | For each model pair and each of overall outcome, functional status and symptom burden, the proportion of shared cases with identical predicted categories. |

| Module / analysis | Metric name | Definition and calculation |
|---|---|---|
| **4. RareLens candidate-level scoring and integration metrics** | | |
| **RareDiagnosis** | | |
| RareDiagnosis | Candidate-score distribution | For primary-diagnosis, follow-up-diagnosis and further-test candidates, summarize score frequencies and/or densities, including the number of unique scores, total candidates, mean and standard deviation where reported. |
| RareDiagnosis | Correct/incorrect candidate-score distributions | Compare the score distributions of candidates judged correct versus those judged incorrect. |
| RareDiagnosis | Within-case score separation | For each eligible case, mean score of correct candidates - mean score of incorrect candidates, calculated separately for primary diagnosis, follow-up diagnosis and further-test recommendation. Cases lacking one candidate class are excluded. |
| RareDiagnosis | Contributing source-model count@k (k = 1, 3, 5) | For each case, the number of distinct source LLMs contributing at least one diagnosis to the first k RareDiagnosis candidates after unioning the source-model sets attached to those candidates; calculated separately at primary and follow-up visits. |
| **RareTreatment** | | |
| RareTreatment | Treatment candidate-score distribution | Summarize RareTreatment candidate-score frequencies and/or densities, including the number of unique scores, total candidates, mean and standard deviation where reported. |
| RareTreatment | Appropriate/inappropriate candidate-score distributions | Compare the score distributions of treatment candidates judged appropriate versus inappropriate. |
| RareTreatment | Within-case treatment score separation | For each case containing both candidate classes, mean score of appropriate candidates - mean score of inappropriate candidates; cases containing only one class are excluded. |
| RareTreatment | Supporting source-model count@k (k = 1, 3, 5) | For each case and k, the number of distinct source LLMs supporting at least one of the first k RareTreatment recommendations; summarize the count distribution and case proportions. |

# Supplementary Table 2 | Run-to-run agreement of principal RareLens outputs

| Module / setting | Output | Level / field | N | Observed agreement ($p_o$) | Mean pairwise Cohen's κ | Fleiss' κ | Gwet's AC1 | ICC(2,1) |
|---|---|---|---|---|---|---|---|---|
| **Risk Screening** | | | | | | | | |
| RareAlert | Risk score | Continuous score (0–100) | 500 | — | — | — | — | 0.930 |
| | Binary screening result | Youden threshold = 70 | 500 | — | 0.895 | 0.895 | 0.897 | — |
| **Diagnosis** | | | | | | | | |
| RareDiagnosis | Correct diagnosis | hit@1 | 500 | 0.806 | 0.740 | 0.740 | 0.743 | — |
| (Primary visit) | | hit@3 | 500 | 0.812 | 0.736 | 0.736 | 0.762 | — |
| | | hit@5 | 500 | 0.850 | 0.776 | 0.776 | 0.819 | — |
| RareDiagnosis | Correct diagnosis | hit@1 | 500 | 0.800 | 0.625 | 0.625 | 0.793 | — |
| (Follow-up visit) | | hit@3 | 500 | 0.880 | 0.653 | 0.654 | 0.896 | — |
| | | hit@5 | 500 | 0.914 | 0.680 | 0.681 | 0.930 | — |
| **Treatment** | | | | | | | | |
| RareTreatment | Appropriate treatment | hit@1 | 456 | 0.770 | 0.221 | 0.220 | 0.809 | — |
| | | hit@3 | 456 | 0.969 | 0.375 | 0.381 | 0.979 | — |
| | | hit@5 | 456 | 0.982 | 0.416 | 0.423 | 0.988 | — |
| **Prognosis** | | | | | | | | |
| RarePrognosis | Outcome category | Ordinal | 500 | 0.820 | 0.879 | 0.821 | 0.852 | — |
| | Functional status | Ordinal | 500 | 0.792 | 0.885 | 0.809 | 0.814 | — |
| | Symptom burden | Ordinal | 500 | 0.822 | 0.845 | 0.820 | 0.847 | — |

*Note: Each module was run independently three times. N denotes the number of eligible cases included in each agreement analysis.*

*Cohen's κ values are reported as the mean of the three pairwise run comparisons (run 1 versus run 2, run 1 versus run 3, and run 2 versus run 3). Unweighted Cohen's κ was used for binary screening and hit@k outcomes, whereas linearly weighted Cohen's κ was used for ordinal prognostic outcomes. Standard unweighted Fleiss' κ was calculated across all three runs, treated as three raters. Fleiss' κ treats the three runs as three raters.*

*Gwet's AC1 is reported as a prevalence-robust complement to κ. For diagnosis and treatment, hit@k indicates whether a correct diagnosis or appropriate treatment appeared among the top k outputs.*

*Binary screening used the Youden-index operating threshold (risk score = 70). $p_o$ denotes observed agreement; —, not applicable or not reported in the source calculation table.*

## Supplementary Table 3a | Comparison of simple ensemble methods and RareLens for risk screening

| Risk screening | | | |
|---|---|---|---|
| **Method** | **n** | **AUROC** | **AUPRC** |
| **Voting** | | | |
| Majority vote (positive-vote fraction) | 31506 | 0.8985 | 0.8067 |
| **Min–max normalized score aggregation** | | | |
| Mean normalized score | 31506 | 0.9128 | 0.8365 |
| Median normalized score | 31506 | 0.9022 | 0.8067 |
| **Raw-score scale-sensitivity controls** | | | |
| Mean raw score | 31506 | 0.9132 | 0.8367 |
| Median raw score | 31506 | 0.8956 | 0.7547 |
| Mode raw score | 31506 | 0.8351 | 0.6168 |
| **RareLens aggregator** | | | |
| **RareLens** | **31506** | **0.9173** | **0.8409** |

## Supplementary Table 3b | Comparison of simple ensemble methods and RareLens for diagnosis candidate ranking

| Primary diagnosis | | | | | | | | |
|---|---|---|---|---|---|---|---|---|
| **Method** | **n** | **Acc@1** | **Acc@3** | **Acc@5** | **MRR** | **nDCG@1** | **nDCG@3** | **nDCG@5** |
| **Score fusion** | | | | | | | | |
| CombSUM | 8696 | 0.5321 | 0.6946 | 0.7455 | 0.6275 | 0.5321 | 0.4749 | 0.5005 |
| Mean | 8696 | 0.5321 | 0.6946 | 0.7455 | 0.6275 | 0.5321 | 0.4749 | 0.5005 |
| CombMNZ | 8696 | 0.5048 | 0.6869 | 0.7399 | 0.6098 | 0.5048 | 0.4549 | 0.4806 |
| CombANZ | 8696 | 0.4656 | 0.6494 | 0.7198 | 0.5766 | 0.4656 | 0.4661 | 0.4991 |
| **Voting** | | | | | | | | |
| Hit-sum | 8696 | 0.4552 | 0.6677 | 0.7284 | 0.5764 | 0.4552 | 0.4224 | 0.4485 |
| **Rank fusion** | | | | | | | | |
| RRF (K = 60) | 8696 | 0.4817 | 0.6818 | 0.7401 | 0.5962 | 0.4817 | 0.4419 | 0.4714 |
| Borda | 8696 | 0.5435 | 0.6978 | 0.7513 | 0.6359 | 0.5435 | 0.4878 | 0.5163 |
| Length-normalized Borda | 8696 | 0.5405 | 0.6957 | 0.7479 | 0.6329 | 0.5405 | 0.4847 | 0.5126 |
| **Central tendency over nominating models** | | | | | | | | |
| Median | 8696 | 0.5314 | 0.6780 | 0.7314 | 0.6206 | 0.5314 | 0.5092 | 0.5335 |
| Mode | 8696 | 0.5440 | 0.6885 | 0.7421 | 0.6316 | 0.5440 | 0.5206 | 0.5460 |
| **RareLens aggregator** | | | | | | | | |
| **RareLens** | **8696** | **0.6552** | **0.7594** | **0.7963** | **0.7102** | **0.6552** | **0.7129** | **0.7201** |

| Follow-up diagnosis | | | | | | | | |
|---|---|---|---|---|---|---|---|---|
| **Method** | **n** | **Acc@1** | **Acc@3** | **Acc@5** | **MRR** | **nDCG@1** | **nDCG@3** | **nDCG@5** |
| **Score fusion** | | | | | | | | |
| CombSUM | 8384 | 0.7587 | 0.9139 | 0.9418 | 0.8401 | 0.7587 | 0.6404 | 0.6572 |
| Mean | 8384 | 0.7587 | 0.9139 | 0.9418 | 0.8401 | 0.7587 | 0.6404 | 0.6572 |
| CombMNZ | 8384 | 0.7185 | 0.9048 | 0.9382 | 0.8155 | 0.7185 | 0.6104 | 0.6284 |
| CombANZ | 8384 | 0.6896 | 0.8692 | 0.9234 | 0.7895 | 0.6896 | 0.6563 | 0.6821 |
| **Voting** | | | | | | | | |
| Hit-sum | 8384 | 0.6471 | 0.8825 | 0.9302 | 0.7696 | 0.6471 | 0.5607 | 0.5800 |
| **Rank fusion** | | | | | | | | |
| RRF (K = 60) | 8384 | 0.6805 | 0.8966 | 0.9374 | 0.7927 | 0.6805 | 0.5870 | 0.6100 |
| Borda | 8384 | 0.7680 | 0.9172 | 0.9448 | 0.8470 | 0.7680 | 0.6563 | 0.6774 |
| Length-normalized Borda | 8384 | 0.7610 | 0.9132 | 0.9413 | 0.8420 | 0.7610 | 0.6528 | 0.6736 |
| **Central tendency over nominating models** | | | | | | | | |
| Median | 8384 | 0.7741 | 0.8979 | 0.9327 | 0.8436 | 0.7741 | 0.7115 | 0.7252 |
| Mode | 8384 | 0.7858 | 0.9059 | 0.9362 | 0.8525 | 0.7858 | 0.7239 | 0.7374 |
| **RareLens aggregator** | | | | | | | | |
| **RareLens** | **8384** | **0.8835** | **0.9454** | **0.9612** | **0.9153** | **0.8835** | **0.9152** | **0.9117** |

## Supplementary Table 3c | Comparison of simple ensemble methods and RareLens for treatment candidate ranking

| Treatment | | | | | | | | | | |
|---|---|---|---|---|---|---|---|---|---|---|
| **Method** | **n** | **Acc@1** | **Acc@2** | **Acc@3** | **Acc@5** | **MRR** | **nDCG@1** | **nDCG@2** | **nDCG@3** | **nDCG@5** |
| **Score fusion** | | | | | | | | | | |
| CombSUM | 452 | 0.8496 | 0.8938 | 0.9248 | 0.9447 | 0.8906 | 0.8496 | 0.8393 | 0.8324 | 0.8309 |
| Mean | 452 | 0.8496 | 0.8938 | 0.9248 | 0.9447 | 0.8906 | 0.8496 | 0.8393 | 0.8324 | 0.8309 |
| CombMNZ | 452 | 0.8496 | 0.8960 | 0.9270 | 0.9535 | 0.8924 | 0.8496 | 0.8410 | 0.8337 | 0.8340 |
| CombANZ | 452 | 0.7987 | 0.8606 | 0.8916 | 0.9292 | 0.8530 | 0.7987 | 0.8081 | 0.8080 | 0.8026 |
| **Voting** | | | | | | | | | | |
| Hit-sum | 452 | 0.8473 | 0.9049 | 0.9314 | 0.9535 | 0.8929 | 0.8473 | 0.8405 | 0.8307 | 0.8145 |
| **Rank fusion** | | | | | | | | | | |
| RRF (K = 60) | 452 | 0.8496 | 0.8960 | 0.9270 | 0.9558 | 0.8924 | 0.8496 | 0.8410 | 0.8347 | 0.8333 |
| Borda | 452 | 0.8407 | 0.8938 | 0.9226 | 0.9336 | 0.8844 | 0.8407 | 0.8339 | 0.8303 | 0.8160 |
| Length-normalized Borda | 452 | 0.8385 | 0.8916 | 0.9137 | 0.9381 | 0.8821 | 0.8385 | 0.8325 | 0.8292 | 0.8177 |
| **Central tendency over nominating models** | | | | | | | | | | |
| Median | 452 | 0.8142 | 0.8739 | 0.9004 | 0.9314 | 0.8641 | 0.8142 | 0.8193 | 0.8171 | 0.8088 |
| Mode | 452 | 0.8274 | 0.8739 | 0.9049 | 0.9336 | 0.8712 | 0.8274 | 0.8240 | 0.8196 | 0.8120 |
| **RareLens aggregator** | | | | | | | | | | |
| **RareLens** | **452** | **0.8982** | **0.9403** | **0.9513** | **0.9668** | **0.9293** | **0.8982** | **0.8928** | **0.8882** | **0.8870** |

# Supplementary Table 3d | Comparison of simple ensemble methods and RareLens for prognosis prediction

| Joint accuracy across all three dimensions | | |
|---|---|---|
| **Method** | **n** | **Accuracy** |
| **Voting** | | |
| Mode | 3802 | 0.2720 |
| Confidence-weighted vote | 3802 | 0.2728 |
| **Ordinal central tendency** | | |
| Ordinal mean | 3802 | 0.2244 |
| Ordinal median | 3802 | 0.2541 |
| Confidence-weighted median | 3802 | 0.2593 |
| **RareLens aggregator** | | |
| **RareLens** | **3802** | **0.3385** |

| Functional status | | | | | | | |
|---|---|---|---|---|---|---|---|
| **Method** | **n** | **Accuracy** | **MCC** | **MAOE** | **OBI** | **Severe recall** | **False optimism** |
| **Voting** | | | | | | | |
| Mode | 3802 | 0.5544 | 0.4317 | 0.5066 | 0.2530 | 0.6727 | 0.3273 |
| Confidence-weighted vote | 3802 | 0.5547 | 0.4319 | 0.5066 | 0.2530 | 0.6727 | 0.3273 |
| **Ordinal central tendency** | | | | | | | |
| Ordinal mean | 3802 | 0.5360 | 0.4139 | 0.5274 | 0.2549 | 0.6145 | 0.3855 |
| Ordinal median | 3802 | 0.5513 | 0.4301 | 0.5060 | 0.2294 | 0.6355 | 0.3645 |
| Confidence-weighted median | 3802 | 0.5523 | 0.4290 | 0.5095 | 0.2549 | 0.6717 | 0.3283 |
| **RareLens aggregator** | | | | | | | |
| **RareLens** | **3802** | **0.6331** | **0.5097** | **0.4335** | **0.0147** | **0.7480** | **0.2520** |

| Overall outcome | | | | | | | |
|---|---|---|---|---|---|---|---|
| **Method** | **n** | **Accuracy** | **MCC** | **MAOE** | **OBI** | **Severe recall** | **False optimism** |
| **Voting** | | | | | | | |
| Mode | 3802 | 0.5702 | 0.4334 | 0.5781 | -0.1241 | 0.6089 | 0.3911 |
| Confidence-weighted vote | 3802 | 0.5705 | 0.4337 | 0.5779 | -0.1234 | 0.6096 | 0.3904 |
| **Ordinal central tendency** | | | | | | | |
| Ordinal mean | 3802 | 0.5171 | 0.3848 | 0.5960 | -0.1052 | 0.4825 | 0.5175 |
| Ordinal median | 3802 | 0.5521 | 0.4126 | 0.5897 | -0.1689 | 0.5387 | 0.4613 |
| Confidence-weighted median | 3802 | 0.5555 | 0.4193 | 0.5744 | -0.1047 | 0.5782 | 0.4218 |
| **RareLens aggregator** | | | | | | | |
| **RareLens** | **3802** | **0.6236** | **0.5008** | **0.5279** | **-0.0581** | **0.7317** | **0.2683** |

| Symptom burden | | | | | | | |
|---|---|---|---|---|---|---|---|
| **Method** | **n** | **Accuracy** | **MCC** | **MAOE** | **OBI** | **Severe recall** | **False optimism** |
| **Voting** | | | | | | | |
| Mode | 3802 | 0.5242 | 0.3971 | 0.6481 | 0.3051 | 0.6059 | 0.3941 |
| Confidence-weighted vote | 3802 | 0.5252 | 0.3983 | 0.6470 | 0.3046 | 0.6065 | 0.3935 |
| **Ordinal central tendency** | | | | | | | |
| Ordinal mean | 3802 | 0.4834 | 0.3531 | 0.6757 | 0.3017 | 0.5630 | 0.4370 |
| Ordinal median | 3802 | 0.5016 | 0.3732 | 0.6528 | 0.2762 | 0.5684 | 0.4316 |
| Confidence-weighted median | 3802 | 0.5124 | 0.3814 | 0.6473 | 0.3012 | 0.6045 | 0.3955 |
| **RareLens aggregator** | | | | | | | |
| **RareLens** | **3802** | **0.6704** | **0.5288** | **0.4829** | **0.0473** | **0.7937** | **0.2063** |

## Supplementary Note 1 | Specialty taxonomy of RarelensBench cases

Clinical specialty was assigned using GPT-5, which selected the most appropriate category from a predefined specialty list based on the disease name. The taxonomy comprised the following categories:

| | | |
|---|---|---|
| Addiction Medicine | Aerospace Medicine | Allergy and Immunology |
| Anesthesiology | Cardiology | Colon and Rectal Surgery |
| Dermatology | Emergency Medicine | Endocrinology |
| Family Medicine | Gastroenterology | General Surgery |
| Hematology | Hepatology | Immunology |
| Infectious Disease | Lymphology | Medical Genetics and Genomics |
| Medical Toxicology | Nephrology | Neurological Surgery |
| Neurology | Nutrition | OBGYN |
| Odontology | Oncology | Ophthalmology |
| Oral and Maxillofacial Surgery | Orthopaedic Surgery | Otolaryngology-Head and Neck Surgery |
| Pain Medicine | Pathology | Pediatrics |
| Physical Medicine and Rehabilitation | Plastic Surgery | Psychiatry |
| Pulmonology | Radiology | Reproductive Medicine |
| Rheumatology | Sleep Medicine | Teratology |
| Thoracic Surgery | Transplant Medicine | Urology |
| Vascular Medicine | Vascular Surgery | |